\documentclass[lettersize,journal]{IEEEtran}
\usepackage{amsmath,amsfonts}
\usepackage[ruled,linesnumbered]{algorithm2e}
\usepackage{algorithmic}
\usepackage{array}
\usepackage{textcomp}
\usepackage{stfloats}
\usepackage{url}
\usepackage{verbatim}
\usepackage{graphicx}
\usepackage{cite}
\usepackage{makecell}

\usepackage{mathrsfs}
\usepackage{color}
\usepackage{multirow}
\usepackage[dvipsnames]{xcolor}
\usepackage{xspace}
\usepackage{tikz}
\usepackage{subfigure}
\usepackage{pifont}
\usepackage{tabularx}
\usepackage{threeparttable}
\usepackage{mathtools}
\usepackage[colorlinks,
            urlcolor=blue,
            linkcolor=blue,       
            anchorcolor=blue,  
            citecolor=green        
            ]{hyperref}
\usepackage{comment}
\usepackage{CJK}

\newcommand{\dnote}[1]{ \color{purple} [Yifan: #1]  \color{black}}
\newcommand{\xnote}[1]{ \color{violet} [Xinran: #1]  \color{black}}

\definecolor{c_m}{HTML}{E7DAD2}
\definecolor{c_b}{HTML}{82B0D2}
\definecolor{c_o}{HTML}{FFBE7A}
\definecolor{c_q}{HTML}{8ECFC9}
\definecolor{c_p}{HTML}{BEB8DC}
\definecolor{c_r}{HTML}{FA7F6F}

\newcommand{\sd}{TERRA\xspace}
\newcommand{\je}{JEEP\xspace}
\newcommand{\fl}{\mathcal{FL}\xspace}
\newcommand{\ho}{hierarchical optimization\xspace}
\newcommand{\ts}{time synchronization\xspace}
\newcommand{\hots}{HOTS\xspace}
\newcommand{\etal}{\textit{et al.}\@}

\begin{document}

\title{Rotation Initialization and  Stepwise Refinement for Universal LiDAR Calibration}
\author{Yifan Duan, Xinran Zhang, Guoliang
You, Yilong Wu, Xingchen Li, Yao Li, Xiaomeng Chu,  Jie Peng, \\Yu Zhang, Jianmin Ji, and Yanyong Zhang*~\IEEEmembership{Fellow,~IEEE,}

\thanks{* The corresponding author.}
\thanks{School of Computer Science and Technology, University of Science and Technology of China, Hefei, 230026, China
{\tt\small \{dyf0202, zxrr, glyou, elonwu, starlet, zkdly, cxmeng, pengjieb\}@mail.ustc.edu.cn, \{yuzhang, jianmin, yanyongz\}@ustc.edu.cn}.}%
}

\markboth{Journal of \LaTeX\ Class Files,~Vol.~14, No.~8, August~2021}%
{Shell \MakeLowercase{\textit{et al.}}: A Sample Article Using IEEEtran.cls for IEEE Journals}

\IEEEpubid{0000--0000/00\$00.00~\copyright~2021 IEEE}

\maketitle

\begin{abstract}
Autonomous systems often employ multiple LiDARs to leverage the integrated advantages,  enhancing perception and robustness.
The most critical prerequisite under this setting is the estimating the extrinsic between each LiDAR, i.e.,  calibration. Despite the exciting progress in multi-LiDAR calibration efforts, a universal, sensor-agnostic calibration method  remains elusive. According to the coarse-to-fine framework, we first design a spherical descriptor \sd for 3-DoF  rotation initialization with no prior knowledge.
To further optimize, we present \je for the joint estimation of extrinsic and pose, integrating geometric and motion information to overcome  factors affecting the point cloud registration. Finally, the  LiDAR poses optimized by the \ho module are input to \ts module to produce the ultimate calibration results, including the time offset.
To verify the effectiveness, we conduct extensive experiments on eight datasets, where 16 diverse types of LiDARs in total and dozens of calibration tasks are tested. In the challenging tasks, the calibration errors can still be controlled within $5cm$ and $1^\circ$ with a high success rate.

\end{abstract}

\section{Introduction}
\IEEEPARstart{I}{n} recent years, the field of 3D vision technology has witnessed remarkable advancements, leading to the widespread adoption of depth sensors such as LiDARs for their ability to obtain accurate depth information~\cite{LOAM, livoxloam}. 
As the demand for enhanced performance continues to grow, there is a notable trend toward equipping  autonomous systems, like self-driving vehicles and robotics, with multiple LiDARs. 
This approach brings numerous benefits, including improved robustness in self-localization and mapping, as well as the ability to prevent system crashes resulting from a single hardware failure~\cite{lin2020decentralized}. Moreover, the utilization of multiple LiDARs enables the generation of a denser point cloud with a wider field of view (FoV), thereby significantly enhancing the overall perception performance~\cite{Hu_2022_CVPR}.
Furthermore, 
by carrying various types of LiDARs, the entire perception system can leverage the unique characteristics of different LiDARs to enhance the performance. 
For example, conventional mechanical LiDARs provide a $360^{\circ}$ horizontal FoV, ensuring uniform spatial point cloud distribution~\cite{velodyne}. Additionally, solid-state LiDARs are becoming increasingly popular in autonomous driving applications, offering denser point clouds within a limited FoV at a relatively lower cost~\cite{livoxloam, r2live, lin2021r3live}.  

To effectively utilize the benefits of multiple LiDAR, it is crucial to establish a common coordinate system by accurately transforming the data from their respective individual coordinate systems. This transformation between different LiDAR coordinate systems is parameterized by the extrinsic parameters, and the process of finding the optimal parameters is known as calibration, 
which plays a pivotal role in ensuring aligned and synchronized data from each sensor.
\begin{figure}[t]
	\centering
	\includegraphics[width = \linewidth]{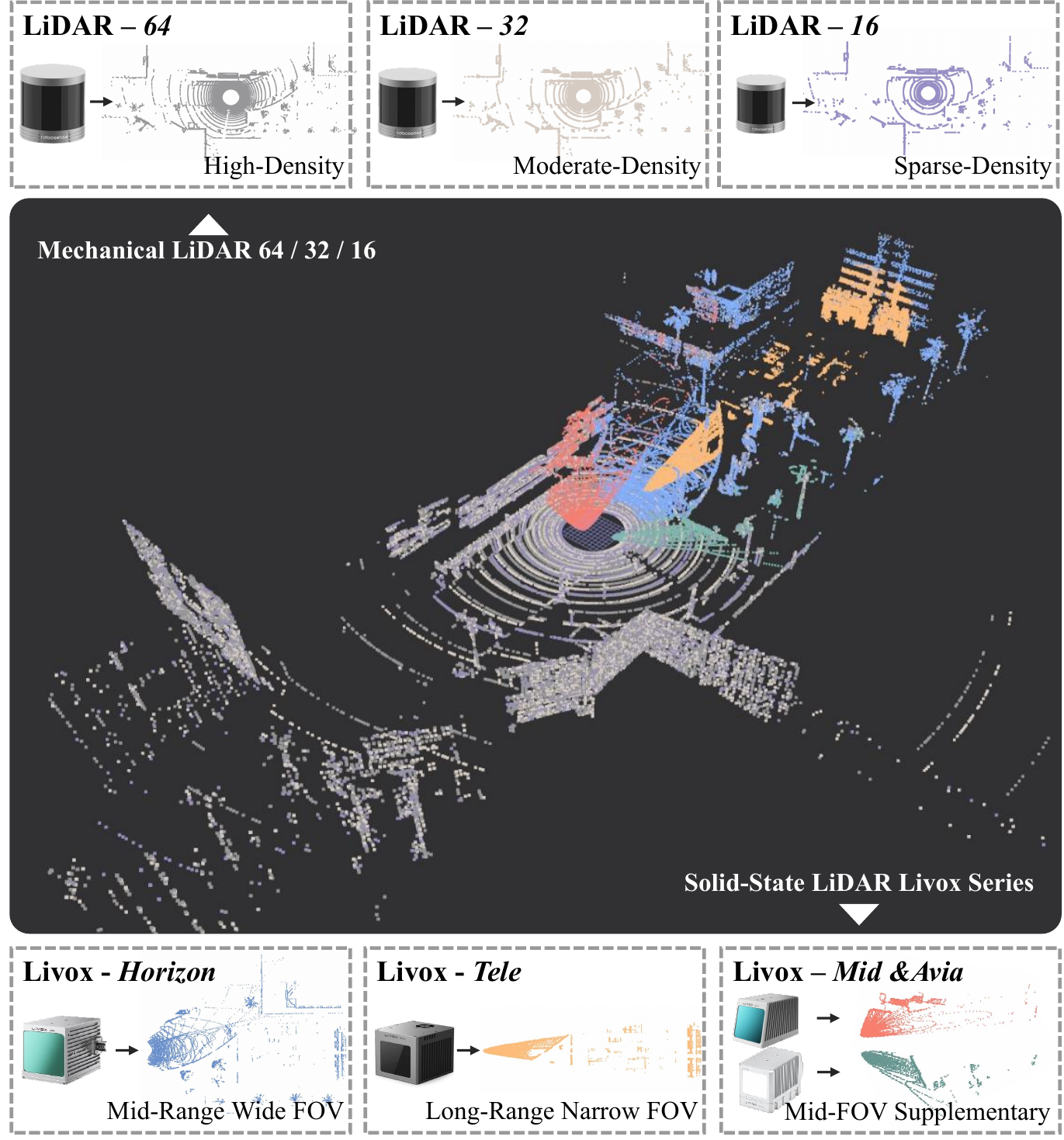}
	\caption{The illustration of seven types of LiDAR in the simulated dataset. Mechanical LiDARs exhibit varying densities, solid-state LiDARs have different FOV, and mechanical and solid-state LiDARs have distinctly different point cloud distributions. This diversity presents new challenges for LiDAR calibration. This paper aims to introduce a universal, sensor-agnostic calibration method.
	}
	\label{fig:fig1}
\end{figure}
\IEEEpubidadjcol
Calibration is a widely researched topic, and we mainly focus on the calibration of LiDARs in this paper.

Existing LiDAR calibration methods typically employ a coarse-to-fine framework to solve for the calibration parameters  progressively.  Finding an appropriate initial value is a prerequisite for obtaining precise parameters, which can be done by  utilizing  Hand-Eye method~\cite{mloam} or prior knowledge of specific scenarios~\cite{CROON}. 
After the initialization step, it's commonly to perform a calibration  refinement process to further improve the accuracy of calibration parameters. Specifically, the fundamental principle of the refinement step is to identify elements that can be correlated between  LiDARs, such as similar features in the point cloud~\cite{jiao2019novel,ma2022automatic} or their similar motion trajectories~\cite{das2022extrinsic,das2023observability}.  Therefore, the refinement step is typically modeled as either a point cloud registration problem or a trajectory alignment problem. It is worth noting that the existing methods are usually tailored for specific LiDAR type, i.e., leveraging the prior knowledge of point cloud distribution or  designing corresponding odometry module for pose estimation.

Although multi-LiDAR calibration has been extensively researched, it becomes increasingly challenging when it comes to the calibration between LiDARs with different types as shown in Fig.~\ref{fig:fig1}. First, the FoV of different LiDARs varies significantly. For instance, the Tele\footnote{https://www.livoxtech.com/tele-15} from Livox has only $14.5^\circ*16.2^\circ$ FoV, whereas the mechanical LiDAR\footnote{https://www.robosense.ai/en/rslidar/RS-Helios} has a $360^\circ*30^\circ$ FoV. Additionally,
due to the different scanning methods of LiDAR, the distribution of point clouds generated by different LiDARs for the same area also differ. Lastly, the overlap between two LiDARs may be limited, as a primary motivation for employing multiple LiDARs is to expand the FoV of the agent. These factors present new challenges to the coarse-to-fine LiDAR calibration framework. The different distribution and density of point clouds make it challenging to develop a universal initialization method that can consistently produce qualified initial values for any two LiDARs. Additionally, the minimal overlap and sparsity of the point cloud contribute to instability in the results during further precise extrinsic parameter refinement, even if the calibration converges. 

To effectively address the aforementioned challenges, 
we propose a new calibration approach for multiple LiDARs of any types. To this end, the issue of parameter initialization needs to be addressed first.
 According to the experiments in Sec.~\ref{sec:exp_refine_abla}, we highlight the importance of initializing rotation parameters rather than translation parameters, given the paper's concentration on single-platform (one vehicle or one robot) sensor calibration.
Therefore, we first introduce a point cloud descriptor for rotation initialization, namely \textbf{\sd}, i.e., a \textbf{T}hree degree of freedom (DoF) \textbf{E}mbedding of point cloud for \textbf{R}obust \textbf{R}otation \textbf{A}lignment.
We carefully design a distance function between multiple TERRAs to overcome the impact of translation on rotation estimation robustly.
After initialization, to address poor registration accuracy caused by the sparsity of the point cloud and scene degeneracy, a method called \textbf{\je} for the \textbf{J}oint \textbf{E}stimation of \textbf{E}xtrinsic and \textbf{P}oses is proposed to obtain optimized, usable extrinsics by integrating geometric and motion information.  To further refine these parameters  with all  sequence data from LiDARs, we implement a straightforward multi-LiDAR SLAM approach and utilize the \ho to optimize the pose of each point cloud frame, subsequently  enhancing the accuracy of the extrinsics estimation. In this way, we also estimate the offset between the data reception time of the two LiDARs by a proposed \ts module.

We conduct experiments on eight datasets, including both simulated and real-world environments with 16 diverse types of LiDARs in total and dozens of LiDAR pairs. Results show our rotation initialization and stepwise refinement methods have strong universality for different calibration tasks. We also conduct comprehensive ablation experiments on the key parameters involved in the methods. For typical calibration tasks for mechanical LiDARs, the calibration error can be controlled within $1cm$ and $0.1^\circ$ and  in tasks involving challenging factors such as partial overlap of point clouds, calibration errors can still be controlled within $5cm$ and $1^\circ$ with a high success rate.

In summary, our main contributions are as follows:
\begin{itemize}
    \item We propose a novel method for universal LiDAR calibration, regardless of the specific type or model of LiDAR. 
    \item We design a point cloud descriptor namely \sd, i.e., a Three-DoF Embedding of point cloud for Robust Rotation Alignment, to perform the rotation parameters initialization between any two LiDARs effectively.
    \item To estimate the extrinsic parameters accurately, we introduce a module called \je, named after the Joint Estimation of Extrinsic and Poses, integrating geometric and motion information to overcome factors affecting the point cloud registration. Following that,  we utilize the \ho module and \ts module to further enhance the accuracy of extrinsics and time offset estimation.
    \item To fully verify the effectiveness of our method, we conduct extensive experiments on eight datasets, including both simulated and real-world environments, where 16 diverse types of LiDARs in total and dozens of LiDAR pairs are tested. The  code and related dataset will be open sourced on \url{https://github.com/yjsx/ULC}.
\end{itemize}

\section{Related Work}
\subsection{LiDAR-LiDAR Calibration}
In this section, we primarily discuss the existing calibration methods for LiDAR-LiDAR. We also briefly introduce the calibration between other sensors due to the similarities among calibration tasks. We divide these methods into two groups: geometry-based methods and motion-based methods.

\subsubsection{Geometry-based methods} Multi-LiDAR calibration is essentially a process to estimate the relative positions of two LiDARs, typically addressed through point cloud registration~\cite{gong2017target}. Point cloud registration is a widely studied problem, including classic methods like point-point ICP~\cite{icp}, NDT~\cite{ndt}, point-plane ICP~\cite{pointplaneicp}, and generalized-ICP~\cite{gicp} etc. However, the accuracy of point cloud registration is influenced by numerous factors, such as the initial transformation, scene degeneracy, and point cloud bias and noise~\cite{covariance1, covariance2, covariance3}. In calibration tasks where the initial values are missing and point cloud distribution varies, these factors become challenges in the solving process. 
Leveraging the typical characteristics of road scenes, CROON~\cite{CROON} initializes the extrinsic parameters by aligning the ground plane, followed by using an octree-based refinement method to optimize the LiDARs' pose. Jiao \etal~\cite{jiao2019novel} and Ma \etal~\cite{ma2022automatic} carefully select the environment for calibration and introduce an automatic multi-LiDAR extrinsic calibration algorithm using corner planes. Brahayam \etal~\cite{ponton2022efficient}  also focus on street scenes, selecting the ground and commonly seen poles as reference objects. By matching and aligning, they solve the 6-DoF parameters in a stepwise manner.
GMMCalib~\cite{tahiraj2024gmmcalib} implements a GMM-based joint registration algorithm to enhance the robustness of the calibration process. 

In multi-LiDAR calibration tasks, a challenge arises when the overlapping areas between LiDARs are minimal, leading to insufficient constraints for point cloud registration. Nie \etal~\cite{nie2022automatic} developed a novel surface normal estimation method for two mechanically tilted placed LiDARs taking account of the uneven distribution of point cloud density and the edge information of planes.  LiDAR-Link~\cite{LiDAR_link}  employs a large-FoV LiDAR as a ``LiDAR-Bridge'' to connect other small-FoV LiDARs, subsequently introducing the LiDAR-Align method to collect point clouds across multiple scenes and utilize the Generalized-ICP in parallel spaces to obtain the extrinsics. 
For small-FoV solid-state LiDAR calibration, the challenge intensifies as there may be no overlapping areas between the LiDARs. To address this problem, MLCC~\cite{small_fov, liu2022targetless} generates overlapping areas by constructing a local map of each LiDAR and formulate the multiple LiDAR extrinsic calibration into a LiDAR bundle adjustment (BA) problem. Heng \etal~\cite{heng2020automatic} also employs the approach of constructing maps to calibrate LiDARs and millimeter-wave radars. 

\subsubsection{Motion-based methods}Since calibration tasks typically involve sensors rigidly connected, the relative motion between two sensors can be established using the hand-eye method~\cite{handeye, hausman2017observability}.
Das \etal~\cite{das2022extrinsic}\cite{das2023observability} estimate the motion trajectories of each sensor separately and recover the initial rotation and translation by aligning these trajectories, subsequently optimizing them using dual quaternion-based hand-eye calibration. 
Taylor \etal~\cite{taylor2016motion} propose a universal method for LiDARs, cameras, GPS, and IMU that sequentially solves for time offsets, rotational offsets, and translational offsets based on motion information. Hand-Eye calibration is also a method used to assign initial values to extrinsic parameters~\cite{jiao2019novel, jiaojianhao2019automatic, mloam}, which typically requires specific trajectories and sufficient accuracy. 

In this paper, 
we emphasize and solely focus on the  rotation initialization. In the refinement phase, we integrate the advantages of geometry-based and motion-based methods to overcome the challenges affecting point cloud registration. Furthermore, through backend optimization, we enhance the calibration accuracy.

\subsection{Point Cloud Descriptor}
Extracting descriptor from point cloud is a common method for transforming point clouds into feature spaces, with numerous applications such as place recognition and pose estimation.  We primarily focus on descriptors generated through projection and categorized them based on the projection method into two types: BEV (Bird's Eye View) projection-based methods and spherical projection-based methods.

\subsubsection{BEV Projection Based Methods}Projecting point clouds onto the BEV plane is a common approach in descriptor generation. He \etal 
~propose M2DP~\cite{he2016m2dp}, which is generated by projecting a 3D point cloud to multiple 2D planes with density signatures and generating the descriptor of the 3D point cloud using the signatures. Scancontext~\cite{scancontext}  partitions the point cloud into bins on the x-y plane based on both azimuthal and radial directions. This process creates a two-dimensional matrix to facilitate global position recognition by finding the optimal rotation for rotational invariance. Their sequel, ScanContext++~\cite{kim2021scan}, introduced a new spatial division strategy to achieve lateral invariance and utilized a sub-descriptor to speed up loop retrieval, showing promising results in urban environments. More recently, some methods~\cite{luo2023bevplace, fu2023overlapnetvlad} use neural networks to generate BEV features and conduct end-to-end training to achieve place recognition.

\subsubsection{Spherical Projection Based Methods} 
PHASER~\cite{bernreiter2021phaser} projects both source and target point cloud onto the sphere and performs a spherical Fourier transform to find the rotation that maximizes the spherical correlation. Similarly, Han \etal ~propose DiLO~\cite{han2021dilo} implementing LiDAR odometry based on the spherical range image (SRI). Shan \etal~\cite{shan2021robust} utilize the traditional BoW algorithm in visual place recognition for LiDAR-based global  localization. This approach involves a projection of high-resolution LiDAR point cloud intensity data into a spherical visual format, facilitating the extraction of features based on ORB~\cite{rublee2011orb}. SphereVLAD++~\cite{zhao2022spherevlad++} processes a query LiDAR scan by converting it into a panorama by spherical projection and encodes local orientation-equivalent features via spherical convolution for descriptor generation.  OverlapNet~\cite{chen2021overlapnet} takes spherical projections of 3D LiDAR data as input and estimates the overlap between two range images by calculating all possible differences for each pixel. Bernreiteret \etal~\cite{bernreiter2021spherical} employ spherical CNNs~\cite{cohen2018spherical}, ~\cite{esteves2018learning}, ~\cite{perraudin2019deepsphere} to learn a multi-modal descriptor from spherical and visual images.

In this paper, we opt to project the point cloud onto the sphere, generating the descriptor \sd, for robust rotation alignment. Compared to other methods, we employ a novel sampling method, diverging from the conventional angle-based sampling, which endows our descriptor with isotropy. This characteristic is  crucial for 3-DoF rotation estimation. What's more, with the help of a carefully designed distance calculation function, it can robustly estimate the rotation even in the presence of translational disturbances. 

\subsection{Multi-LiDAR SLAM}
LiDAR SLAM has seen substantial development in recent years. Inspired by LOAM~\cite{LOAM}, numerous variants   improve it from different perspectives, including enhancing accuracy~\cite{wang2021f,dellenbach2022ct, vizzo2023kiss}, feature extraction~\cite{pan2021mulls,shan2018lego}, and feature selection~\cite{duan2022pfilter, jiao2021greedy}. 
Compared to single-LiDAR SLAM or LiDAR-IMU fusion SLAM, multi-LiDAR SLAM is relatively a nascent field. M-LOAM~\cite{mloam} proposes a complete and robust solution for uncertainty-aware multi-LiDAR extrinsic calibration and SLAM.
Lin \etal~\cite{lin2020decentralized} develop a decentralized framework for simultaneous calibration, localization and mapping with multiple LiDARs to remove the single point of failure problem. 
Traj-LO~\cite{zheng2024traj, zheng2024traj2} regards the data of LiDAR as streaming points continuously captured at high frequency and applies the continuous-time trajectory estimation into multi-LiDAR SLAM. Locus 2.0~\cite{reinke2022locus} employs a novel normals-based  G-ICP formulation, an adaptive voxel grid filter  and a sliding-window map approach to reduce memory consumption in multi-LiDAR SLAM. It has been successfully deployed on heterogeneous robotic platforms in the DARPA Subterranean Challenge.

In this paper, multi-LiDAR is not the primary issue to be addressed. However, to achieve precise extrinsic estimation, we implement a simple multi-LiDAR SLAM to provide the poses of each frame and optimize them using \ho. Subsequently, the final calibration results and the time offsets between the LiDARs are outputted through \ts.

\section{Preliminaries}
\label{sec:formulation}
\begin{table}[t]
\renewcommand{\arraystretch}{1.5}
\center
\caption{Nomenclature }

\label{table:notation}
\begin{tabular}{clc}

\hline
Notation & Description                                        & Type            \\ \hline
 $\mathbf{p}$       & Point                                              & $\mathbb{R}^3$              \\ 
$^t\mathcal{S}_a$        & Scan from sensor ``$a$'' at time $t$               & $\{\mathbb{R}^3\}$ \\ 
$^t\mathcal{M}_a$      & Map stacked by scans from ``$a$'' at time $t$                & $\{\mathbb{R}^3\}$  \\ 
$^t\mathbf{P}_a$        & Pose of sensor ``$a$'' at time $t$                   & $SE(3)$           \\ 
$^t_{t+1}\mathbf{T}_a$       & Transformation of ``$a$'' between $t$ and $t+1$ & $SE(3)$          \\ 
$\mathbf{C}$        & Extrinsic parameters between ``$a$'' and ``$b$''        & $SE(3)$           \\ 
$\mathbf{C}^*$        & Result of calibration        & $SE(3)$           \\ 
$\mathbf{C}^{'}$        & Perturbation of calibration, i.e.,  $\mathbf{C}^* = \mathbf{C}^{'} \cdot \mathbf{C}$   & $SE(3)$           \\ 
$\mathbb{F}$        & \sd, descriptor for rotation initializaiton    & $\mathbb{R}^N$           \\ \hline
\end{tabular}
\end{table}

The objective of calibration is to estimate the extrinsic parameters between two independent sensors, which can transform the data from each sensor into the same coordinate system. This is the foundation for many downstream tasks. We refer to one of these sensors as the ``base'' sensor, denoted by $a$, and the other as the ``target'' sensor, denoted by $b$. The related notations are listed in Tab.~\ref{table:notation}, where $SE(3)$ denotes a rigid transformation, which means  $\mathbf{P}, \mathbf{T}, \mathbf{C}$ are composed of a rotation matrix $\mathbf{R} \in SO(3)$ and a translation vector $\mathbf{t} \in \mathbb{R}^3$.
The relationships between the notations are detailed in the following three sections. In the following, we  describe the processes of point cloud registration, LiDAR-based SLAM, and hand-eye calibration with the notations mentioned above.

\subsection{Point Cloud Registration}
The pose transformations between point clouds are typically estimated through point cloud registration. As specified in \cite{registrationsurvey}, given two point clouds $\mathcal{P}_0, \mathcal{P}_1$, the goal of registration is to estimate the transformation matrix $\mathbf{T}^*$ to align $\mathcal{P}_1$ with $\mathcal{P}_0$. The Iterative Closest Point (ICP) algorithm~\cite{icp} is the most common solution to the registration problem, where $\mathbf{T}^*$ can be calculated by

\begin{equation}
\label{eq:registration}
\begin{aligned}
    \mathbf{T}^* &= \mathop{\arg\min}_\mathbf{T} \ \ \|d(\mathcal{P}_0, \mathbf{T}(\mathcal{P}_1))\|^2, \\
\end{aligned}
\end{equation}
where $d(\cdot,\, \cdot)$ denotes the loss function to measure the distance between the two inputs and $\mathbf{T}(\mathcal{P})$ means each point in $\mathcal{P}$ is translated and rotated to form a new point cloud, i.e., $\{\mathbf{R} \mathbf{p} +\mathbf{t} | for~\mathbf{p}~in~\mathcal{P}\}$. 

In this paper, we utilize a more robust point-to-plane ICP algorithm~\cite{pointplaneicp} due to the sparsity of the LiDAR point cloud, which results in the point-to-point ICP lacking sufficient correct correspondences to work properly.
For two frame of point cloud $\mathcal{P}_0, \mathcal{P}_1$, the $d(\mathcal{P}_0, \mathbf{T}(\mathcal{P}_1))$ of point-to-plane ICP is defined as:
\begin{equation}
    d(\mathcal{P}_0, \mathbf{T}(\mathcal{P}_1)) = \sum_ {p \in \mathcal{P}_1} (\mathbf{T}(\mathbf{p}) - p_0) ~\mathbf{n}_0,
\label{eq:pserror}
\end{equation}
where  $p_0$ is any 
point on the corresponding plane and $\mathbf{n}_a$ is the unit normal vector of the plane. The optimal pose estimation can be derived by solving the non-linear equation through Gauss-Newton method~\cite{dennis1996numerical}. The Jacobian matrix can be solved by the chain rule as follows:
\begin{equation}
\begin{aligned}
    \mathbf{J}_d &= \frac{\partial\,  d(\mathcal{P}_0, \mathbf{T}(\mathcal{P}_1))}{\partial \mathbf{T}\mathbf{p}}
    \frac{\partial\, \mathbf{T}\mathbf{p}}{\partial \delta \xi} \\
    &= \mathbf{n}_0 \mathbf{J}_p,
\end{aligned}
\end{equation}
where $\mathbf{J}_p$ can be estimated by applying left perturbation model with $\delta \xi \in \mathfrak{ se}(3)$:
\begin{equation}
\begin{aligned}
    \mathbf{J}_p = \frac{\partial\, \mathbf{T}\mathbf{p}}{\partial \delta \xi} &= \lim_{\delta \xi \rightarrow 0}\frac{exp(\delta \xi)\mathbf{T}\mathbf{p} - \mathbf{T}\mathbf{p}}{\delta \xi}\\
    &=\begin{bmatrix} \mathbf{I}_{3\times3}  & -[\mathbf{T}\mathbf{p}]_\times \\ \mathbf{0}_{1\times3} & \mathbf{0}_{1\times3} \end{bmatrix},
\end{aligned}
\end{equation}
where $[\mathbf{T}\mathbf{p}]_\times$ means the skew symmetric matrix of the transformed point. It is important to note that the accuracy of point cloud registration is often influenced by various factors such as the initial transformation, scene degeneracy, and point cloud bias and noise~\cite{covariance1, covariance2, covariance3}. This paper primarily addresses registration failures caused by the first two factors.

\subsection{SLAM}
In this paper, we also use some concepts commonly used in the SLAM, i.e., simultaneous localization and mapping, which plays a vital role in the navigation of autonomous systems, such as autonomous carscite and robots. The problem that SLAM aims to solve is to estimate the position of a sensor when each frame of data is received. This position is then used to build a map using the sensor data. In specific, for LiDAR-based SLAM, a sequence of point clouds $\{^0\mathcal{S},^1\mathcal{S},\dots,^t\mathcal{S}\}$ is received at consecutive time intervals, and the poses $\mathbf{P}$ at each time is estimated using point cloud registration. The map is then constructed as follow:
\begin{equation}
  ^t\mathcal{M} = \bigcup_{i \in [0, t]}~^i\mathbf{P} ( ^i\mathcal{S} ),  \label{eq:build_map}
\end{equation}
where the pose  is calculated by
$$
^i\mathbf{P} = ^0_i\mathbf{T} = ^0_1\mathbf{T} ~ ^1_2\mathbf{T} \dots  ^{i-1}_i\mathbf{T}.
$$

\subsection{Hand-Eye Method}
Hand-eye calibration is originally a method of calibrating the end-effector of the robot and the camera fixed on it. The position changes $\mathbf{A} \in SE(3)$  of end-effector are obtained through robot kinematics, meanwhile the corresponding position changes $\mathbf{B} \in SE(3)$ of the camera are obtained through pattern pose estimation. 
The transformation  $\mathbf{A}$ and $\mathbf{B}$ conform to the following equation:
\begin{equation}
    \mathbf{A~X}=\mathbf{X~B},
\end{equation}
where the transformation between the end-effector and the camera is denoted as $\mathbf{X}$.

In our calibration task, 
since the base sensor and the target sensor are also rigidly connected, their motion conforms to the following equation as well:
\begin{equation}
    \label{eq:hand_eye_eq}
    ^t_{t+1}\mathbf{T}_a ~ \mathbf{C} =  \mathbf{C}~ ^t_{t+1}\mathbf{T}_b.
\end{equation}

Please note that in the above equation, the $^t_{t+1}\mathbf{T}_a, ^t_{t+1}\mathbf{T}_b$ and $\mathbf{C}$ are all unknown variables, which need to be calculated by the registration mentioned before.
\section{System Overview}
\label{sec:overview}

\begin{figure}[t]
	\centering
	\includegraphics[width = \linewidth]{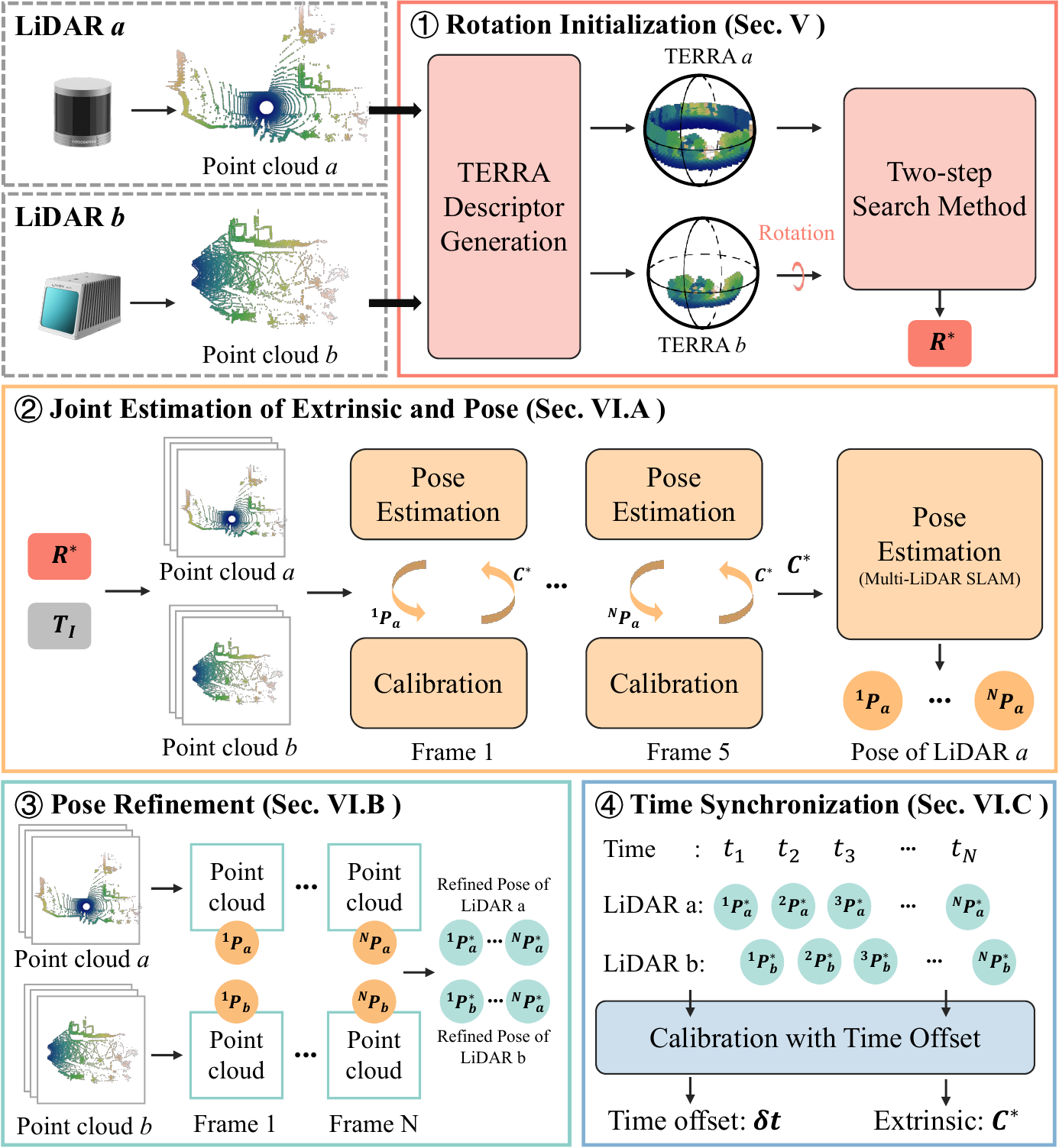}
	\caption{The pipeline of the universal LiDAR calibration involving four core steps, i.e., \sd based rotation initialization, \je for joint estimation of extrinsic and poses,  \ho based pose refinement and \ts for enhancing the accuracy of extrinsic parameters and estimating the time offset.}
	\label{fig:pipeline}
\end{figure}

The objective of this study is to calibrate two LiDARs without any prior knowledge of the initial extrinsic parameters, the point cloud distribution, or the field of view. To address this challenge, we design a universal LiDAR calibration method consisting of four main steps as shown in Fig.~\ref{fig:pipeline}. Firstly, we develop a versatile rotation initialization method based on the proposed \sd which is detailed in Sec.~\ref{sec:ri}. Secondly, building upon the initial rotation, we use \je to jointly estimate the extrinsic parameters and poses with consecutive frames, yielding refined extrinsic parameters. The method is elaborated in Sec.~\ref{sec:je}.
Finally, we implement a hierarchical optimization that not only reduces the calibration errors to the millimeter-level but also achieves time synchronization of the LiDARs, as described in Sec.~\ref{sec:ho} and Sec.~\ref{sec:time_syn}.
\section{Rotation Initialization}
\label{sec:ri}
In this section, we propose a novel method for initializing the rotation part of the extrinsic parameter to ensure the convergence of subsequent steps. It is important to note that, when calibrating sensors on a single mobile platform, i.e., a car or a robot, the translation offset typically does not result in a considerable distance between point pairs which should be matched during registration. This is because the distance between sensors is limited by the size of the mobile platform, and we will conduct experiments to demonstrate it in Sec.~\ref{sec:exp_refine_abla}.
However, the rotation offset can create a significant distance, potentially leading to registration failure. To address this problem, we simplify the initialization problem by focusing solely on estimating the  3-DoF rotation part. We will begin by presenting the newly proposed TERRA, a Three-DoF Embedding of point cloud for Robust Rotation Alignment. Subsequently, we will explain how \sd is utilized to assess rotation and introduce a well-designed method to calculate the distance between two TERRAs to enhance the robustness. Finally, a two-step search method is designed to quickly identify the most suitable initial rotation value. 
For clarity, we employ the term ``descriptor'' interchangeably with ``embedding'' in this paper, following common usage.


\subsection{Design of \sd}

\begin{figure}[t]
	\centering
	\includegraphics[width = \linewidth]{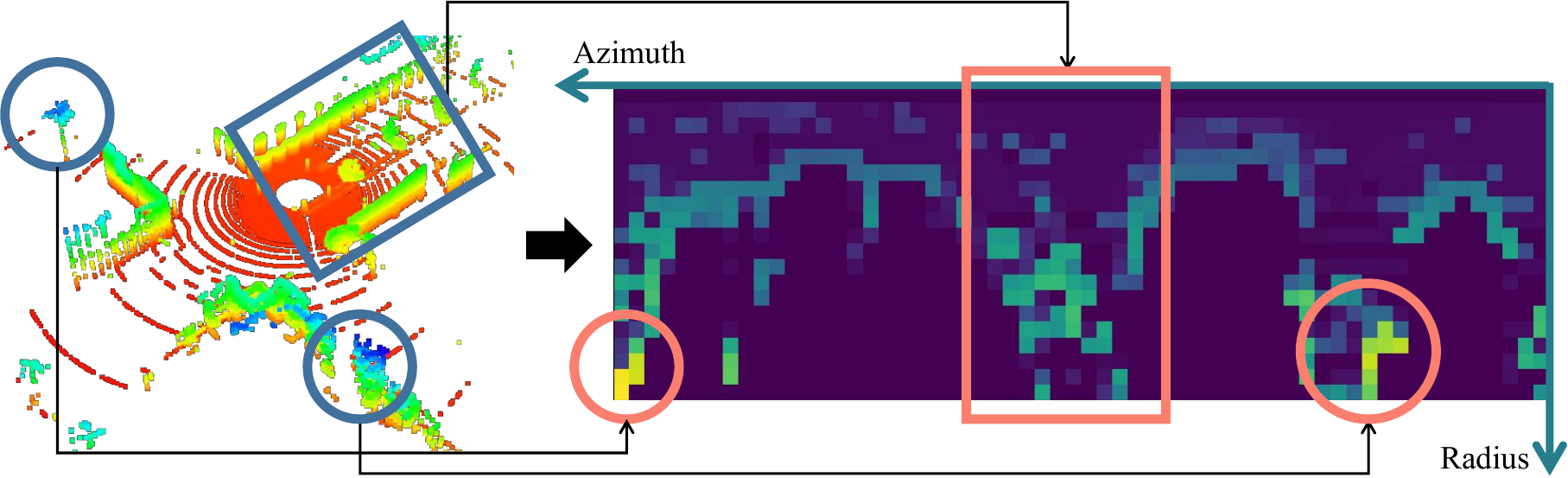}
	\caption{Instructions for the generation of scan context. By projecting the point cloud onto the $x-y$ plane and dividing it into bins based on azimuth and radius, a two-dimensional descriptor is generated. 
	}
	\label{fig:scgeneration}
\end{figure}

Extracting embeddings or descriptors is a common task, often used in place recognition or pose estimation tasks~\cite{meng2024mmplace, chen2021overlapnet, yuan2023std, fu2024crplace}. However, these tasks often require the rotation invariance of the descriptors, in order to improve the algorithm's robustness in dealing with diverse data. 
This requirement conflicts with our need to initialize rotation, which demands accurately reflecting changes in angles.
A classic rotation-aware descriptor is scan context~\cite{scancontext}, which partitions the point cloud into bins on the $x-y$ plane according to both azimuthal and radial directions. Each bin is encoded with the maximum height of the segmented point cloud in the bin, i.e., the max value of $z$. In the end, a matrix is extracted as the descriptor to represent the point cloud as shown in Fig.~\ref{fig:scgeneration} and the rows and columns of the matrix represent different azimuths and radii, respectively. Shifting the columns of the original matrix to create the new matrix represents rotating the point cloud around the $z$ axis, which enables comparing the similarity of descriptors at different angles and thus get the most suitable rotation. However, this method can only search for the optimal angle in 1-DoF, while the 3-DoF estimation is necessary for the task of multi LiDAR calibration.

To solve this problem, inspired by globe in a gimbal, which can rotate on three axes to display the earth, 
we project the point cloud onto a sphere to generate descriptors instead of projecting onto the $x-y$ plane as previously done with the scan context.
The sphere, being a 3D structure, intuitively supports 3-DoF rotation, similar to a gimbal. Additionally, we will not lose any information during the projection, that is, a position on the sphere corresponds to one and only one point in the point cloud. This is because the imaging principle of LiDAR dictates that the laser is unable to penetrate obstacles. 

The next point of consideration is the selection of the most suitable data structure to represent the sphere. For spherical data, a readily conceivable technique is the application of the Mercator projection~\cite{Mercator}, which is a type of cartographic mapping method signifying a sphere's surface into a plane. Upon application of Mercator projection, a 2D matrix can serve as the data structure for \sd.
However, on one hand, this method lacks isotropy, subject to considerable distortion toward the poles, resulting in a misrepresentation of relative size and distance of land masses as seen in world maps. On the other hand, the only reasonable rotation for world maps is around the Earth's axis. In fact, Gauss's Theorema Egregium~\cite{gauss1828disquisitiones} proves that a sphere's surface cannot be represented on a plane without distortion. 
Based on the  discussion, we can summarize that the optimal data structure should encompass two fundamental characteristics: (1) Isotropy, which implies that the descriptor should remain unbiased regardless of the position of the points when encoding them. (2) The capability to effectively perform rotational operations for comparing similarities under different rotations.

\begin{figure}[t]
	\centering
	\includegraphics[width = \linewidth]{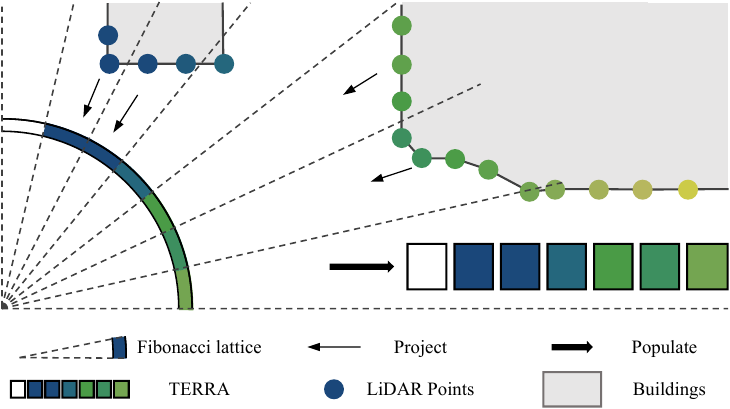}
	\caption{Instructions for the generation of \sd. The process is divided into three steps: sampling the sphere using the Fibonacci lattice, projecting the point cloud from LiDAR onto the  unit sphere, and populating the values into \sd based on the  correspondences.
	}
	\label{fig:sphere_projection}
\end{figure}

To meet the above requirements, we propose TERRA, a Three-DoF Embedding of point cloud for Robust Rotation Alignment.
We conduct three steps to convert a frame of point cloud into the corresponding \sd as shown in Fig.~\ref{fig:sphere_projection}, i.e., spherical sampling, projection, and populating. Specifically, we employ a uniform spherical sampling method known as the Fibonacci lattice~\cite{Fibonacci}. This method uniformly distributes $N$ points across the surface of a sphere with each point approximately covering the same area, which demonstrates the high isotropy of \sd. The Fibonacci lattice sampling approach is detailed in Alg.~\ref{alg:fibonacci}, which takes an integer $N$ as input and returns a spherical point cloud, denoted as $\fl$, consisting of $N$  uniformly distributed points over the sphere. 
\begin{figure}[t]
	\centering
	\includegraphics[width = \linewidth]{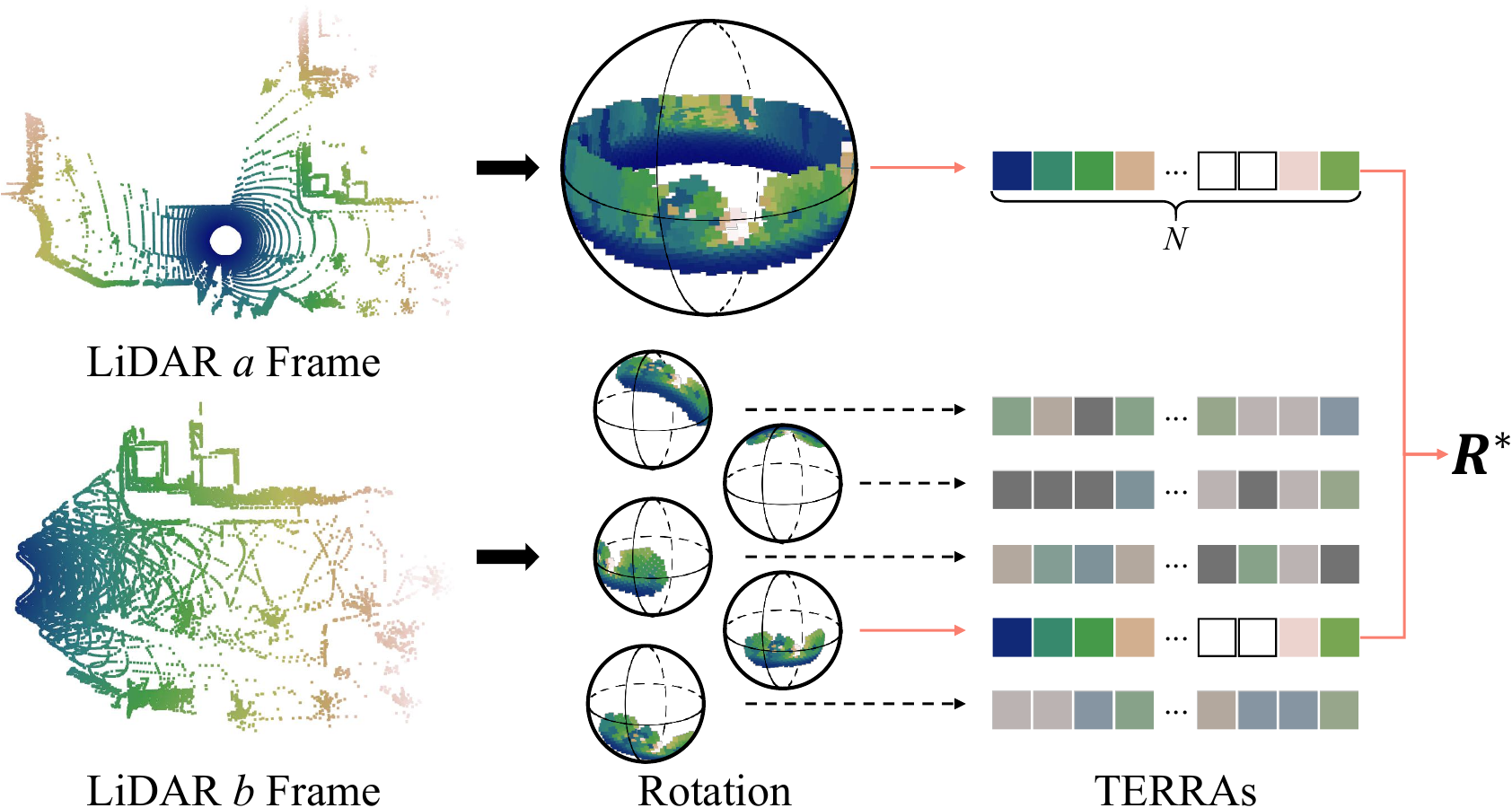}
	\caption{Instructions for rotation initialization. Each LiDAR frame has a \sd extracted from it. Then the descriptor of one LiDAR is fixed,  while rotating the other to find the optimal match. The rotation corresponding to the optimal match is the result of the  rotation initialization. Different colors represent different values and the \sd is a vector of length $N$.
	}
	\label{fig:rotation_init}
\end{figure}

After preparing the point cloud $\fl$, we can encode any given LiDAR frame to \sd, which is a vector of length $N$. In specific, the LiDAR frame is first normalized, i.e., projecting each point onto a unit sphere. Then, the closest point in $\fl$ for each point is efficiently searched by KD-tree~\cite{kdtree}. LiDAR points corresponding to the same point in the $\fl$ are considered the smallest unit to extract \sd, which means that the  ``characteristic'' of these points forms one of the $N$ values in the \sd. In this paper, we use the minimum distance from these covered points to the origin as the ``characteristic'' to populate \sd, aiming to minimize the effects of occlusion.
Finally, a descriptor of length $N$ is generated, as illustrated in the Fig.~\ref{fig:sphere_projection}. 
\begin{algorithm}[t]
\caption{Fibonacci lattice}
\label{alg:fibonacci}
\begin{algorithmic}[1]
\REQUIRE The number of points: $N$
\ENSURE Sphere Points: $\fl$
\STATE $\phi \leftarrow\frac{\sqrt{5} + 1}{2}$ (golden ratio)
\STATE $\fl \leftarrow \emptyset$\
\FOR{$i = 0$ \TO $N-1$}
    \STATE $y \leftarrow 1 - \frac{2i}{N-1} $
    \STATE $radius \leftarrow \sqrt{1 - y^2}$
    \STATE $\theta \leftarrow \frac{2\pi i}{\phi}$
    \STATE $x \leftarrow radius \times \cos(\theta)$
    \STATE $z \leftarrow radius \times \sin(\theta)$
    \STATE Append point $(x, y, z)$ to $\fl$
\ENDFOR
\end{algorithmic}
\end{algorithm}

\subsection{Rotation of the \sd}
After preparing the \sd for each LiDAR frame, we need to design a rotation operation to search for the optimal rotation required. 
For the reason that each value in \sd corresponds one-to-one with points in $\fl$.
Therefore, rotating the point cloud $\fl$ is equivalent to rotating \sd.

Specifically, we rotate the point cloud $\fl$ by a certain rotation $\textbf{R}$, and the resulting rotated point cloud is denoted as $\fl'$. Similarly, KD-tree is employed to establish a one-to-one correspondence between point pairs in the rotated point cloud $\fl'$ and the original $\fl$. In this case, each pair consists of two points with different indices. By taking the value from the original index of \sd and filling it into the corresponding point's index position, we create a rotated \sd. Essentially, we're reorganizing the original \sd values according to the new index alignment post-rotation, thus forming a rotated version of \sd. Fig.~\ref{fig:rotation_init} shows the process of $\fl$ rotation and rotated \sd generation. 

For a descriptor \sd of length $N$, the rotation process entails a time complexity of $O(N)$ for rotating the point cloud $\fl$, and $O(Nlog N)$ for matching points using a KD-tree, followed by an additional $O(N)$ for generating the updated descriptor. Thus, the total time for performing a rotation is $O(Nlog N)$, offering a substantial efficiency over sequentially rotating a LiDAR frame and then generating a new \sd. Additionally, since each point independently searches for its match, the entire rotation process can be further accelerated through parallel computing.

\subsection{Distance between \sd}

We evaluate the distance between TERRAs to identify the most suitable rotation. 
The small distance between the \sd indicates that the landmarks projected to sphere are in the correct corresponding relationship, meaning the correct rotation has been identified.
An effective distance function typically needs to exhibit two key qualities. On one hand, it should show a monotonically increasing or decreasing trend within a local area, thus enabling the use of gradient descent to find the optimal solution. On the other hand, having fewer local optima in the function's domain is beneficial as it facilitates the more efficient identification of the global optimum.

\begin{figure}[t]
	\centering
 \subfigure[The projection positions of different landmarks. The projection of the black dot shows no shift, while the red dot exhibit the maximum shift.] {\includegraphics[width = 0.4\linewidth] {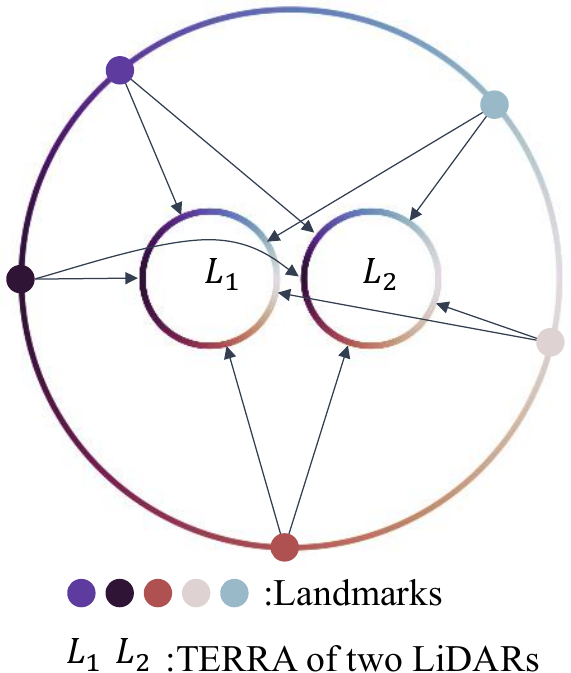} \label{fig:translation_exp1}} 
 \subfigure[Comparing the impact of landmarks at different distances. As the radius of the large circle increases, the color differences gradually decrease while the radius of the smaller circle remains constant.] {\includegraphics[width = 0.55\linewidth]    {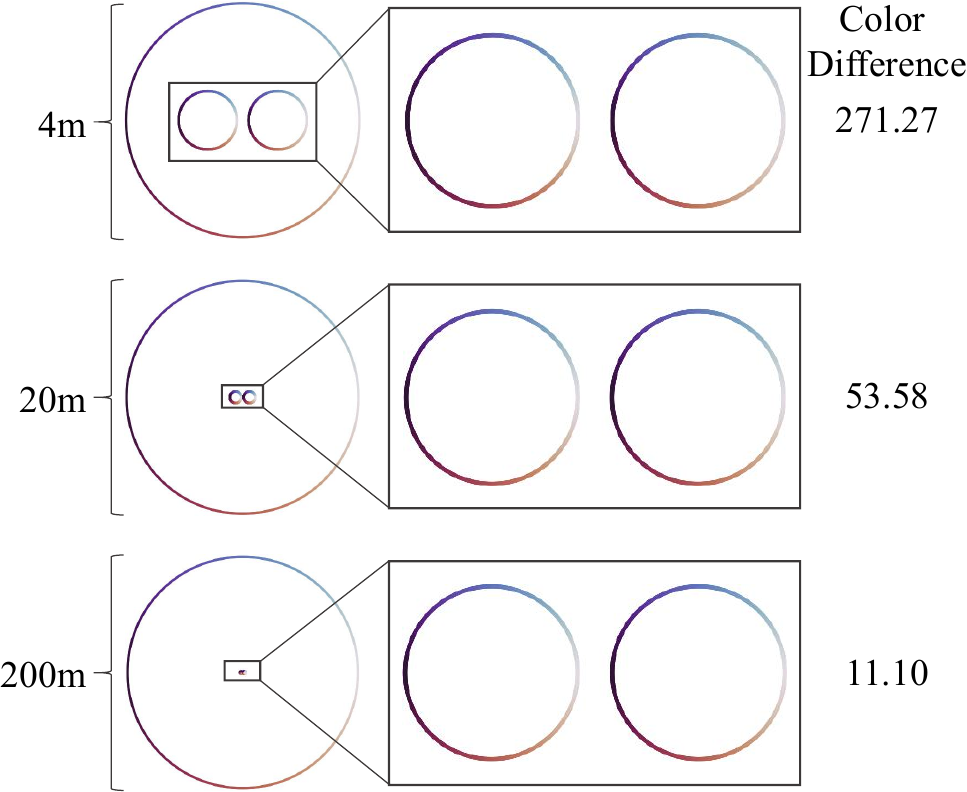} \label{fig:translation_exp2}} 
	\caption{The diagram illustrates the impact of translation between two LiDARs on the projected positions of landmarks. Landmarks are represented by the large circle, while the unit spheres used to generate \sd are depicted as small circles. 
	}
	\label{fig:translation_exp}
\end{figure}

When the origins of two LiDARs coincide perfectly, the angles between landmarks projected onto the two unit spheres remain constant, because the positions of the two unit spheres are identical. In this ideal scenario, 
comparing \sd to estimate rotation is theoretically correct.
However, this assumption is not feasible in reality, as the translation between two LiDARs causes shifts in the projection positions of objects. As illustrated in Fig.~\ref{fig:translation_exp}, we abstract this effect into a task where a big circle (landmarks) colors two small circles (the unit sphere). In Fig.~\ref{fig:translation_exp1}, the black dot, representing landmarks in the same direction as the translation, do not shift in projection position, whereas the red dot, representing landmarks perpendicular to the translation direction, show the maximum shift. 

To address the impact of translation, we draw inspiration from the phenomenon that distant scenery moves slower in the field of view when traveling by train, which indicates that distant landmarks should have smaller shifts in their projection positions. To test this hypothesis, we change the radius of the big circle and quantitatively calculate the coloring differences between the two small circles as shown in Fig.~\ref{fig:translation_exp2}. The results show that as the radius of the great circle increases, the coloring differences decrease.

Based on the observations, we define a mask $\mathbf{M} = \left[ m_i \right]_{i=1}^{N}$ to indicate which positions are suitable for distance calculation for two \sd $\mathbb{F}_1$ and $\mathbb{F}_2$. $m_i$ is defined as:  
\begin{equation}
\label{eq:mask}
     m_i = \begin{cases} 
1, & \text{if } (\mathbb{F}_1^i > d_1) \land (\mathbb{F}_2^i > d_1) \land (|\mathbb{F}_1^i - \mathbb{F}_2^i| \leq d_2) \\
0, & \text{otherwise}
\end{cases},
\end{equation}
where $\mathbb{F}_1^i$ is the $i-th$ position of $\mathbb{F}_1$ and we use the parameter $d_1$ to filter out the distant landmarks from the LiDARs to reduce the interference caused by translation between LiDARs and  $d_1$ is set to $20m$ in the practice.  The parameter $d_2$ is used to  select positions that may belong to the same object for subsequent calculations. $d_2$ can be adjusted according to the actual distance between LiDARs. In our experiments, it is set to $5m$ based on the typical size of vehicles. 

Therefore, we define the distance calculation function as follows:
\begin{equation}
\label{eq:feature_dis}
\begin{aligned}
    D(F_1, F_2) &= \frac{||(\mathbb{F}_1-\mathbb{F}_2)\mathbf{M}^T||_1}{||\mathbf{M}||_1^\alpha}, \\
\end{aligned}
\end{equation}
where $\alpha$ set to 2 is used to increase the influence of valid number on the distance, preventing scenarios with few correspondences but also minimal distances. For instance, even if there's only one valid position with a short distance, the distance remains small.
The smaller distance indicates greater similarity between the two \sd, representing that the rotation is more likely to be close to the ground truth. 

\subsection{Two-step Search Method}
Despite the minor time expenditure for a single rotation, traversing all feasible rotations still demands an unacceptable amount of time, given the vastness of the rotation space. Therefore, we need to devise a search strategy that efficiently identifies the optimal rotation.


In the first step, we start by checking if the initial rotation will result in an overlap between the two point clouds using  Eq.~\eqref{eq:mask}, that is, whether the length of mask $\mathbf{M}$ is not zero. If there is no overlap, the rotation space will be traversed with a significantly large step, e.g., $30^\circ$, until a suitable initial rotation that produces an overlap area is found. 
This step is time-efficient. 
Starting from this initial rotation, we explore the surrounding rotation space with a large step size, e.g., $10^\circ$, sampling among all rotations that lead to an overlap area and scoring each rotation using Eq.~\eqref{eq:feature_dis}. The top ten rotations with the smallest distances will be input into the next stage.

In the second stage, building on the groundwork laid by the first, a heuristic search is employed to find improved solutions in the vicinity of each discrete rotation identified earlier. Specifically, in each area, we explore with a step size of $1^\circ$, moving along the path of decreasing distance as calculated in Eq.~\eqref{eq:feature_dis}, until a local optimum is reached.



\section{Stepwise Refinement with Data Sequences}
\label{sec:reg_datasequences}

After obtaining a appropriate initial rotation value, we first integrate hand-eye calibration, using the initial frames of the sequences to obtain optimized extrinsic parameters. Next, to mitigate the impact of scene degeneracy, we employ a hierarchical optimization approach, continuously optimizing the extrinsic parameters during the multi-LiDAR mapping process. Finally, we provide a method for synchronizing the timing between LiDARs.

\subsection{Joint Estimation of Extrinsic and Pose}
\label{sec:je}
As previously mentioned in Sec.~\ref{sec:formulation}, we model the calibration problem as a point cloud registration problem. Through scan-to-scan registration~\cite{CROON}, we form the most direct calibration method as follows:
\begin{equation}
\mathbf{C}^* = \mathop{\arg\min}_\mathbf{T} \ \ \|d(^t\mathcal{S}_a, \mathbf{T}(^t \mathcal{S}_b))\|^2,\\
\label{eq:calib_scan}
\end{equation}
where each calibration takes the scans $^t\mathcal{S}_a\text{ and }^t\mathcal{S}_b$ from two LiDARs at the same time $t$ as input, and outputs the transformation $\mathbf{C}^*$ between them.

However, since only a single frame of data provides constraints for estimating transformation, the impact of noise and scene degeneracy becomes significant~\cite{mloam}. To mitigate these disadvantages, we propose \je for Joint Estimation of Extrinsic and Pose. By utilizing the motion principle of sensors connected by rigid links, we utilize multiple frames from both LiDARs to refine extrinsic parameters, while estimating the pose at different moments.
\begin{figure}[t]
	\centering
	\includegraphics[width = \linewidth]{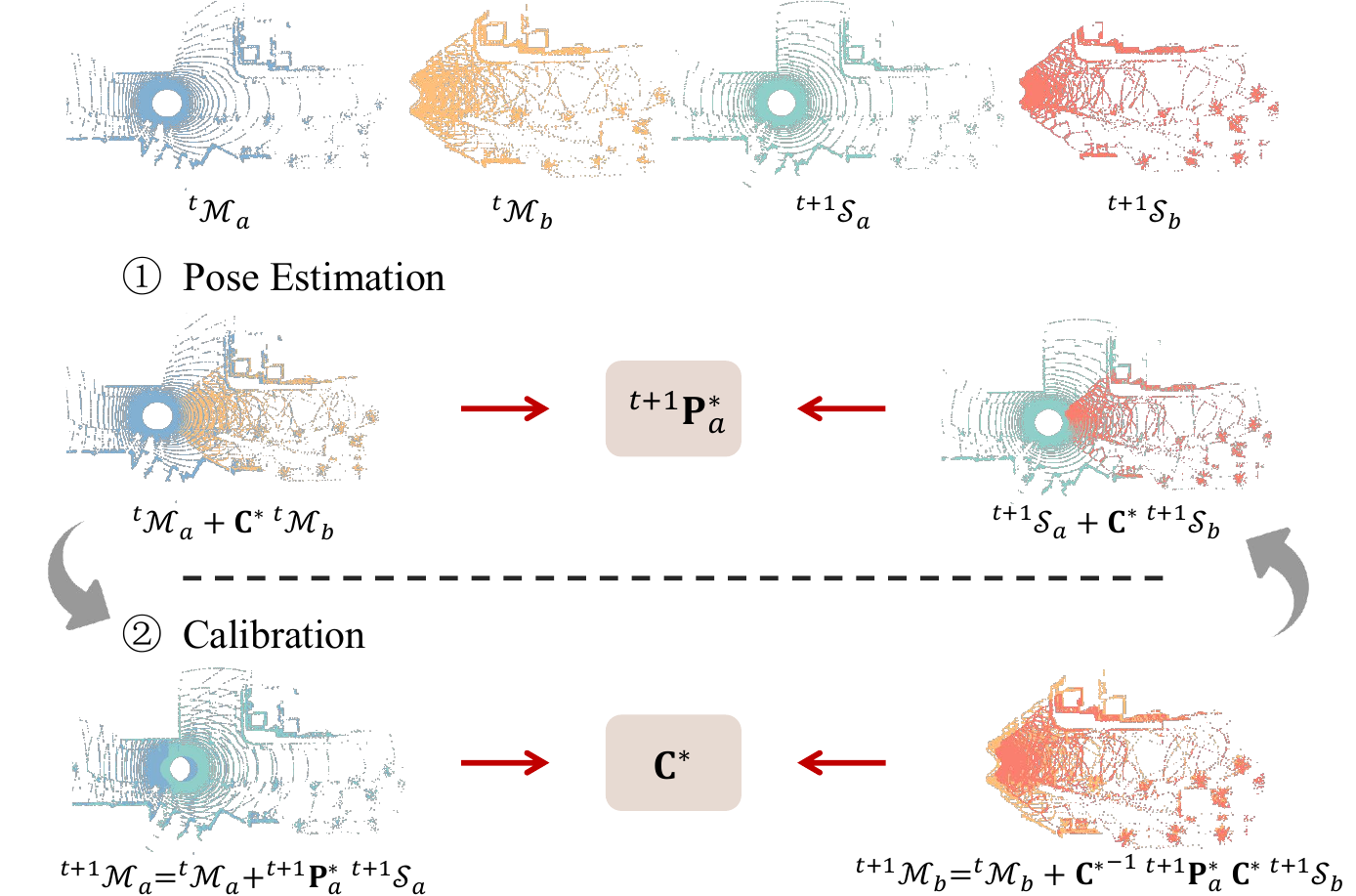}
	\caption{Instructions for \je. For point cloud pairs input at the same time, their transformation relative to the map, i.e., poses, and the transformation between them, i.e., extrinsic, are iteratively solved until either convergence is achieved or a convergence failure is indicated by the number of iterations. 
	}
	\label{fig:refine1}
\end{figure}

In detail, \je is divided into two phases: pose estimation phase and calibration phase as shown in Fig.~\ref{fig:refine1}. 
During the pose estimation phase, we utilize the equation of hand-eye calibration~Eq.~\eqref{eq:hand_eye_eq}, combining the extrinsic parameters and the motion of the base LiDAR $a$ to derive the motion of the target LiDAR $b$. This means that we can define the pose estimation problem as:
\begin{equation}
\begin{aligned}
\label{eq:je_odom}
   ^t\mathbf{P}^*_a = &\mathop{\arg\min}_\mathbf{T} (\ \ \|d(^{t-1}\mathcal{M}_a, \mathbf{T}(^t \mathcal{S}_a))\|^2 + \\
   &\|d(^{t-1}\mathcal{M}_b, (^{t-1}\mathbf{C}^{*-1} ~ \mathbf{T} ~ ^{t-1}\mathbf{C}^*)(^t \mathcal{S}_b))\|^2), \\
\end{aligned}
\end{equation}
where  $^{t-1}\mathcal{M}_a\text{ and }^{t-1}\mathcal{M}_b$ are the maps built according to Eq.~\eqref{eq:build_map}, and the pose of LiDAR $b$ $^t\mathbf{P}^*_b$ is replaced by $^{t-1}\mathbf{C}^{*-1}~^t\mathbf{P}^*_a~ ^{t-1}\mathbf{C}^*$. 
At startup, the map uses the first frame as a substitute, i.e., $^{0}\mathcal{M}_a = ^{0}\mathcal{S}_a$ and $^{0}\mathcal{M}_b = ^{0}\mathcal{S}_b$. The rotation part of $^0C^*$ is initialized using \sd, and the translation part is set to $(0, 0, 0)$.
The Jacobian matrix for the first part is presented in Sec.~\ref{sec:formulation}, and the second part is as follows:
\begin{equation}
\begin{aligned}
    \mathbf{J}_p &= \frac{\partial\, \mathbf{C}^{-1}\mathbf{T}\mathbf{C}\mathbf{p}}{\partial \delta \xi} = \lim_{\delta \xi \rightarrow 0}\frac{\mathbf{C}^{-1}exp(\delta \xi)\mathbf{T}\mathbf{C}\mathbf{p} - \mathbf{C}^{-1}\mathbf{T}\mathbf{C}\mathbf{p}}{\delta \xi}\\
    &=\begin{bmatrix}\mathbf{R}^{\mathbf{C}^{-1}}_{3\times3}\\\mathbf{0}_{1\times3}\end{bmatrix}\begin{bmatrix} \mathbf{I}_{3\times3}  & -[\mathbf{R}^\mathbf{T}\mathbf{R}^\mathbf{C}  \mathbf{p} + \mathbf{R}^\mathbf{T} \mathbf{t}^\mathbf{C} + \mathbf{t}^\mathbf{T}]_\times \\ \mathbf{0}_{1\times3} & \mathbf{0}_{1\times3} \end{bmatrix},
\end{aligned}
\end{equation}
where $\mathbf{R}^C$ is the rotation of  $^{t-1}\mathbf{C}$,  and $\mathbf{R}^\mathbf{T}$, $\mathbf{t}^\mathbf{T}$ are the rotation and translation of  $\mathbf{T}$.

In the calibration phase, we employ the following objective function to align two maps:
\begin{equation}
\label{eq:calib_map}
^t\mathbf{C}^* = \mathop{\arg\min}_\mathbf{T} \ \ \|d(^t\mathcal{M}_a, \mathbf{T}(^t \mathcal{M}_b))\|^2.\\    
\end{equation}
We iteratively execute these two processes until both $^t\mathbf{C}^*$ and $^t\mathbf{P}^*_a$ stabilize and converge. During this process, when $^t \mathbf{C}^*$ is inaccurate, estimating $^t \mathbf{P}^*_a$ is difficult to converge due to the incorrect position of $^t \mathcal{S}_b$. Additionally, using a local map for the calibration introduces more constraints, making the point cloud registration in Eq.~\ref{eq:calib_map} more accurate than in  Eq.~\ref{eq:calib_scan}.  Exceeding a certain number of iterations will be considered a failure of \je.

However, repetitively executing this procedure for every frame yields little benefit and is impractical. 
On one hand, the iterative computation of each frame cannot meet real-time requirements, and the continuous growth in map size will further intensify this issue. On the other hand, due to the use of a map-to-map registration for calibration as specified in Eq.~\eqref{eq:calib_map}, the cumulative error in pose estimation inevitably leads to map deformation, which negatively impacts the accuracy of the calibration.
Therefore, we restrict its execution to the initial phase of the data sequence as illustrated in Fig.~\ref{fig:pipeline}, where it proves to be cost-effective and highly accurate. After \je, we develop a straightforward multi-LiDAR SLAM method using only the pose estimation phase to provide poses for subsequent steps.

\subsection{Pose Refinement by Hierarchical Optimization}
\label{sec:ho}
\begin{figure}[t]
	\centering
	\includegraphics[width = \linewidth]{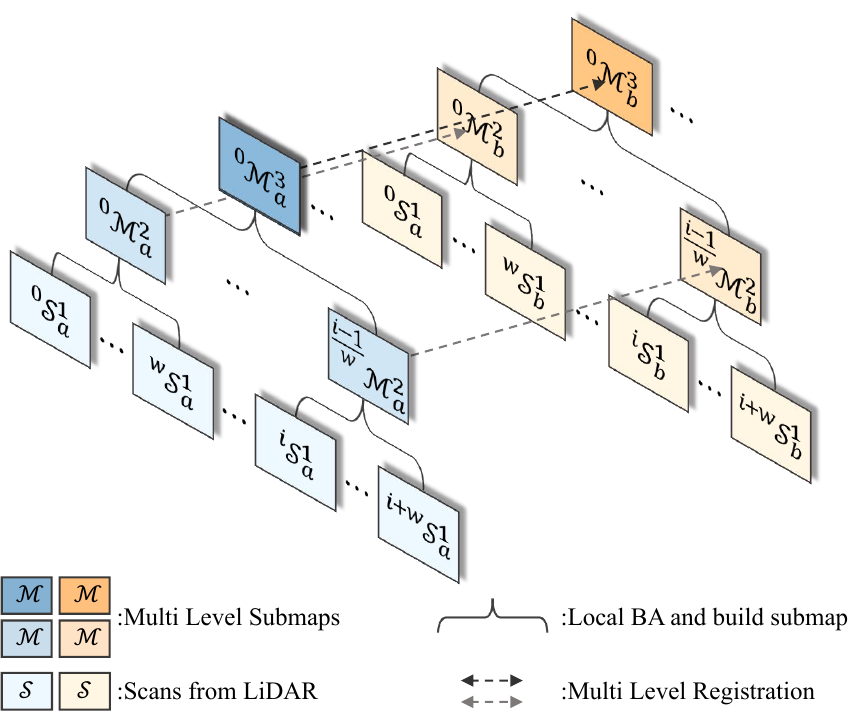}
	\caption{Hierarchical optimization  for pose refinement. Each LiDAR constructs local maps at different levels and optimizes the pose through local LiDAR BA with in the same window. }
	\label{fig:hba}
\end{figure}
Despite using multi-frame data and incorporating motion constraints in \je, calibrating in a single environment remains risky due to the impact of scene degeneracy on point cloud registration accuracy~\cite{zhang2016degeneracy, mloam}. Collecting data across different environments by SLAM to jointly produce extrinsic parameters is advisable. However, errors in motion estimation can lead to incorrect calibration results, as analyzed in the previous section. Therefore, we employ a hierarchical optimization module to optimize the poses generated by multi-LiDAR SLAM, aiming to reduce pose errors and use all data collectively to estimate the extrinsics.

In the field of SLAM, a common method used for backend optimization involves using factor graphs to optimize all variables together, achieving global consistency. However, this approach typically optimizes only at the level of poses, overlooking the constraints introduced by the point cloud itself. Inspired by HBA~\cite{hba}, we employ a similar hierarchical optimization architecture here to achieve more accurate poses for each data frame from both LiDARs. 
Specifically,  as shown in Fig.~\ref{fig:hba}, 
we first construct a hierarchical structure for each LiDAR's data by building submaps at various levels. For each LiDAR, scans are grouped according to a window size $w$. A local LiDAR bundle adjustment (BA)~\cite{liu2021balm} is performed in each local window using the initial pose estimated by the multi-LiDAR SLAM, optimizing the relative pose between each frame and the first frame in this window. Scans within the same window with the poses optimized 
by BA, form a local submap which serves as a frame in the next layer. This process can be repeatedly performed from the lower layer to the upper layer. Subsequently, the hierarchical structures of the two LiDARs are linked through the registration of different layer submaps.
Finally, the poses of all frames and submaps are linked into a factor graph and solved using the Levenberg-Marquardt method with GTSAM~\cite{gtsam}.

In addition, since the BA process within each submap is independent, it is suitable to use parallel computing to accelerate the process. To achieve that, we maintain a thread pool and a task queue. When each submap needs to be stitched, the task is added to the task queue and the idle thread will retrieve tasks from the task queue. 
\subsection{Time Synchronization}
\begin{figure}[t]
	\centering
	\includegraphics[width = \linewidth]{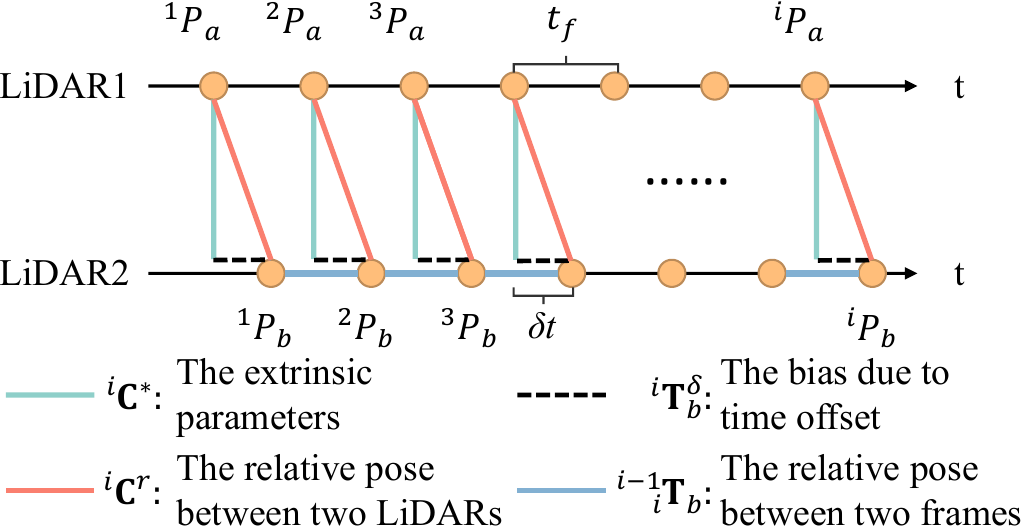}
	\caption{The estimated extrinsic is disturbed by motion due to time offset. We identify the optimal extrinsics by determining the time offset.}
	\label{fig:time_syn}
\end{figure}
\label{sec:time_syn}
After \je and \ho, each LiDAR frame's pose is as accurate as possible. However, as illustrated in Fig.~\ref{fig:time_syn}, in the absence of time synchronization for the LiDAR pairs, the pose between two aligned frames from two LiDARs, denoted as $^i \mathbf{C}^r= ^i \mathbf{P}_a^{-1}~^i\mathbf{P}_b$  here, is not the actual extrinsic parameters, denoted as $^i \mathbf{C}^*$, between the LiDARs.
Therefore, we propose a method for time synchronization to estimate the time offset between different LiDARs.

It's worth noting that we assume a consistent frame rate for the LiDAR, meaning the time offset is a constant value. Moreover, between adjacent frames from the same LiDAR, the motion status remains consistent, allowing us to obtain the position at any given moment through interpolation.

More specifically, we hypothesize the time offset between two LiDARs as $\delta t$ and denote the interval of LiDAR capturing data as $t_f$. The ratio between $\delta t$ and $t_f$ is denoted as $\alpha=\frac{\delta t}{t_f}$. The pose bias caused by the time offset is denoted as:
\begin{equation}
    ^i \mathbf{T}_b^\delta=\alpha~^{i-1} \mathbf{P}_{b}^{-1}~ ^i \mathbf{P}_b,\\
\end{equation}
where $\alpha$ means interpolating the pose. 
The translation part is linear interpolation under constant velocity assumption, while the rotation part is spherical linear interpolation (Slerp)~\cite{dam1998quaternions}. 
Following that, the triangular relationship in the Fig.~\ref{fig:time_syn} can be expressed as:
\begin{equation}
  ^i \mathbf{C}^*~ ^i \mathbf{T}_b^\delta = ^i \mathbf{C}^r.
  \label{eq:relationship}
\end{equation}
Substituting $^i \mathbf{T}_b^\delta$ and $\mathbf{C}^r_i$ into Eq.~\ref{eq:relationship}, we get
\begin{equation}
    ^i \mathbf{C}^* = ^i \mathbf{P}_a^{-1}~^i\mathbf{P}_b (\alpha~^{i-1} \mathbf{P}_{b}^{-1}~ ^i \mathbf{P}_b)^{-1},
\end{equation}
where $^i \mathbf{C}^*$ is a function of $\alpha$. We model the process of finding $\mathbf{C}^*$ and $\alpha$ as minimizing the variance of $\mathbf{C}^*$, as $\mathbf{C}^*$ is considered constant over short periods once the time offset interference is excluded. The calculation process is as follows:
\begin{align}
    \alpha ^* & = \mathop{\arg\min}_\alpha Var(\mathbb{C}^*|\alpha), \label{eq:var}\\
    \mathbf{C} ^* & = Mean( \mathbb{C}^*|\alpha ^*),\label{eq:mean}
\end{align}
where $\mathbb{C}^* = \{^1 \mathbf{C}^*, ^2 \mathbf{C}^*, \cdots, ^i \mathbf{C}^*\}$, and the $\mathbf{C}^*$ is the final output of the whole algorithm. 

\section{Experiments}
\label{sec:exp}
To fully verify the effectiveness of our method, we conduct comprehensive experiments using both simulated and real-world datasets to evaluate the performance of the proposed methods compared to existing approaches. We first introduce the datasets used in our experiments in Sec.~\ref{sec:dataset}. In Sec.~\ref{sec:Metrics}, we describe the quantitative evaluation metrics used in the subsequent experiments. 
In Sec.~\ref{sec:initexp}, the comparison between \sd and five common methods demonstrates that our method provides significant robustness for diverse LiDAR pairs. Quantitative experiments are conducted on both simulated and real-world datasets, along with essential ablation studies. 
In Sec.~\ref{sec:refineexp}, we test our extrinsic refinement methods \je, the \ho module and the \ts module against two open-source calibration methods as well as the direct scan-to-scan method. Since the last two modules need to work together, we  abbreviate them as \hots.
Please note that all registration processes are implemented using the widely adopted point-to-plane ICP method~\cite{pointplaneicp}.  The parameters and their values used in our method are as shown in the Tab.~\ref{table:parameters}. Except for the ablation studies, all experiments are conducted according to this setup.
\begin{table}[t]
\center
\caption{Parameters and Their Values Used in Our Method}
\label{table:parameters}
\renewcommand{\arraystretch}{1.5}
\begin{tabular}{c|lc}
Module                 & \multicolumn{1}{c}{Description} &  Value  \\ \hline
\multirow{5}{*}{\sd}   &  $N$, sphere sampling number    & 10000       \\
                       & $d_1$, used in distance calculation  & 20                                  \\
                       & $d_2$, used in distance calculation  & 5 \\      
                       &   First-stage search stride         & 10 \\  
                       &   Second-stage search stride         & 1 \\ \hline  
\multirow{2}{*}{\je}  &  Input frame number to \je & 5                   \\
                       &  Failure judgment threshold    & 10                                   \\ \hline
\multirow{2}{*}{\makecell{Hierarchical \\ optimization}}   & $w$, submap window size         &   5    \\
               & Layer number    & 3      
\end{tabular}
\renewcommand{\arraystretch}{1}

\end{table}
\subsection{Datasets}

\label{sec:dataset} 
\begin{figure*}[ht]
	\centering
	\subfigure[COLORFUL] {\includegraphics[width = 0.24\linewidth]{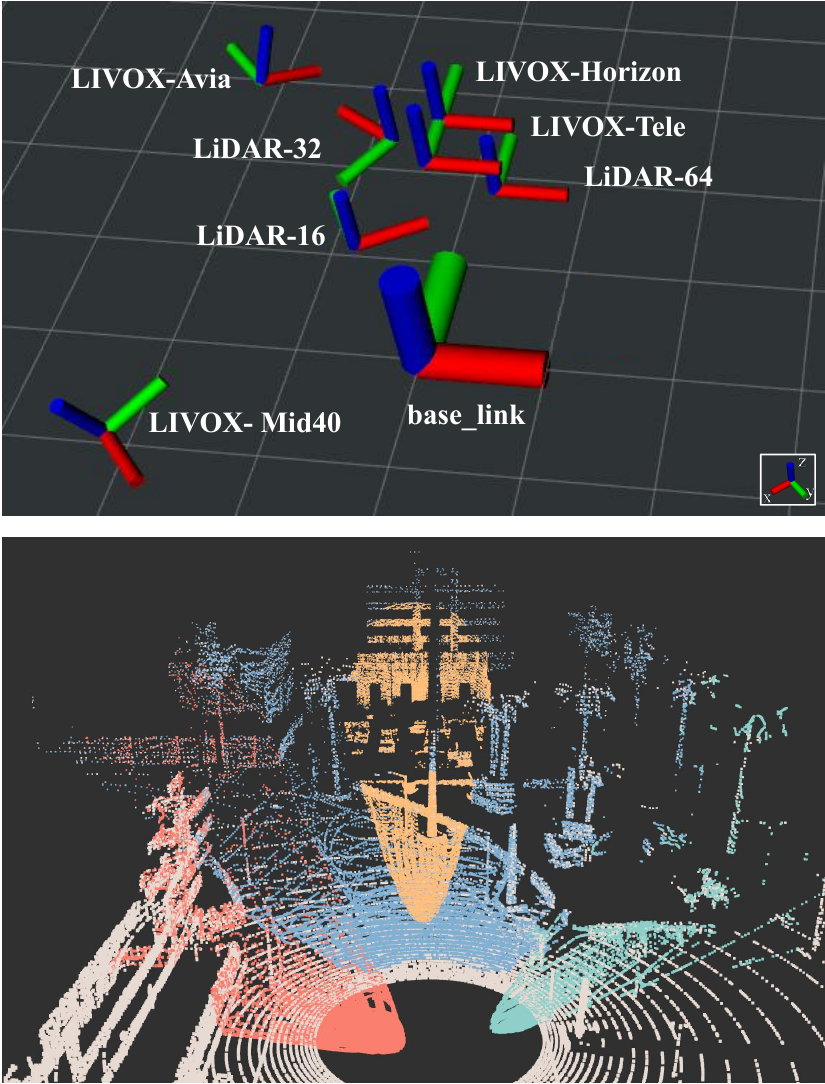}\label{fig:dataset_a}}
	\subfigure[Campus-SS] {\includegraphics[width = 0.24\linewidth]{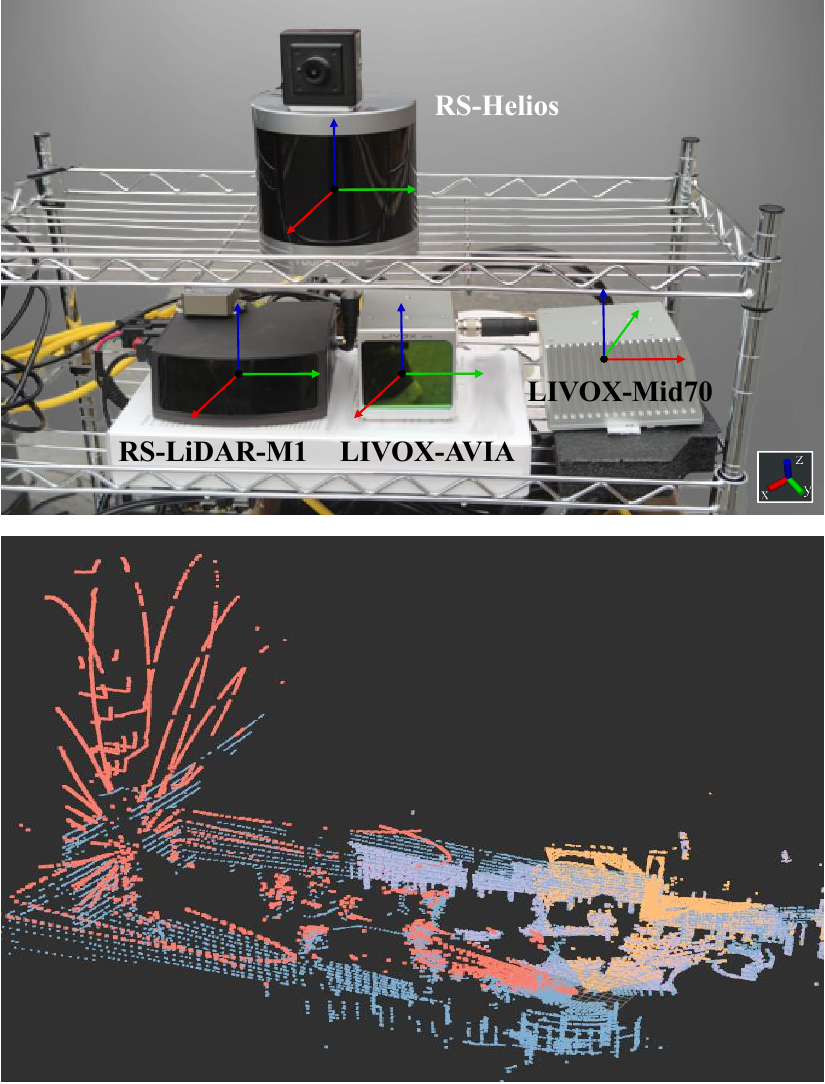}\label{fig:dataset_c}}
	\subfigure[Campus-BS] {\includegraphics[width = 0.24\linewidth]{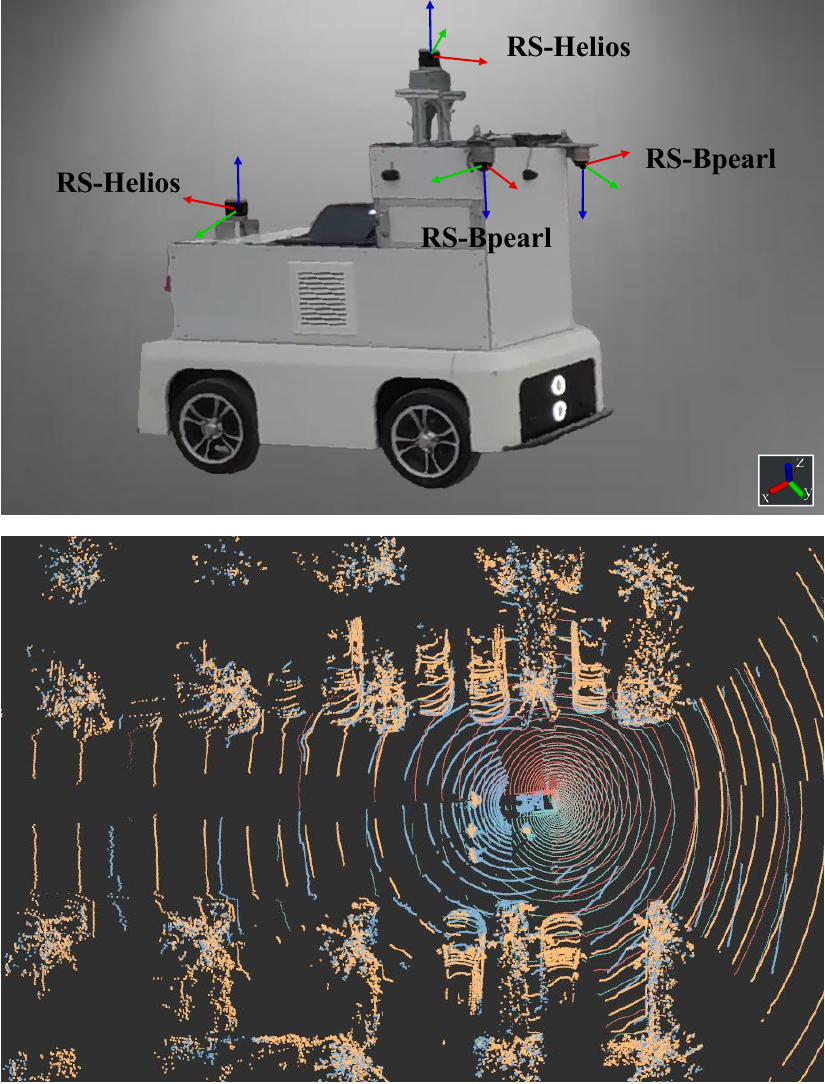}\label{fig:dataset_d}}
    \subfigure[USTC FLICAR] {\includegraphics[width = 0.24\linewidth]{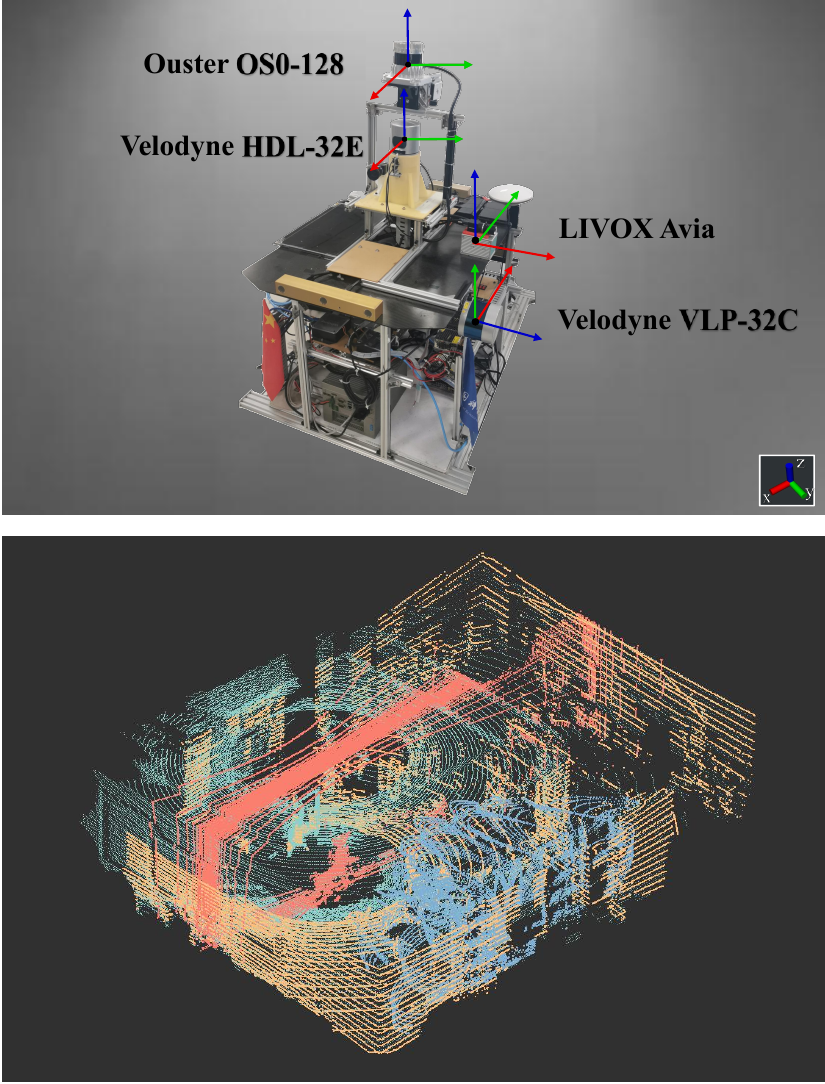}\label{fig:dataset_f}}\\
    \subfigure[TIERS] {\includegraphics[width = 0.24\linewidth]{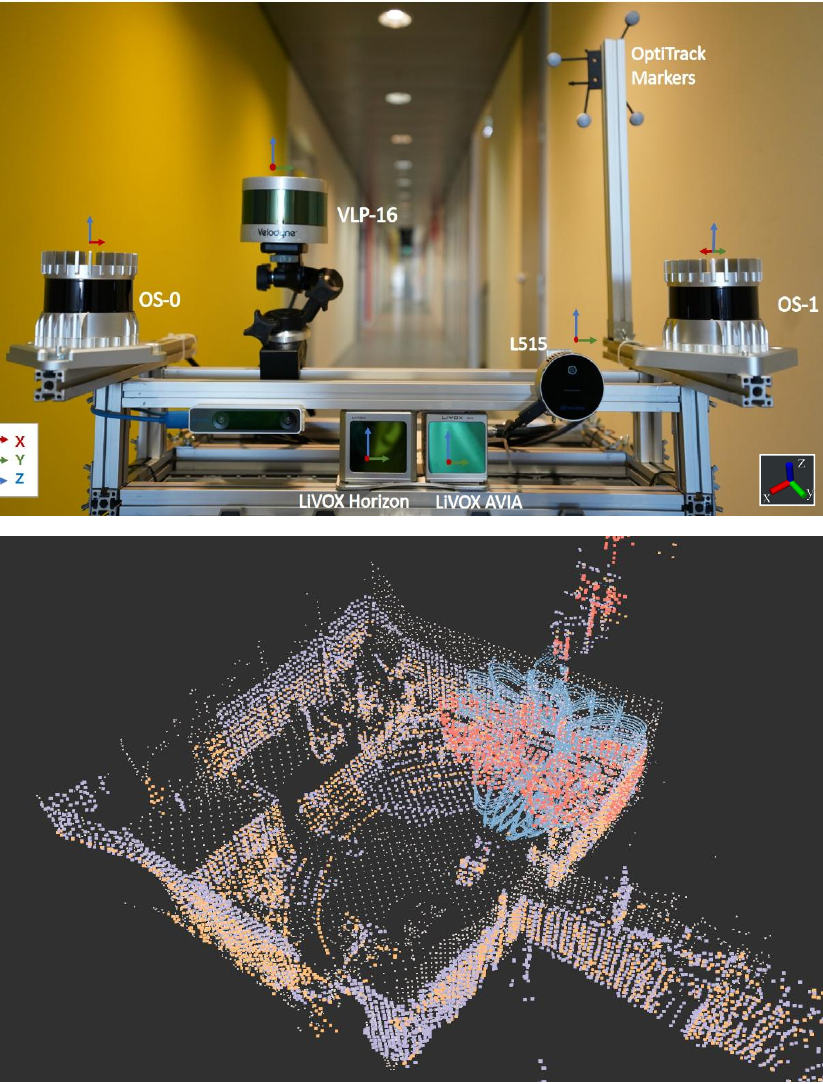}\label{fig:dataset_b}}
    \subfigure[PandaSet] {\includegraphics[width = 0.24\linewidth]{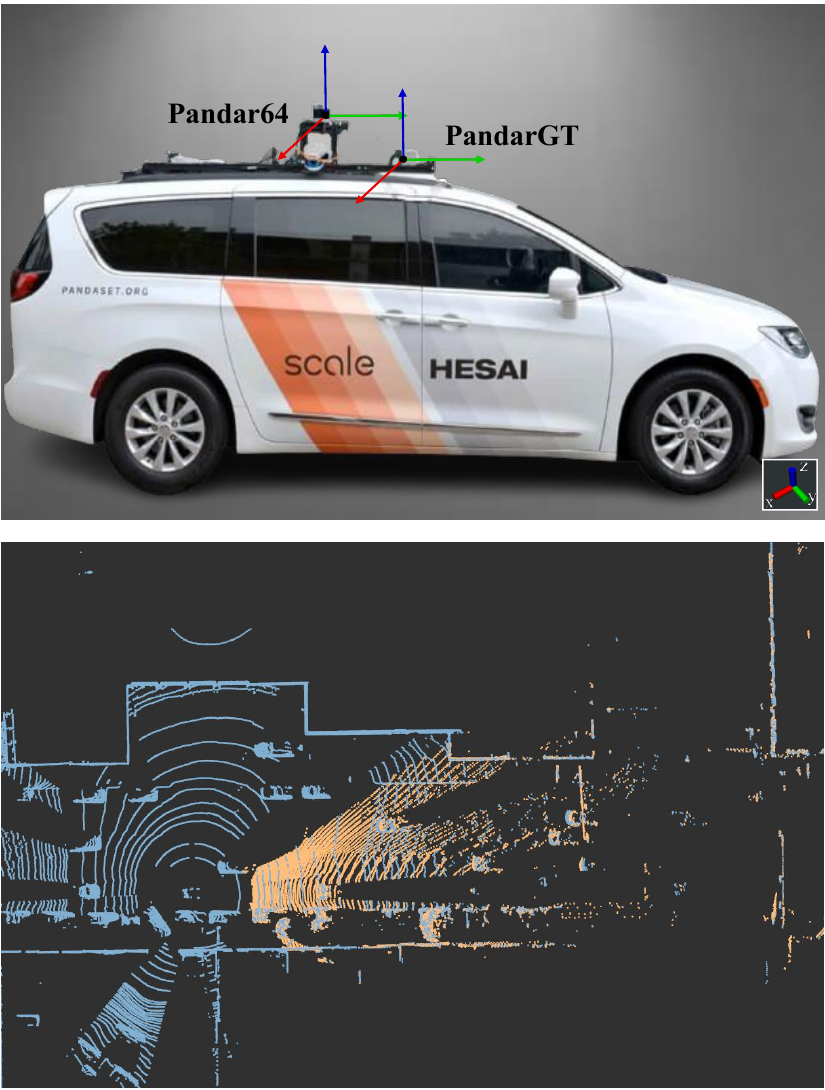}\label{fig:dataset_e}}
    \subfigure[A2D2] {\includegraphics[width = 0.24\linewidth]{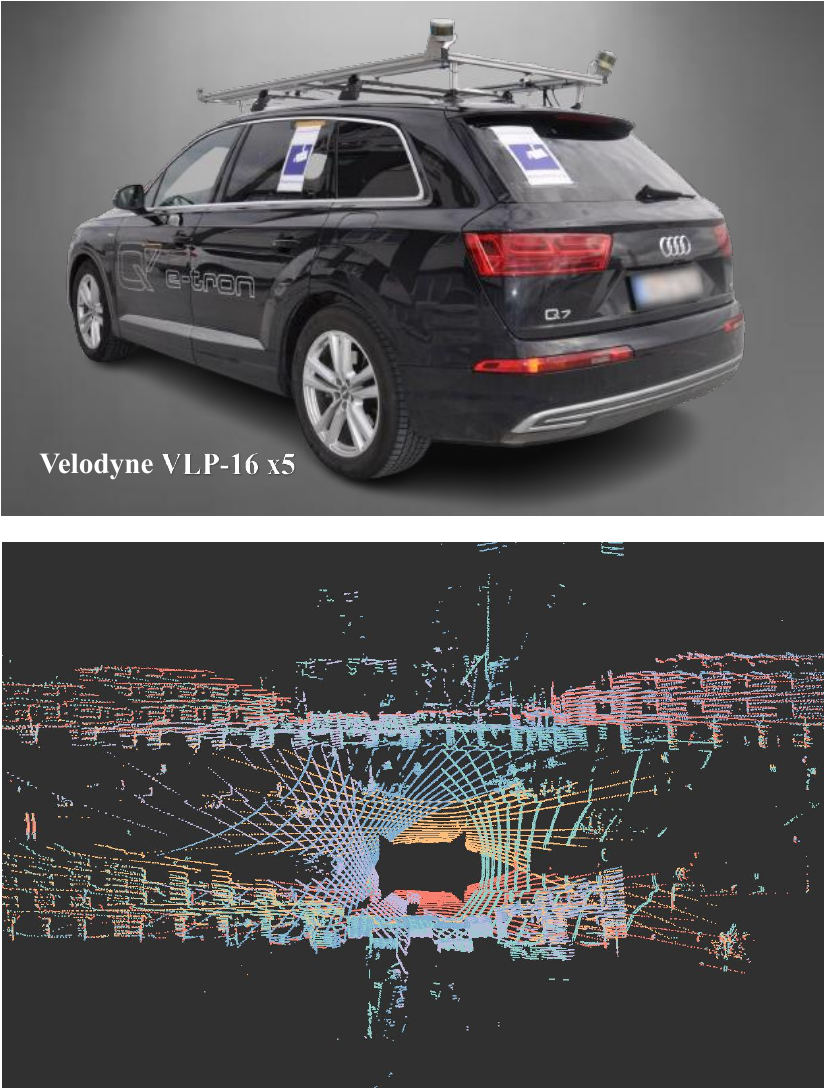}\label{fig:dataset_g}}
    \subfigure[UrbanNav] {\includegraphics[width = 0.24\linewidth]{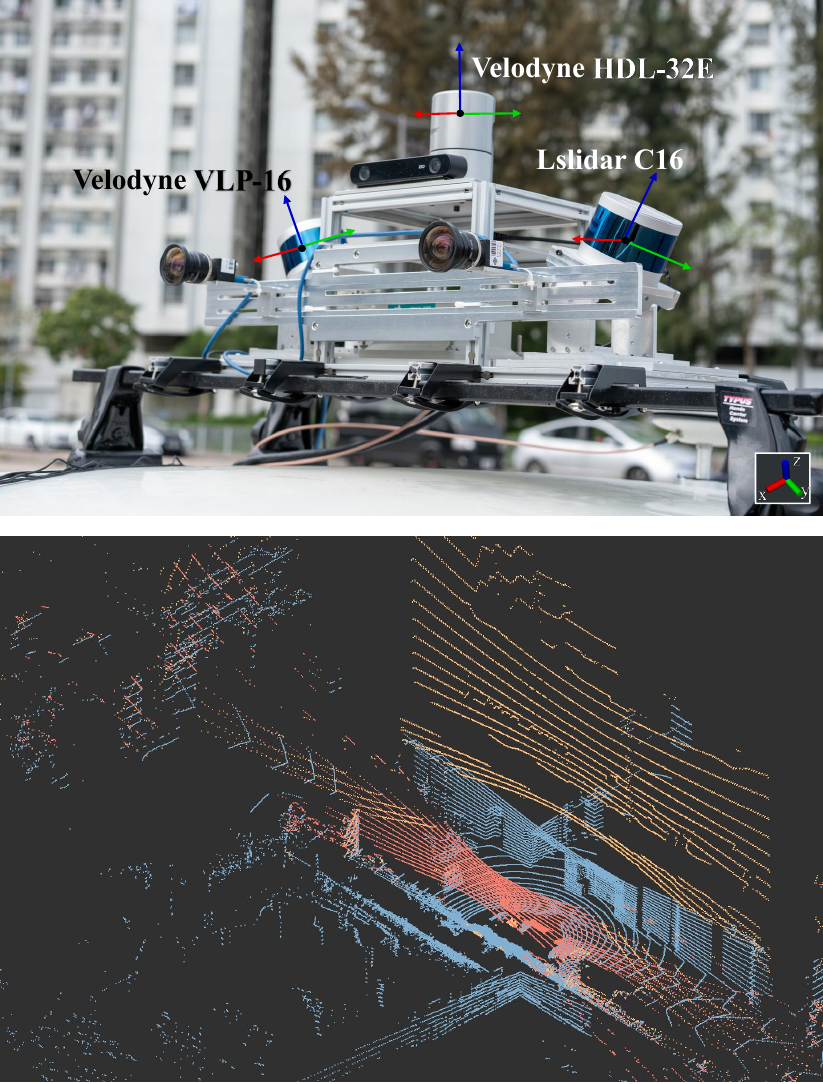}\label{fig:dataset_h}}
	\caption{The platform and point clouds of utilized datasets. Three datasets are self-collected, five are publicly available; one is a simulated dataset, and seven are real-world datasets. These encompass 16 diverse types of LiDAR. For each dataset, the upper figure represents the actual position of the LiDARs, and the lower one illustrates the calibrated point clouds from LiDARs.}
	\label{fig:dataset}
\end{figure*}
\begin{table*}[ht]
\center
\caption{Datasets for Evaluating the Proposed Method.  16 Different Types of LiDAR are Involved.}
\label{table:dataset}
\renewcommand{\arraystretch}{1 }
\begin{tabular}{c|ccccccc|c|c}
\hline
\multicolumn{1}{c|}{\multirow{2}{*}{Dataset}} & \multicolumn{7}{c|}{LiDAR}                                                                                                    & \multirow{2}{*}{Type} & \multirow{2}{*}{Self-Collected} \\ \cline{2-8}
\multicolumn{1}{c|}{}                         & \multicolumn{3}{c|}{Spinning}                                     & \multicolumn{4}{c|}{Solid-State}                          &                        &                                 \\ \hline
COLORFUL                                       & \multicolumn{3}{c|}{LiDAR-\{16, 32, 64\}}                         & \multicolumn{4}{c|}{Livox-\{Horizon, Mid40, Avia, Tele\}} & Simulated              & \ding{51}  \\ 
Campus-SS                                        & \multicolumn{3}{c|}{RS-Helios}                                    & \multicolumn{4}{c|}{RS-LiDAR-M1, Livox-\{Mid70, Avia\}}   & Realistic              &  \ding{51}   \\ 
Campus-BS                                       & \multicolumn{3}{c|}{RS-\{Bpearl$\times$2, Helios$\times$2\}}                 & \multicolumn{4}{c|}{/}                                     & Realistic              &  \ding{51}      \\ 
USTC FLICAR~\cite{ustcfly}                                    & \multicolumn{3}{c|}{Ouster OS0-128, Velodyne-\{VLP-32C,HDL-32E\}} & \multicolumn{4}{c|}{Livox Avia}                           & Realistic              &   \ding{55}\\ 
TIERS~\cite{tiers}                                       & \multicolumn{3}{c|}{VLP-16, Ouster-\{OS1-64, OS0-128\}}           & \multicolumn{4}{c|}{Livox-\{Horizon, Avia\}}              & Realistic              &     \ding{55}  \\ 
PandaSet ~\cite{xiao2021pandaset}                                      & \multicolumn{3}{c|}{Pandar64}                                     & \multicolumn{4}{c|}{PandarGT}                             & Realistic              &    \ding{55}\\ 

A2D2~\cite{geyer2020a2d2}                                           & \multicolumn{3}{c|}{VLP-16$\times$5}                                   & \multicolumn{4}{c|}{/}                                     & Realistic              &  \ding{55}      \\ 
UrbanNav~\cite{hsu2021urbannav}                                       & \multicolumn{3}{c|}{Velodyne-\{VLP16, HDL-32E\}, Lslidar C16}     & \multicolumn{4}{c|}{/}                                     & Realistic              &    \ding{55}\\ \hline

\end{tabular}
\renewcommand{\arraystretch}{1}
\end{table*}


To demonstrate the versatility of our method, a variety of real-world and simulated datasets are utilized, encompassing 16 diverse types of LiDARs. The datasets used in our experiments are listed in Tab.~\ref{table:dataset}. The calibrated point clouds and the platforms are shown in Fig.~\ref{fig:dataset}.
A brief description of each dataset is provided below:
\begin{itemize}
\item \textbf{COLORFUL}: We collected a multi-LiDAR dataset in  CARLA~\cite{carla}, an open urban driving simulator. 
This dataset is captured using three mechanical LiDARs and four different Livox LiDARs\footnote{https://www.livoxtech.com/}, each displaying data in various colors as illustrated in Fig.~\ref{fig:dataset_a}, hence the name ``colorful''. The dataset covers a duration of 120 seconds, with each sensor operating at a rate of 20 Hz. 
The simulated environment provides accurate relative extrinsic parameters among the sensors, which serve as the ground truth for the calibration task.
Therefore, the quantitative experiments described below are primarily performed using this dataset. 

\item \textbf{Campus-SS}  We collected this dataset using a low-speed vehicle carrying 4 LiDARs on the campus of USTC. This dataset is used to investigate the performance of algorithms in calibrating various solid-state LiDARs, such as Livox models and the RS-LiDAR-M1\footnote{https://www.robosense.ai/en/rslidar/RS-LiDAR-M1}. For this reason, we have named it Campus-SS. The ground truth for this dataset is established by manually providing initial values for point cloud registration, performing the calculation five times, and taking the average. The calibration scenes are manually selected to be suitable for calibration tasks, ensuring sufficient constraints in all directions.

\item \textbf{Campus-BS} We collected the dataset on an autonomous logistic vehicle on  USTC campus, equipped with two 32-beam LiDARs (RS-Helios\footnote{https://www.robosense.ai/en/rslidar/RS-Helios})
and two special LiDARs for near-field blind-spots detection (RS-Bpearl\footnote{https://www.robosense.ai/en/rslidar/RS-Bpearl}). 
We use this dataset to investigate the performance related to blind-spots LiDARs. For this reason, we have named it Campus-BS. The method for obtaining the ground truth is the same as Campus-SS dataset. 

\item \textbf{USTC FLICAR}~\cite{ustcfly}: 
This dataset, collected by Wang \etal, used an aerial platform on a bucket truck carrying four LiDARs for aerial work. It's important to highlight that this dataset features two vertically positioned LiDARs with a reduced overlapping FoV, introducing challenges for calibration. Meanwhile, this dataset features movement trajectories distinct from those found in autonomous driving datasets, including vertical movements.
\item \textbf{TIERS}~\cite{tiers}: The TIERS dataset, collected by Li \etal, encompasses data captured by different types of LiDARs, including spinning LiDARs with varying resolutions (16, 64, and 128 beams) and solid-state LiDARs with two different scan patterns. The dataset offers a diverse collection of environments, including indoor, outdoor, urban roads, forests, and various other scenes.  

\item \textbf{PandaSet~\cite{xiao2021pandaset}}: This dataset was collected by Xiao \etal, equipped with one mechanical spinning LiDAR (Pandar64) and one forward-facing LiDAR (PandarGT) for autonomous driving scenarios. 

\item \textbf{A2D2~\cite{geyer2020a2d2}}: The A2D2 dataset was collected by Geyer \etal~with five 16-beam LiDARs on highways, country roads, and cities in the south of Germany.  This dataset represents a practical vehicle sensor layout strategy, utilizing five LiDARs to provide comprehensive surround sensor coverage.

\item \textbf{UrbanNav}~\cite{hsu2021urbannav}: The UrbanNav dataset, collected by Hsu \etal~in Hong Kong with 3 mechanical LiDARs, focuses on urban canyon scenarios. In our experiments, it is primarily used to test the performance of online calibration in these typical urban driving situations.
\end{itemize}
\vspace{-1em}

\subsection{Metrics}
\label{sec:Metrics}
\subsubsection{Accuracy}
To measure the difference between the calibration results and the ground truth, we define two metrics to quantify the errors in the rotation and translation components.

The rotation error is calculated using the following formula with the reference to \cite{roterror, rotationeverage}:
\begin{equation}
e_{rot} = \arccos((trace(\textbf{R}_{gt}^{-1} \textbf{R}_{est}) - 1 )/2),
\end{equation}
where $\textbf{R}_{est}$ and $\textbf{R}_{gt}$ respectively represent the estimated and ground truth values of the rotation.
This formula calculates the angle between two rotation matrices in radians. The smaller the angle, the more accurate the estimation of rotation is.

The translation error is calculated using the formula that calculates the Euclidean distance in meters between two sensor positions, as follows:
\begin{equation}
e_{trans} = \|\textbf{t}_{est} - \textbf{t}_{gt}\|_2,
\end{equation}
where $\textbf{t}_{est}$ and $\textbf{t}_{gt}$ respectively represent the estimated and ground truth values of the translation.
The smaller the distance, the more accurate the estimation of translation is.
\subsubsection{Success rate}

When calibration failures, the rotation and translation errors can be significantly large, indicating only the failure of this particular instance, not the accuracy of the algorithm, so averaging all errors is not an effective method for comprehensively evaluating the calibration algorithm.
For better evaluation, 
we define ``calibration success'' to filter samples for computing accuracy and use success rate, abbreviated as $SR$, to measure the robustness of algorithms.
For rotation initialization, we define success as the rotation errors less than $10^\circ$, which is. For the final outcome, success is defined as rotation errors less than $1^\circ$ and translation errors less than $10~cm$. The $SR$ is calculated as the number of successful samples divided by the total number of samples.

\subsection{Study of Rotation Initialization}
\label{sec:initexp}

\begin{figure*}[t]
	\centering
	\includegraphics[width = .9\linewidth]{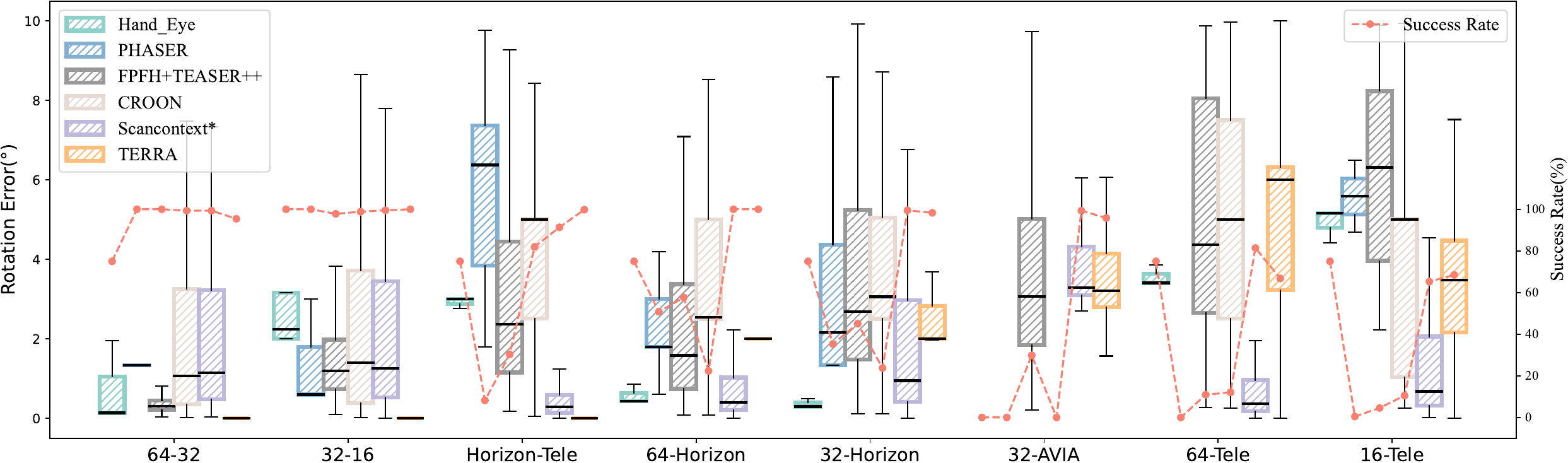}
	\caption{The rotation error and success rate of rotation initialization tasks.  We select eight representative LiDAR pairs from the COLORFUL dataset for calibration and test six methods. The x-axis of the graph represents the LiDAR types involved in the eight calibration tasks. The box plot indicates rotation error, while the line graph represents the success rate. 
	}
	\label{fig:init_result}
\end{figure*}


In this section, we conduct experiments to evaluate the accuracy of the proposed \sd for rotation initialization. We compare its performance with several commonly used methods, including the motion-based Hand-Eye method~\cite{mloam}, the feature-based FPFH method with Teaser++~\cite{fpfh, yang2020teaser}, the geometric-based CROON method~\cite{CROON}, and  the PHASER method~\cite{bernreiter2021phaser}, which also project the LiDAR frame onto the sphere for global registration. In addition, we enhance the classic scancontext~\cite{scancontext}. 
Specifically, we extract and align the largest planes considered as the ground from the LiDAR scans first, to estimate the roll and pitch angles. 
The detail of plane alignment can be found in~\cite{CROON}. Subsequently, we leverage scancontext~\cite{scancontext} to calculate the yaw, thereby accomplishing the 3-DoF rotation initialization. This method is denoted as scancontext* for comparison.

We conduct comprehensive tests on the proposed rotation initialization method, e.g., \sd, starting with quantitative assessments of eight various LiDAR pairs within COLORFUL dataset, and comparing the results with five commonly used algorithms. Subsequently, the method is tested across various real-world datasets, encompassing different scenes and sensor types, demonstrating the versatility of \sd. Finally, we perform ablation studies on some of the key parameters. 
\subsubsection{Quantitative experiment in the simulated dataset}
Given that the COLORFUL dataset encompasses the most diverse sensor types and precise ground truth, we select eight typical calibration tasks to test \sd.   The difficulty of the tasks increases from left to right in Fig.~\ref{fig:init_result}.

The experimental results demonstrate that our method achieves the highest accuracy among the six methods tested. The Hand-Eye method requires sufficient motion, which cannot be satisfied in our randomly picked sequences, leading to the algorithm failing to converge. The FPFH-based method fails in almost all tasks due to the different distributions of point clouds in different LiDARs, which makes it difficult to use the same set of parameters to extract similar features locally. For the same reasons, PHASER performs poorly in calibrations related to solid-state LiDARs. Similarly, the geometry-based CROON method performs well in mechanical-mechanical and solid-solid initialization but fails in cross-modal tasks. The scancontext* achieves a precision comparable to \sd, but it requires a substantial ground area in the LiDARs' FoV to align roll and pitch. This condition is satisfied in the sensor layout of the COLORFUL dataset but may not be fulfilled in all actual scenarios, as demonstrated in the subsequent section.

\begin{figure}[t]
	\centering
	\subfigure[64-32] {\includegraphics[width = 0.45\linewidth]{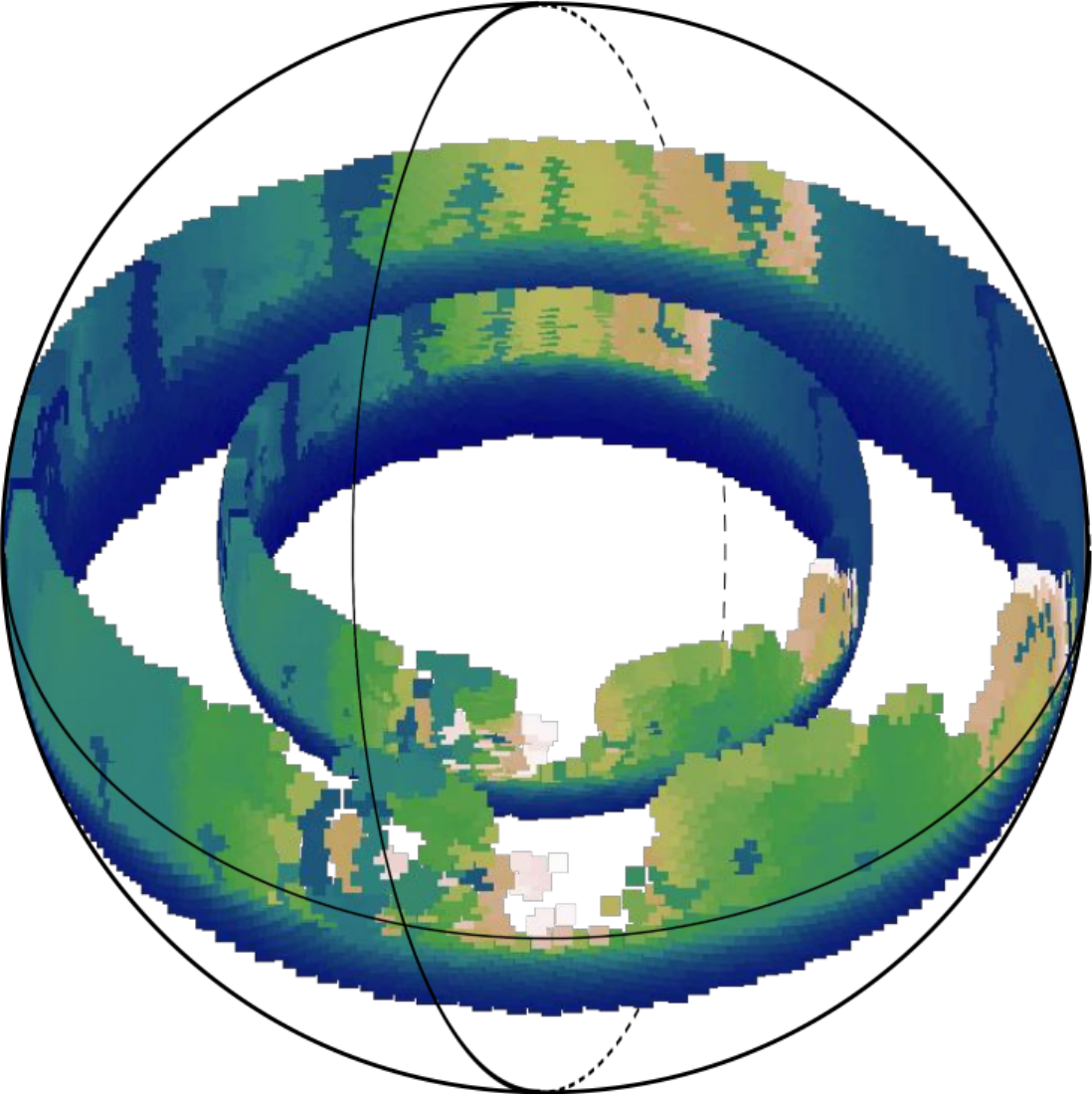}}
	\subfigure[Horizon-Tele] {\includegraphics[width = 0.45\linewidth]{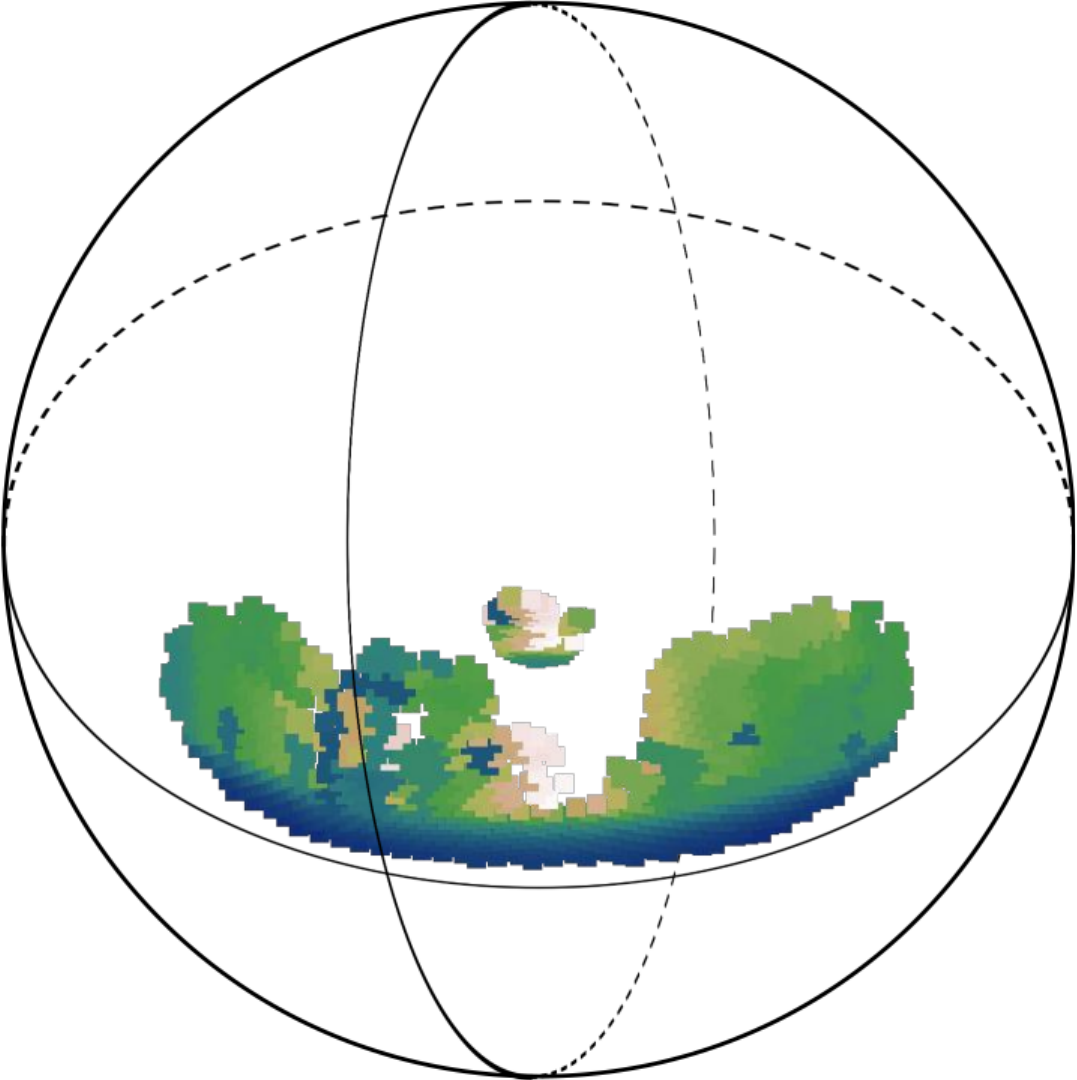}}
	\subfigure[32-Mid40] {\includegraphics[width = 0.45\linewidth]{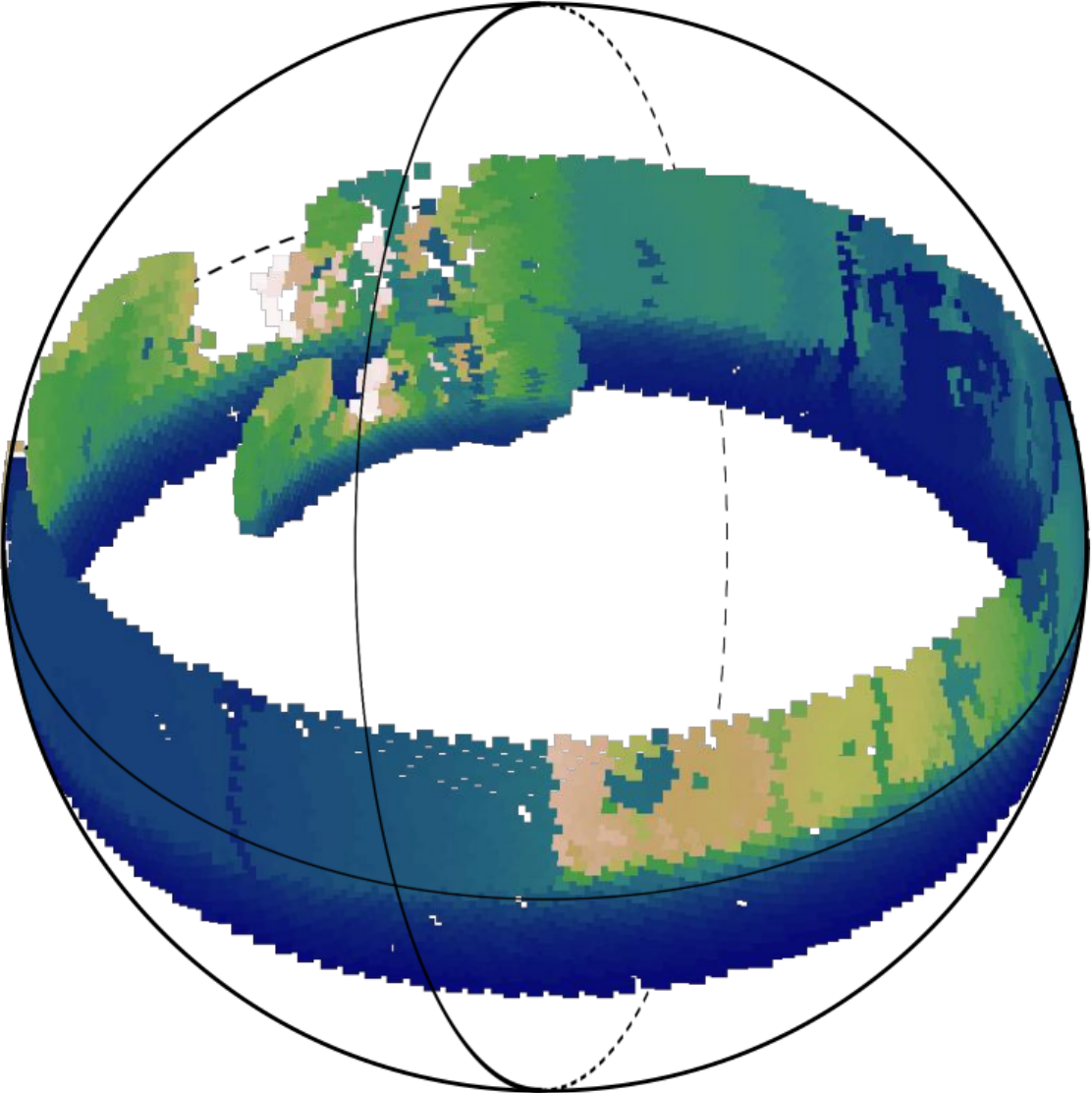}}
	\subfigure[16-Tele] {\includegraphics[width = 0.45\linewidth]{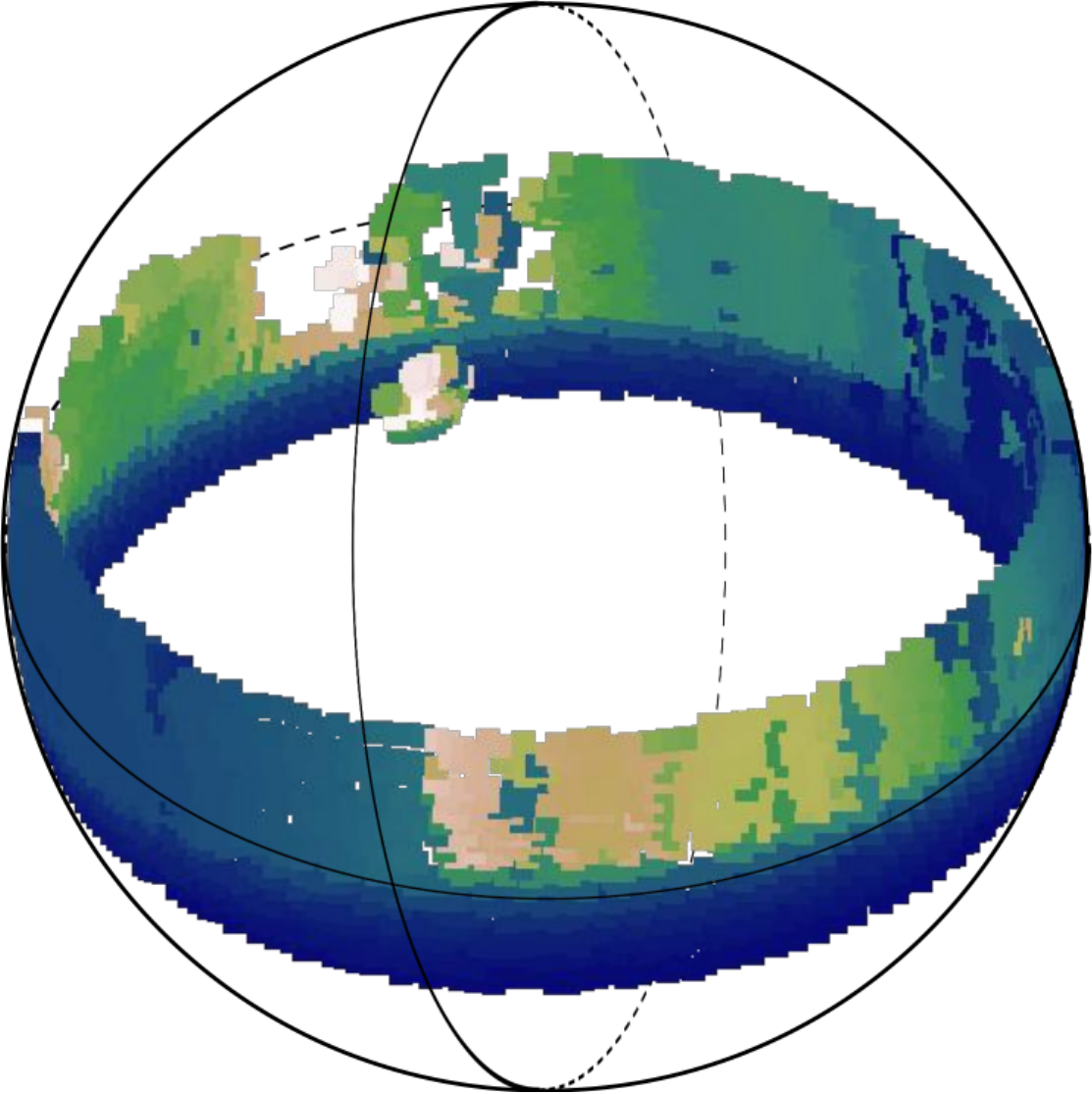}}
	\caption{The example of the \sd in COLORFUL dataset. The \sd for the first LiDAR is represented by the outer sphere, while the second
is illustrated by the inner sphere.}
	\label{fig:example_of_sd}
\end{figure}

In contrast, the proposed \sd demonstrates robustness across all calibration tasks. For simple tasks, the success rate approaches $100\%$, and even in the two most challenging tasks, the success rate exceeds $60\%$. 
In terms of accuracy, for the left four tasks, the variances of \sd's results is very small, indicating the  stability of \sd across different environments. Notably, for the 64-Horizon calibration task, although the variance is small, the mean error is not near zero. This phenomenon is caused by the search step and \sd's resolution. We also consider this an excellent completion of the initialization task, as this error can be further reduced through subsequent refinement step.
Overall, our experimental results demonstrate the superiority of our \sd method, which can provide accurate and reliable rotation initialization for a wide range of LiDAR pairs. Part of the schematic diagram for \sd is shown in Fig.~\ref{fig:example_of_sd}. For clearer viewing, 
the \sd for the first LiDAR is represented by the outer sphere, while the second is illustrated by the inner sphere.

\begin{table}[t]
\center
\resizebox{\linewidth}{!}{
\begin{threeparttable}
\caption{The rotation initial result of \sd in the real-world dataset.}
\label{table:init_real_table}
\scriptsize
\renewcommand{\arraystretch}{1.5}

\begin{tabular}{c|c|cccc}
\hline
\multirow{2}{*}{Dataset} & \multirow{2}{*}{Calibration Target} & \multicolumn{2}{c}{Scancontext*}               & \multicolumn{2}{c}{\sd} \\ \cline{3-6} 
                         &                             & \multicolumn{1}{c}{$e_{rot}$($^\circ$)\ $\downarrow$} & $SR$(\%)\ $\uparrow$ & \multicolumn{1}{c}{$e_{rot}$($^\circ$)\ $\downarrow$} & $SR$(\%)\ $\uparrow$\\ \hline
Campus1                  & \makecell{ RS-LiDAR-M1\\  LIVOX-AVIA}         & \multicolumn{1}{c}{/ }         &   0     & \multicolumn{1}{c}{1.144}      &  100   \\ \hline
\multirow{2}{*}{Campus2}  & \makecell{ RS-Helios(top)\\  RS-Helios(back)}          & \multicolumn{1}{c}{7.491}          &  46.46 & \multicolumn{1}{c}{1.521}          &  100  \\ \cline{2-6} 
                          & \makecell{ RS-Helios(top)\\ RS-Bpearl(right)} & \multicolumn{1}{c}{/}          &    0 & \multicolumn{1}{c}{2.243}          &  100     \\ \hline
\makecell{ USTC\\ FLICAR} & \makecell{ Ouster OS0-128\\ VLP-32C}               & \multicolumn{1}{c}{/}          &    0    & \multicolumn{1}{c}{3.178}          &  63.54  \\ \hline
\multirow{3}{*}{TIERS}   & \makecell{ LIVOX Horizon\\  LIVOX AVIA }                      & \multicolumn{1}{c}{2.643}          & 66.67   & \multicolumn{1}{c}{2.018}          &  100   \\ \cline{2-6} 
                         & \makecell{   OS-1 \\  LIVOX AVIA }                  & \multicolumn{1}{c}{4.84}          &  21.73  & \multicolumn{1}{c}{1.918}          &  100   \\ \cline{2-6} 
                         & \makecell{  OS-0 \\  OS-1  }                & \multicolumn{1}{c}{4.008}          &   67.57  & \multicolumn{1}{c}{0.5819}          &  100   \\ \hline
Pandaset                 & \makecell{  Pandar64\\  PandarGT }                        & \multicolumn{1}{c}{5.399}          &  92.40  & \multicolumn{1}{c}{1.438}          & 93.67 \\ \hline
A2D2                     & \makecell{  frontcenter\\  frontleft}       & \multicolumn{1}{c}{/}          & 0  & \multicolumn{1}{c}{8.726}          & 61.37  \\ \hline
UrbanNav                 & \makecell{  VLP16\\   Lslidar C16}                  & \multicolumn{1}{c}{/}          &  0 & \multicolumn{1}{c}{3.785}          & 67.78  \\ \hline
\end{tabular}
\renewcommand{\arraystretch}{1}
\begin{tablenotes}
    \item \tiny * For A2D2 dataset, the ``frontcenter'' and ``frontleft'' is the positions of the LiDARs, which are both VLP-16.
\end{tablenotes}
\end{threeparttable}}
\end{table}

\subsubsection{Quantitative experiment in the real world}
\label{sec:exp_init_realworld}
To verify the algorithm's performance in real-world scenarios, we select ten representative LiDAR pairs from seven datasets as shown in Tab.~\ref{table:init_real_table}. In this section, we compare our method with the best-performing method, Scancontext*, in last part. 

Due to the variety of installation methods of LiDARs, the assumption of a large planar surface in LiDAR scan does not hold, resulting in the complete failure of Scancontext* in some scenarios, with a success rate of zero.
In contrast, the proposed \sd still achieves stable and successful rotational initialization across most of scenarios without making any assumptions about the environment.
In most tasks, the rotation error is within 5 cm. In the A2D2 dataset, due to severe distortions of the point cloud caused by motion, the error is relatively larger, but the success rate still exceeds $60\%$. What's more,
it's worth noting that \sd requires the input of two point clouds to have overlapping areas, but this does not mean that only LiDARs with overlapping areas can be calibrated. Similar to the approach~\cite{small_fov} designed for non-overlapping LiDAR calibration, if a map is constructed for each LiDAR and then the overlapping local maps are input into the \sd, it can still work effectively.

\begin{figure}[t]
	\centering 
	\subfigure[Campus-BS (top-right)] {\includegraphics[width = 0.45\linewidth]{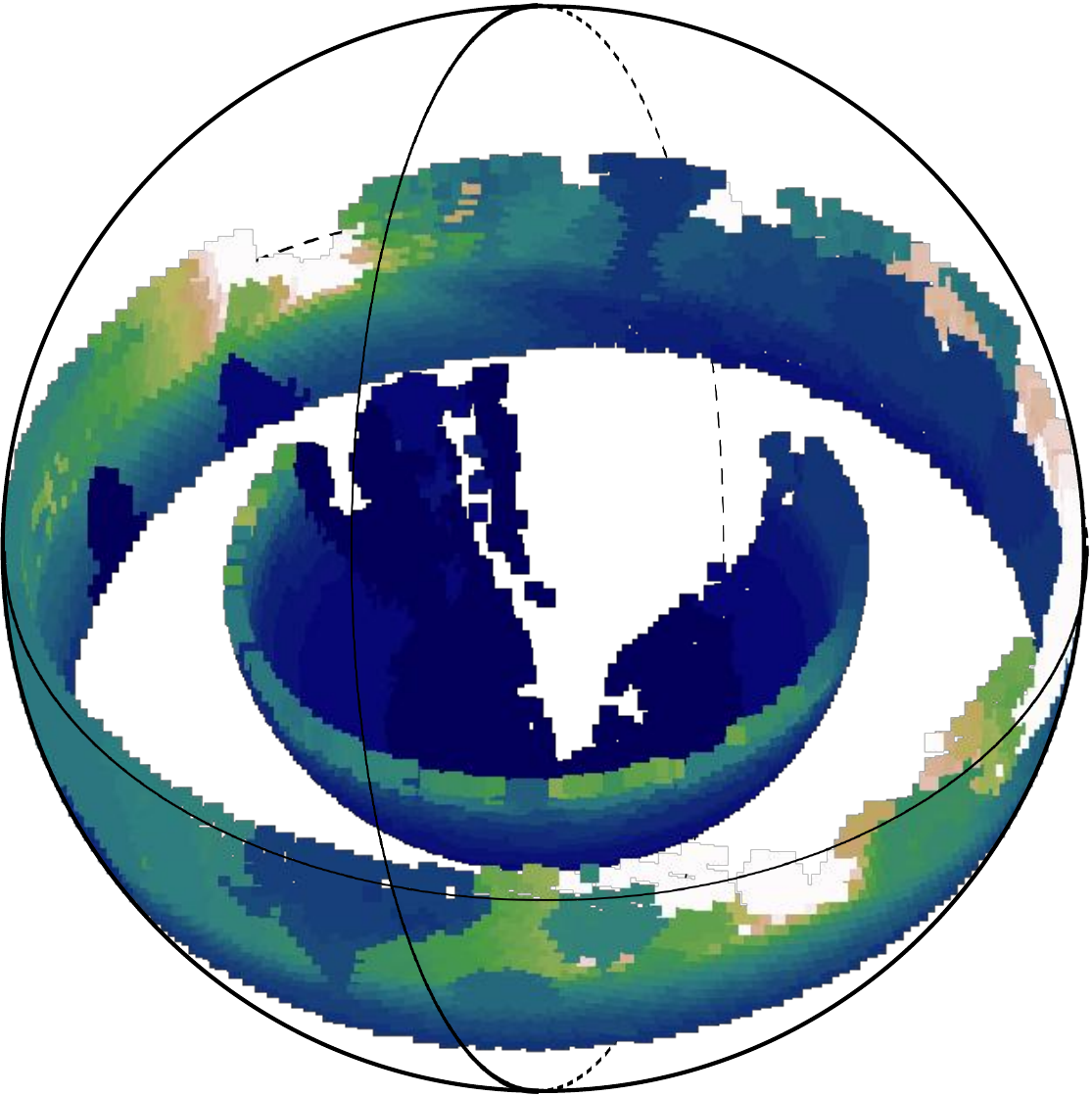}}
	\subfigure[USTC FLICAR] {\includegraphics[width = 0.45\linewidth]{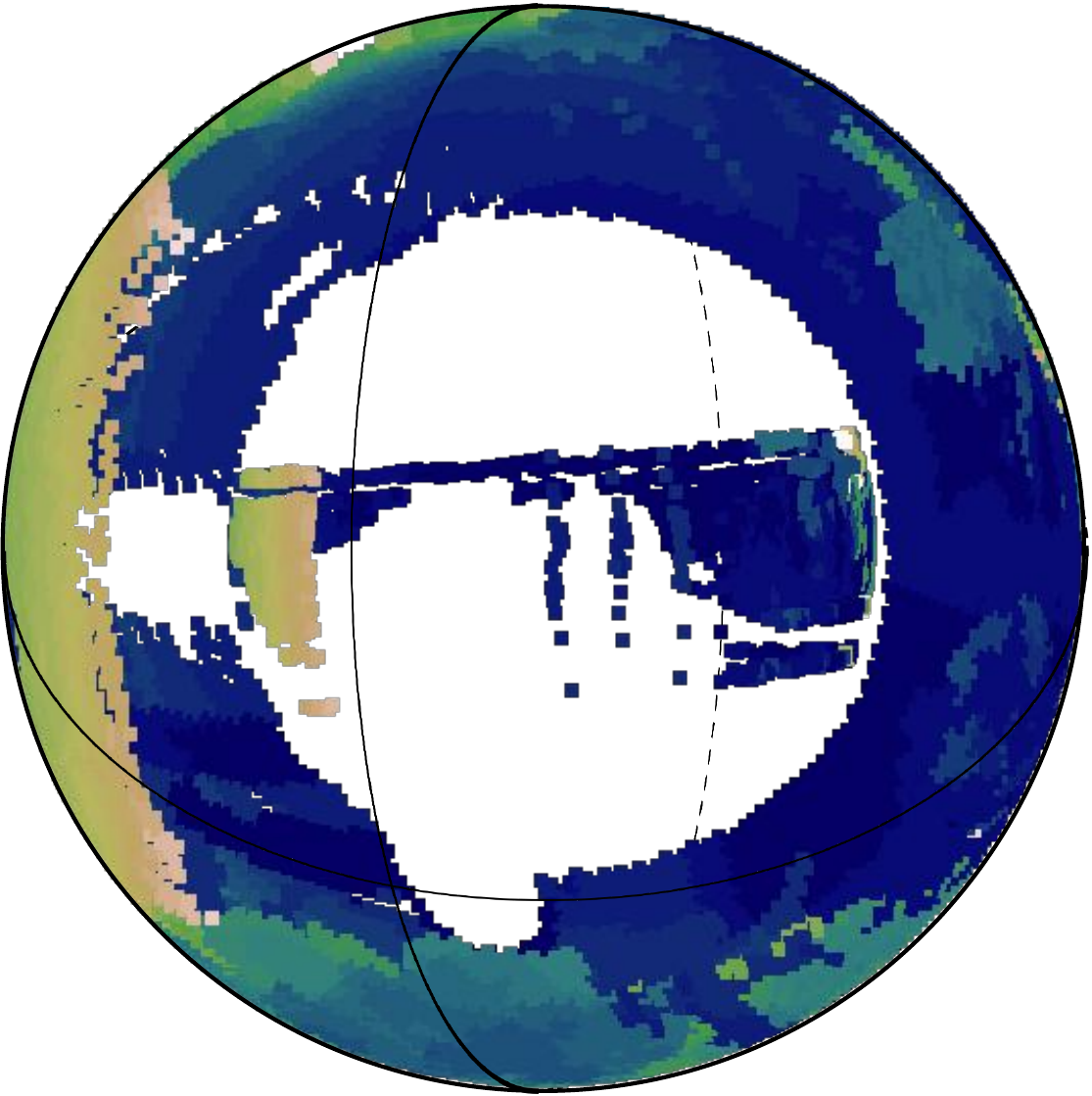}}
	\subfigure[A2D2] {\includegraphics[width = 0.45\linewidth]{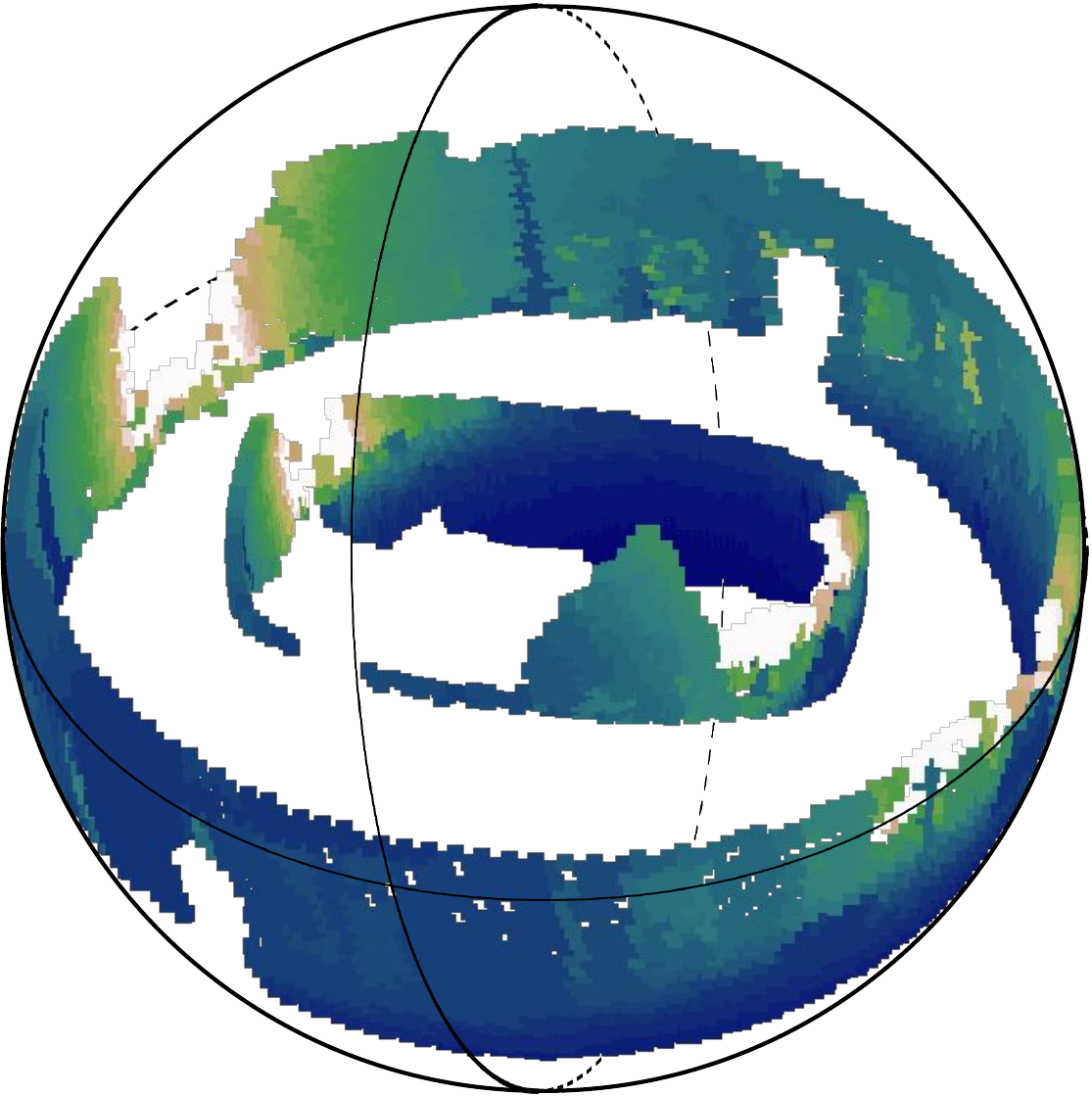}}
        \subfigure[UrbanNav]  {\includegraphics[width = 0.45\linewidth]{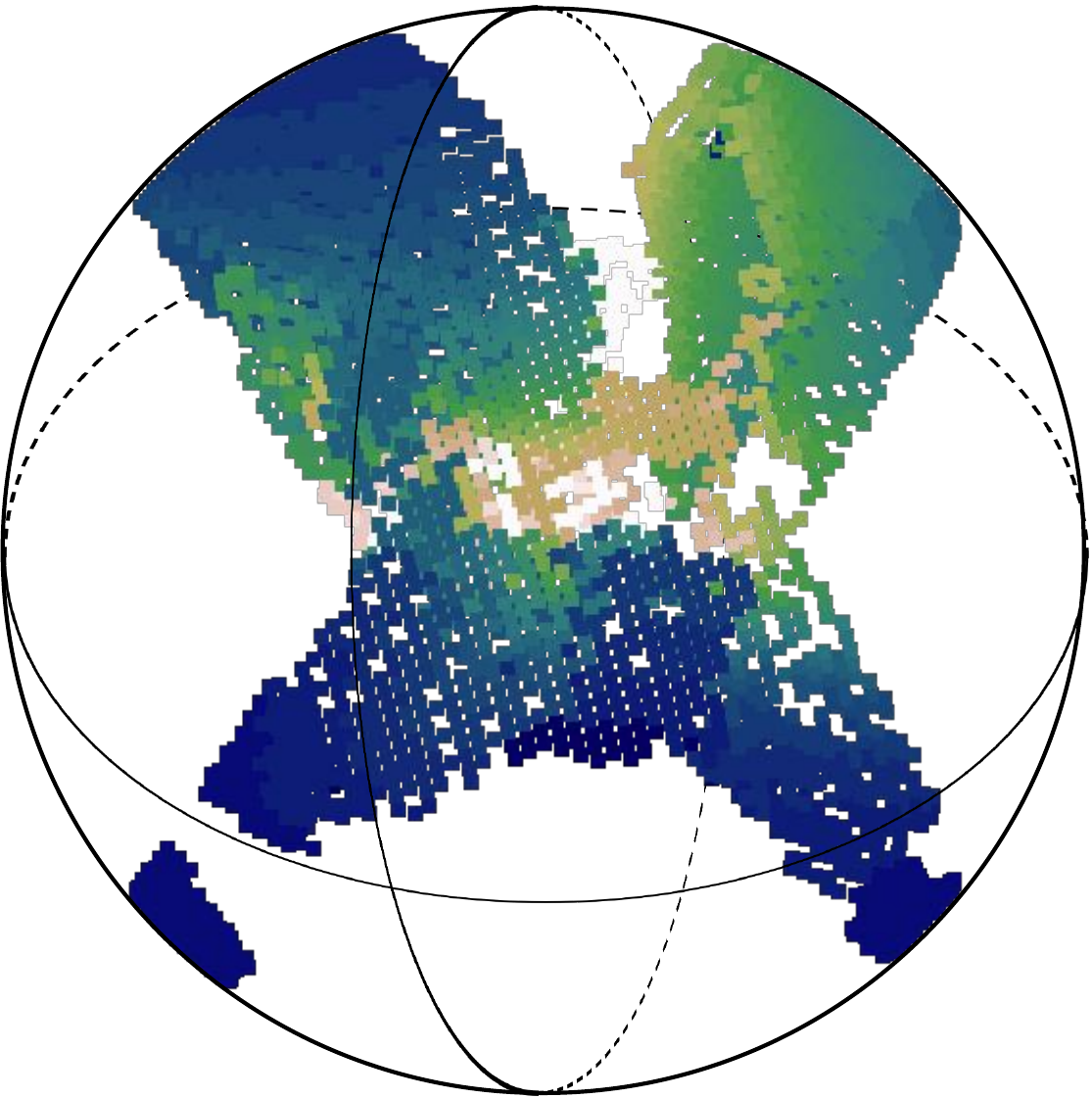}}
	\caption{The challenging example of the \sd in the real-world dataset. The \sd for the first LiDAR is represented by the outer sphere, while the second
is illustrated by the inner sphere. For UrbanNav, both spheres are the same size.
	}
	\label{fig:example_of_sd_real_world}
\end{figure}

Fig.~\ref{fig:example_of_sd_real_world} presents several challenging tasks. 
The LiDAR pairs showcased in USTC FLICAR, A2D2, and UrbanNav are all positioned with the LiDARs tilted, resulting in minimal overlapping areas between the two LiDARs. Additionally, the unique hemispherical FoV of the RS-Bpearl in the Campus-BS dataset presents significant challenges.

\subsubsection{Ablation study}
In this section, we test some key parameters affecting the performance of \sd, including the translation distance between LiDARs, the number of spherical sampling points, and the stride during the search.

\begin{figure}[t]
	\centering 
	\subfigure[64-32, parallel] {\includegraphics[width = 0.48\linewidth]{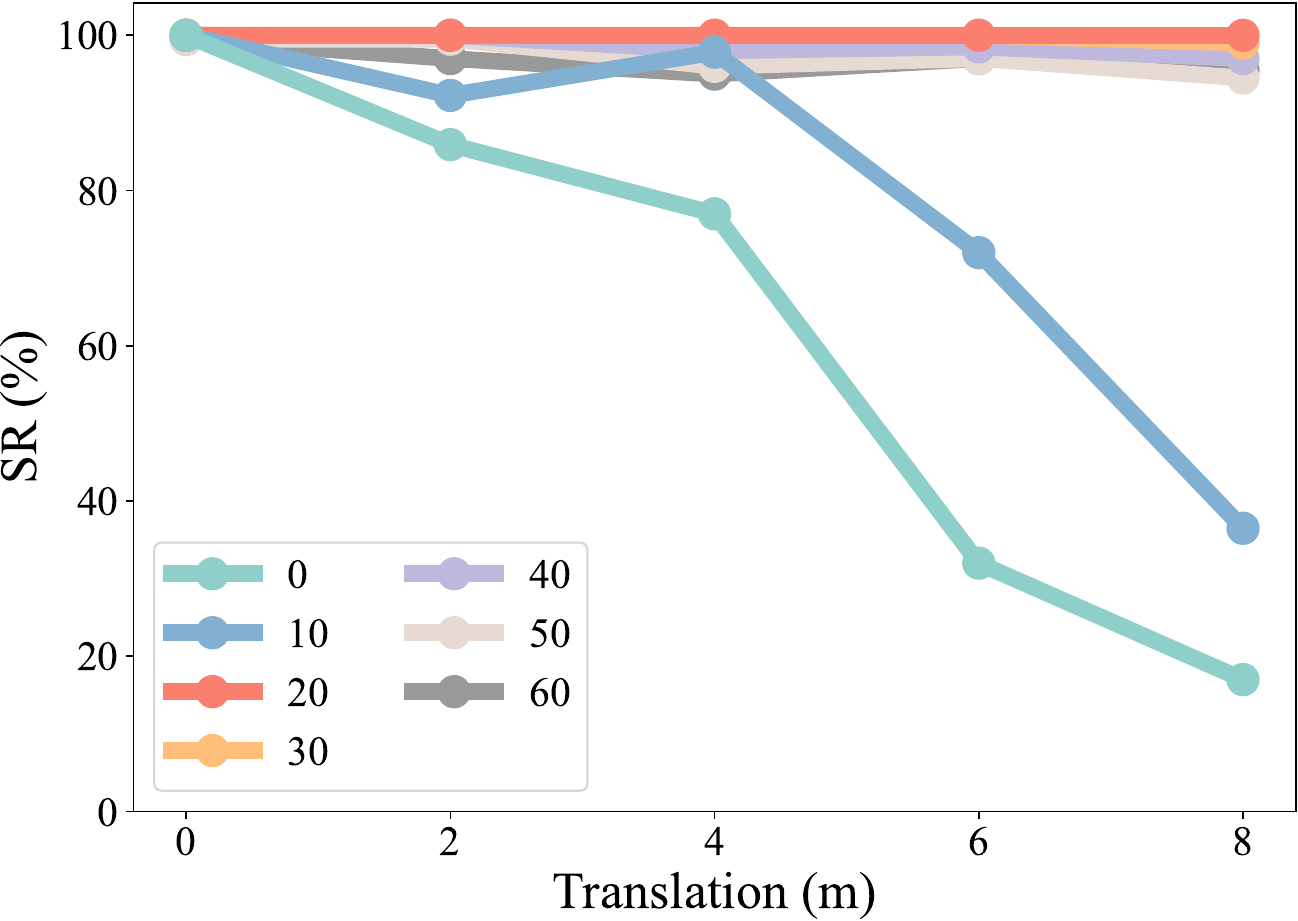}}
	\subfigure[64-32, perpendicular] {\includegraphics[width = 0.48\linewidth]{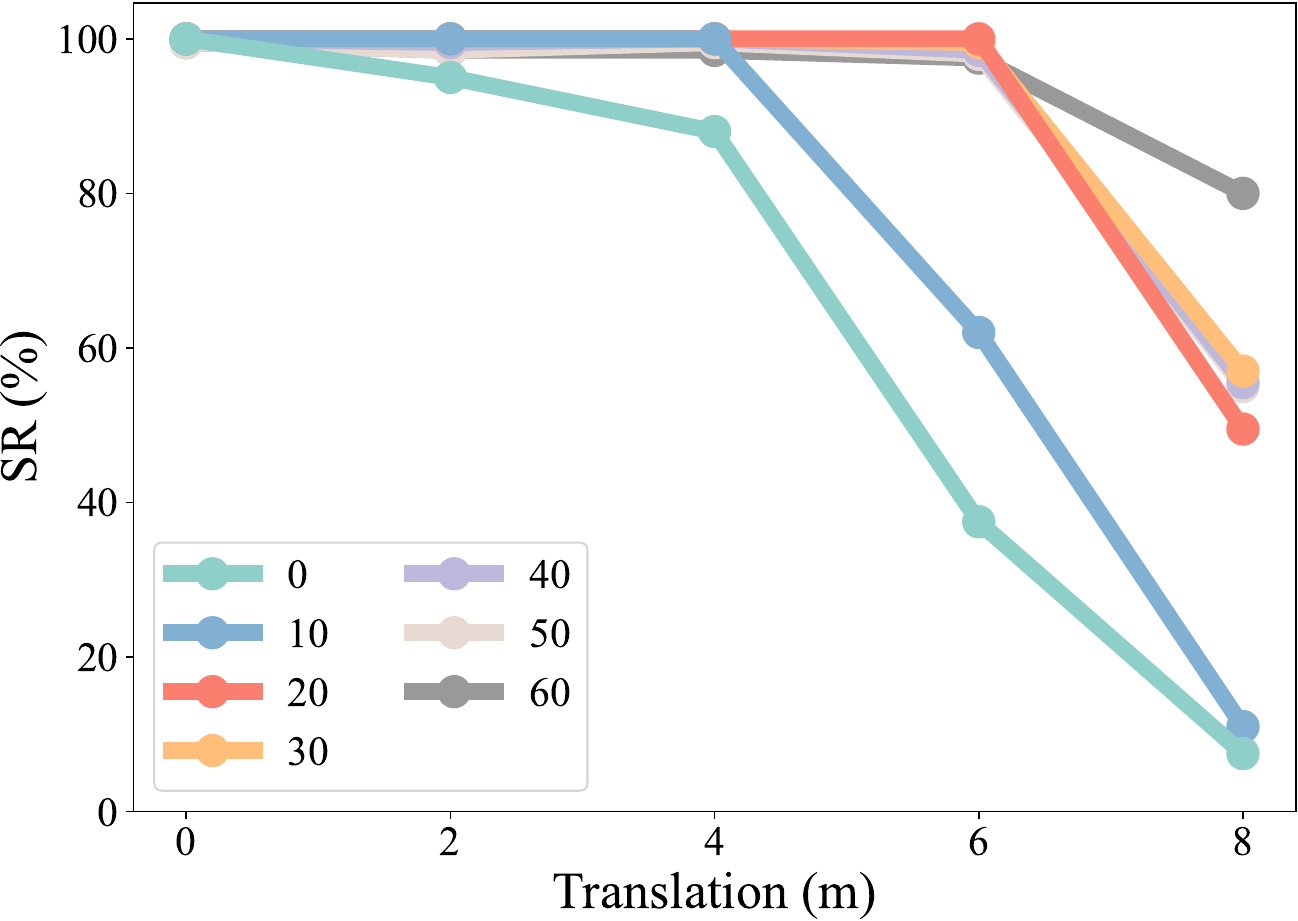}}
	\subfigure[64-Horizon, parallel] {\includegraphics[width = 0.48\linewidth]{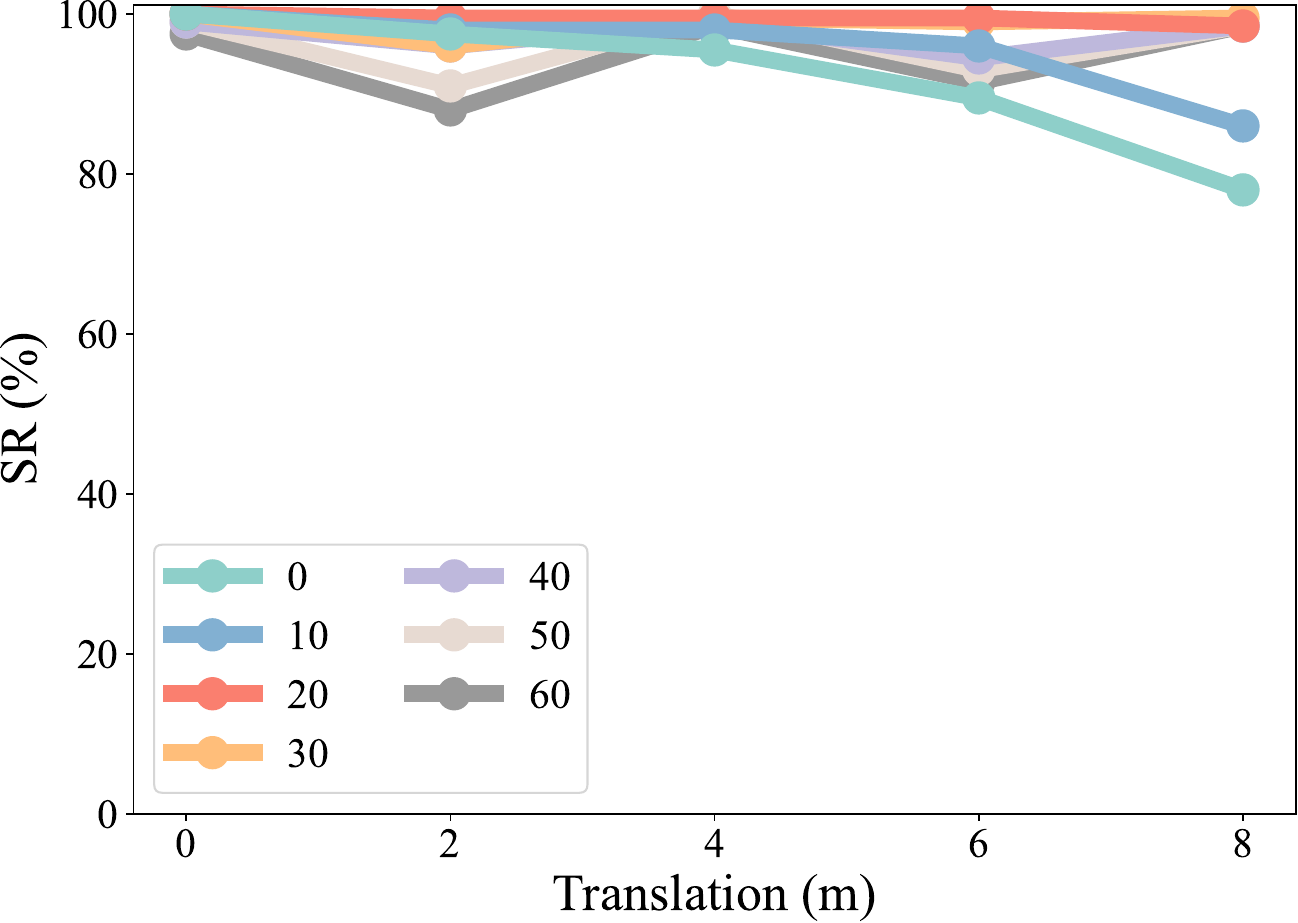}}
        \subfigure[64-Horizon, perpendicular]  {\includegraphics[width = 0.48\linewidth]{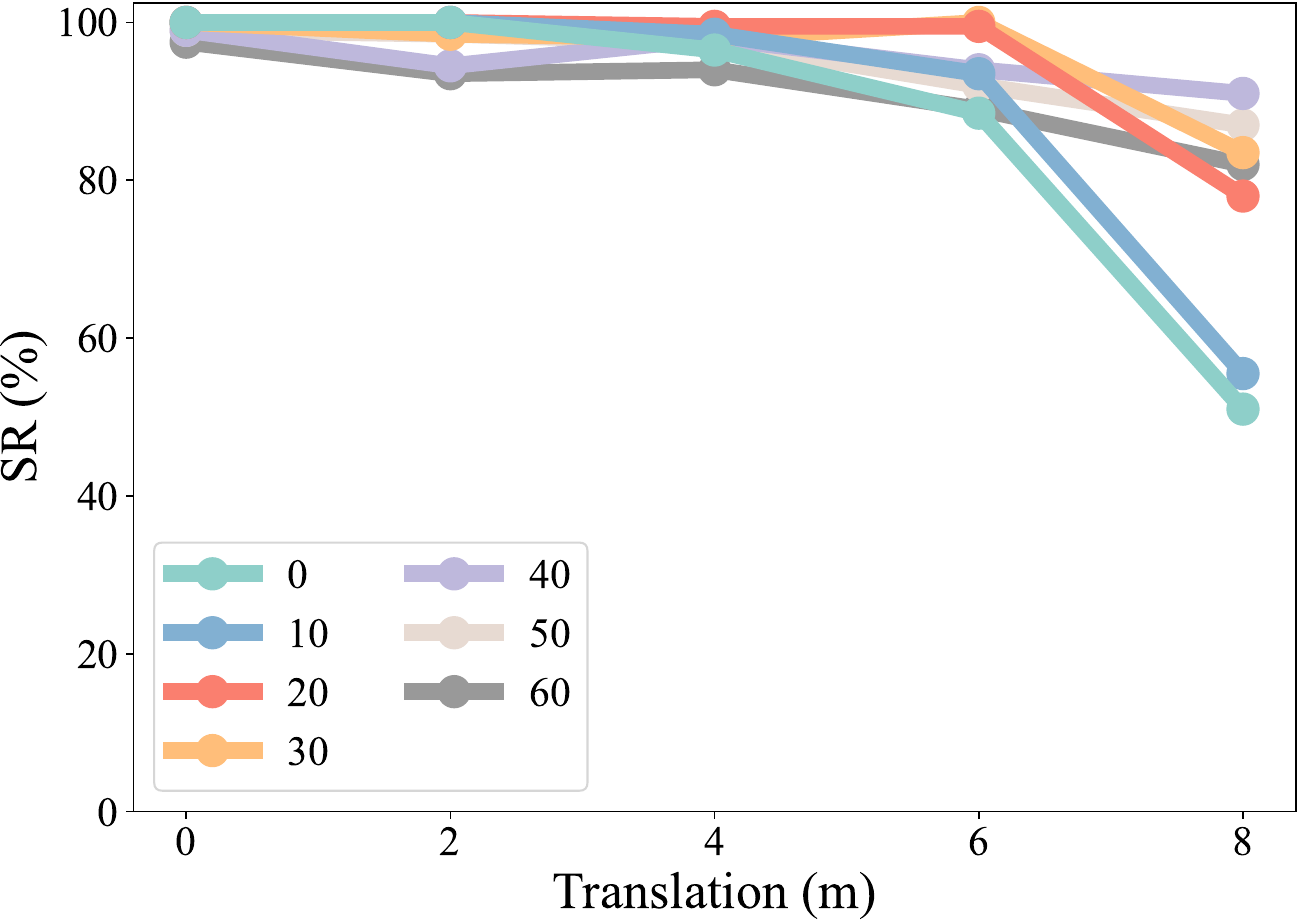}}
	\caption{The impact of translation on the success rate of rotation initialization and how the $d_1$ value mitigates this impact.}
	\label{fig:translation_abla}
\end{figure}

    \textbf{The impact of translation:} As \sd can only initialize rotation, the translation component introduces an error into the distance calculation in Eq.~\eqref{eq:feature_dis}. Within the COLORFUL dataset, for the same calibration tasks, i.e.,  64-32 and 64-Horizon, we introduce translations ranging from 0 to 8 meters in both the direction parallel and perpendicular to the vehicle's movement, thereby collecting  9 additional data sequences. The 0-meter translation implies the same location of the two LiDARs, which is impractical in reality but feasible in a simulation environment. Additionally, we conduct comparative experiments by selecting different values for $d_1$ in the Eq.~\eqref{eq:mask} to mitigate translation effects. The quantitative experimental results are illustrated in the Fig.~\ref{fig:translation_abla}. With a fixed $d_1$, the success rate of initialization decreases as the translation distance increases. When the translation distance is fixed, choosing $d_1$ values of $0m$ and $10m$ results in significantly poorer algorithm performance due to the disruption caused by nearby objects, compared to other values. Conversely, opting for larger  values, such as $50m$ or $60m$, the threshold becomes too high, reducing the number of available values in the distance calculation and thereby weakening the algorithm. After comprehensive evaluation, we fix $d_1$ at $20m$ for our algorithm.
    \begin{table}[t]
\center
\caption{The impact of sampling number $N$ on the \sd's precision}
\label{table:resolution}
\renewcommand{\arraystretch}{1.5}
\resizebox{\linewidth}{!}{
\begin{tabularx}{\linewidth}{>{\centering\arraybackslash}Xl|>{\centering\arraybackslash}X|>{\centering\arraybackslash}X|>{\centering\arraybackslash}X|>{\centering\arraybackslash}X|>{\centering\arraybackslash}X}
\hline
\multicolumn{2}{c|}{$N$}                                    & 100   &500    &1000  & 3000 &10000 \\ \hline
\multicolumn{2}{c|}{Resolution($^\circ$)}                                    & 18.98   &8.70    &6.10  & 3.55 &2.76 \\ \hline
\multicolumn{1}{l}{\multirow{2}{*}{64-32}}   & $e_{rot}$($^\circ$)\ $\downarrow$ &6.204  &2.953   &2.176&2.061 &0.935\\ 
\multicolumn{1}{l}{}                         & $SR$(\%)\ $\uparrow $           &13.08  &69.4    &100  &100   &100 \\ \hline
\multicolumn{1}{l}{\multirow{2}{*}{64-Horizon}} & $e_{rot}$($^\circ$)\ $\downarrow$ &4.978 &6.469   &4.835 &3.542 & 2.743\\ 
\multicolumn{1}{l}{}                         & $SR$(\%)\ $\uparrow  $          &5.70   &98.65  &99.66 &100   &100 \\ \hline
\multicolumn{1}{l}{\multirow{2}{*}{64-Tele}} & $e_{rot}$($^\circ$)\ $\downarrow$ & \textbackslash{} &   \textbackslash{}   &   \textbackslash{}  & 5.183 &4.995 \\ 
\multicolumn{1}{l}{}                         & $SR$(\%)\ $\uparrow$            & 0     & 0     &0     & 26.17  & 45.6 \\ \hline
\end{tabularx}}
\renewcommand{\arraystretch}{1}
\end{table}

    \begin{figure}[t]
	\centering
\subfigure[64-32-Eq.~\eqref{eq:feature_dis}] 
{\begin{tikzpicture}
\node[anchor=south west,inner sep=0] (image) at (0,0) {\includegraphics[width=0.48\linewidth]{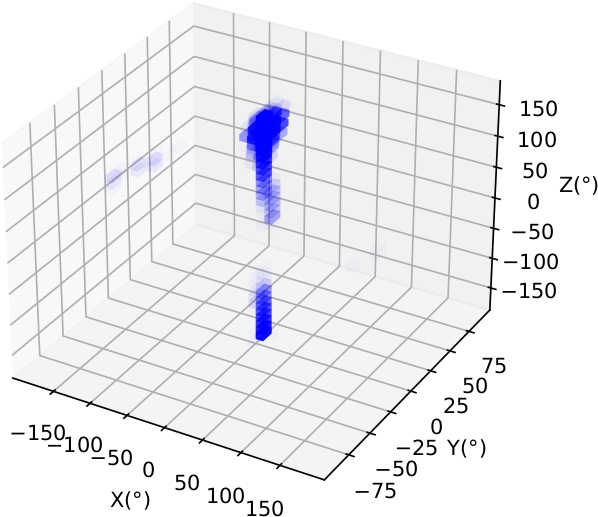}};
\begin{scope}[x={(image.south east)},y={(image.north west)}]
    \fill [red] (0.445,0.75) circle (2pt); 
\end{scope}
\end{tikzpicture}\label{fig:initgrid_a}}
\subfigure[64-Tele-Eq.~\eqref{eq:feature_dis}] 
{\begin{tikzpicture}
\node[anchor=south west,inner sep=0] (image) at (0,0) {\includegraphics[width=0.48\linewidth]{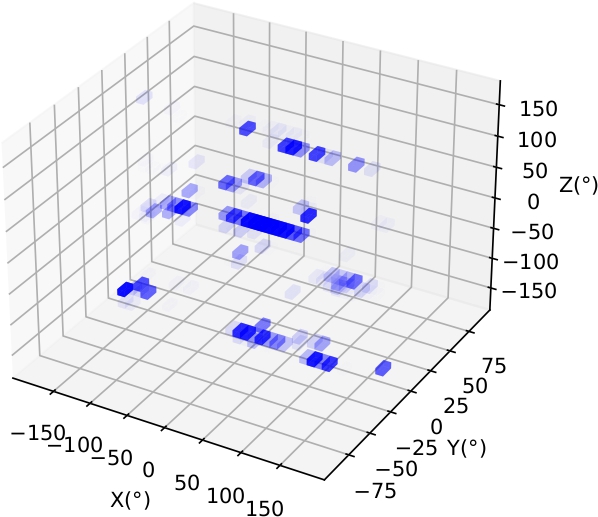}};
\begin{scope}[x={(image.south east)},y={(image.north west)}]
    \fill [red] (0.445, 0.565) circle (2pt); 
\end{scope}
\end{tikzpicture}\label{fig:initgrid_b}}
\subfigure[64-32-ED] 
{\begin{tikzpicture}
\node[anchor=south west,inner sep=0] (image) at (0,0) {\includegraphics[width=0.48\linewidth]{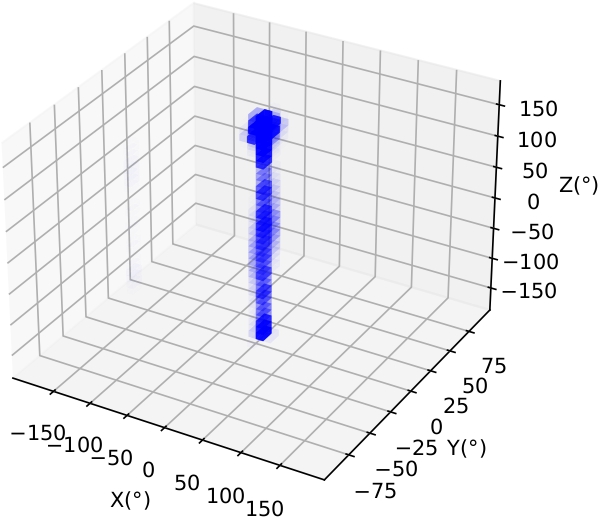}};
\begin{scope}[x={(image.south east)},y={(image.north west)}]
    \fill [red] (0.445,0.75) circle (2pt); 
\end{scope}
\end{tikzpicture}\label{fig:initgrid_c}}
\subfigure[64-Tele-ED] 
{\begin{tikzpicture}
\node[anchor=south west,inner sep=0] (image) at (0,0) {\includegraphics[width=0.48\linewidth]{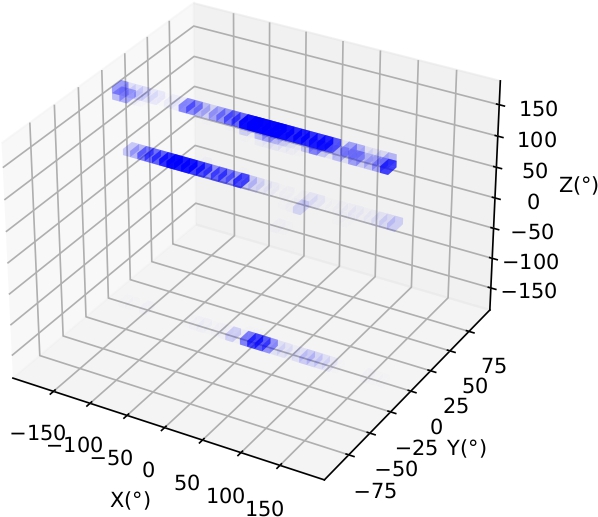}};
\begin{scope}[x={(image.south east)},y={(image.north west)}]
    \fill [red] (0.445, 0.565) circle (2pt); 
\end{scope}
\end{tikzpicture}\label{fig:initgrid_d}}\\
 	\caption{Rotations and their corresponding distances. The axes represent the Euler angles for rotation around the x, y, and z axes, respectively. The opacity signifies the value of the distance, where less opacity indicates a smaller distance. The red point represents the location of the ground truth. ED is an abbreviation for Euclidean Distance.}
	\label{fig:initgrid}
\end{figure}
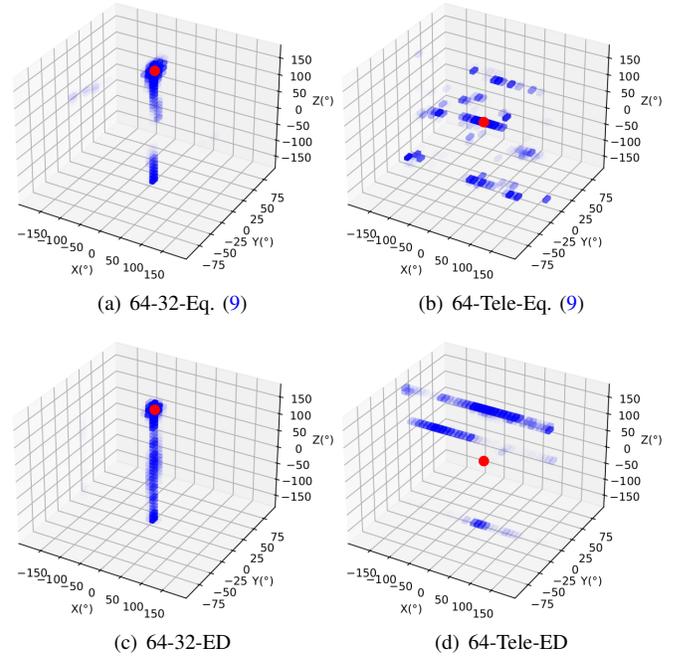
   \textbf{The impact of sampling number:} The sample number $N$  of Fibonacci lattice determines the resolution of \sd. We use 64-32, 64-Horizon and 64-Tele calibration tasks as examples to illustrate the performance of \sd at varying resolutions.  As demonstrated in Tab.~\ref{table:resolution}, the precision decreases as the resolution reduces. However, because the 32-beam LiDAR and Horizon has a wider FoV, it can still achieve acceptable success rate even at low resolutions. Meanwhile, Tele has only a $14.5^\circ \times 16.2^\circ$ FoV. When \sd's resolution exceeds $16.2^\circ$, it means that there is only one valid data in the entire descriptor, which will understandably lead to the algorithm's failure.
   
    \textbf{ The impact of  search stride:} Similar to the sampling number, the search stride also has an impact on performance. It's worth noting that the stride we talk here is the stride in the first stage, not the heuristic search stride in the second stage. The reason being, in the second stage, we only use a small stride such as $1^\circ$ to search, which is still not particularly time-consuming. However, in the first stage, the stride will increase the search time cubically, so a stride as large as possible is essential. Yet, an exceedingly large stride may result in a failure to fall into the locally convex space close to the ground truth during traversal, leading to a convergence to incorrect results. 
    
    As shown in Fig.~\ref{fig:initgrid_a} and Fig.~\ref{fig:initgrid_b}, we traverse the rotation space with a resolution of $10^\circ$ and calculate the corresponding distance for each rotation. Opacity signifies the value of the distance, with lesser opacity indicating a smaller distance. The small FoV of the Tele results in a small locally convex space, while in contrast, the locally convex space of the 64-32 is relatively large. In summary, for LiDARs like Tele, a large step size could likely cause the algorithm to miss local convex space, leading to failure. Therefore, in our algorithm, we use a $10^\circ$ step size, which is smaller than Tele's  FoV.

    \textbf{The impact of distance function:} As depicted in Fig.~\ref{fig:initgrid}, we compare the distance from Eq.~\ref{eq:feature_dis} and the Euclidean distance (ED). The two tasks tested perfectly demonstrate the two beneficial properties of our designed distance function. Firstly, as shown in Fig.~\ref{fig:initgrid_a} and Fig.~\ref{fig:initgrid_c}, our distance function has a clear gradient near the ground truth, which facilitates faster search. Additionally, for 64-Tele,  although there are some local optima in  the solution space shown in Fig.~\ref{fig:initgrid_b}, the ground truth remains the global optimum. However, using ED makes the global optimum far from the ground truth, which is the worst case.

\begin{table*}[t]
    \center
    \caption{Rotation Error($e_{rot}(^\circ)$), Translation Error($e_{trans}(cm)$) and Success Rate($SR(\%)$) of Different Approaches. \\The initial value is ground truth with $2m$ and $5^\circ$ translation and rotation disturbance.\\ Rotational Error and Translation Error are only recorded when the experiment is successful ($SR$\textgreater 0).}
    \label{table:refine}
    \renewcommand{\arraystretch}{1.5}
    \resizebox{\textwidth}{!}{
    \begin{tabular}{cc|ccccccccc|cccccc}
    \hline
    \multicolumn{2}{c|}{\multirow{2}{*}{Target}}                                                                                                 & \multicolumn{3}{c}{CROON~\cite{CROON}}                                  & \multicolumn{3}{c}{scan-scan}                                            & \multicolumn{3}{c|}{\je}                                                  & \multicolumn{3}{c}{MLCC~\cite{small_fov}}                               & \multicolumn{3}{c}{\hots  }      \\ \cline{3-17} 
    \multicolumn{2}{c|}{}                                                                                                                        & \multicolumn{1}{c}{$e_{rot}$\ $\downarrow$} & \multicolumn{1}{c}{$e_{trans}$\ $\downarrow$} & $SR$ \ $\uparrow$ & \multicolumn{1}{c}{$e_{rot}$\ $\downarrow$} & \multicolumn{1}{c}{$e_{trans}$\ $\downarrow$} & $SR$\ $\uparrow$   & \multicolumn{1}{c}{$e_{rot}$\ $\downarrow$} & \multicolumn{1}{c}{$e_{trans}$\ $\downarrow$} & $SR$\ $\uparrow$    & \multicolumn{1}{c}{$e_{rot}$\ $\downarrow$} & \multicolumn{1}{c}{$e_{trans}$\ $\downarrow$} & $SR$\ $\uparrow$  & \multicolumn{1}{c}{$e_{rot}$\ $\downarrow$} & \multicolumn{1}{c}{$e_{trans}$\ $\downarrow$} & $SR$\ $\uparrow$   \\ \hline
    \multicolumn{1}{c|}{}                          & \begin{tabular}[c]{@{}c@{}}LiDAR-64 \\ LiDAR-32\end{tabular}                                & 0.032   & 1.269   &25.63                                                & 0.176   & 4.911   & 20.84                                                & \textbf{0.007}      & \textbf{0.229}   & \textbf{ 100 }                   & 0.007   & \textbf{0.127}   & 100                                                 & \textbf{0.006}   & {0.144}   &\textbf{100}   \\ \cline{2-17} 
    \multicolumn{1}{c|}{\multirow{6}{*}{COLORFUL}} & \begin{tabular}[c]{@{}c@{}}LiDAR-32 \\ LiDAR-16\end{tabular}                                & 0.110   & 2.338   & 26.39                                               & 0.215   & 4.885   &  22.81                                               & \textbf{0.074}      & \textbf{1.378}   & \textbf{ 100 }                   & 0.130   & 1.820   & 100                                                 & \textbf{0.027}   & \textbf{0.197}   &\textbf{100}   \\ \cline{2-17} 
    \multicolumn{1}{c|}{}                          & \begin{tabular}[c]{@{}c@{}}LIVOX-Horizon \\ LIVOX-Tele\end{tabular}                         & 0.199   & 5.779   &0.830                                                & 0.133   & 5.351   & 1.740                                                & \textbf{0.061}      & \textbf{4.486}   & \textbf{49.12}                   & \textbf{0.026}   & 6.048   & 100                                                 & {0.030}   & \textbf{1.786}   &\textbf{100}   \\ \cline{2-17} 
    \multicolumn{1}{c|}{}                          & \begin{tabular}[c]{@{}c@{}}LiDAR-64 \\ LIVOX-Horizon\end{tabular}                           & 0.122   & 5.386   & 14.24                                               & 0.135   & 4.141   &  8.932                                               & \textbf{0.010}      & \textbf{0.793}   & \textbf{98.84}                   & \textbf{0.008}   & 1.640   & 100                                                 & {0.009}   & \textbf{1.374}   &\textbf{100}   \\ \cline{2-17} 
    \multicolumn{1}{c|}{}                          & \begin{tabular}[c]{@{}c@{}}LiDAR-32 \\ LIVOX-Avia\end{tabular}                             & 0.425   & 5.699   &1.044                                                & 0.275   & 5.852   & 6.148                                                 & \textbf{0.089}      & \textbf{3.079}   & \textbf{91.10}                   & 0.041   & 2.862   & 100                                                 & \textbf{0.025}   & \textbf{0.926}   &\textbf{100}   \\ \cline{2-17} 
    \multicolumn{1}{c|}{}                          & \begin{tabular}[c]{@{}c@{}}LiDAR-64 \\ LIVOX-Tele\end{tabular}                              & 0.215   & 5.290   & 0.573                                               & 0.164   & 5.824   &  1.082                                               & \textbf{0.079}      & \textbf{4.060}   & \textbf{55.89}                   & \textbf{0.031}   & 6.482   & 100                                                 & {0.032}   & \textbf{2.356}   &\textbf{100}   \\ \hline 
    \multicolumn{1}{c|}{Campus-SS}                 & \begin{tabular}[c]{@{}c@{}}RS-LiDAR-M1 \\ LIVOX-AVIA\end{tabular}                           & 0.313  & 7.609   & 10.4                                                 & 0.534   & 6.962   & 1.301                                                & \textbf{0.205}      & \textbf{3.003}   & \textbf{95.12}                   & \textbf{0.159}   & 2.277   & 100                                                 & {0.234}   & \textbf{0.461}   & \textbf{100} \\ \cline{1-17} 
    \multicolumn{1}{c|}{\multirow{2}{*}{\raisebox{-3ex}{Campus-BS}}}& \begin{tabular}[c]{@{}c@{}}RS-Helios(top) \\ RS-Helios(back)\end{tabular}  & /       &    /    & 0                                                   & /       &    /    & 0                                                    & \textbf{0.325}      & \textbf{2.623}   & \textbf{96.34}                   & \textbf{0.284}   & 2.215   & 100                                                 & {0.349}   & \textbf{1.555}   & \textbf{100}   \\ \cline{2-17} 
    \multicolumn{1}{c|}{}                          & \begin{tabular}[c]{@{}c@{}}RS-Helios(top) \\ RS-Bpearl(right)\end{tabular}                  & /       &    /    & 0                                                   & \textbf{0.815}   & 7.524   & 0.975                                                & {0.820}      & \textbf{7.285}   & \textbf{50.19}                   & \textbf{0.428}   & \textbf{3.46}    & 100                                                 & 0.6914           & 4.635            & \textbf{100}  \\ \cline{1-17} 
    \multicolumn{1}{c|}{USTC FLICAR}               & \begin{tabular}[c]{@{}c@{}}Ouster OS0-128 \\ VLP-32C\end{tabular}                           & /       &    /    & 0                                                   & /       &    /    & 0                                                    & \textbf{0.791}      & \textbf{4.854}   & \textbf{41.25}                   & \textbf{0.505}   & 6.521   & 100                                                 & 0.5123           & \textbf{4.151}            & \textbf{100} \\ \cline{1-17} 
    \multicolumn{1}{c|}{\multirow{3}{*}{\raisebox{-3ex}{TIERS}}}    & \begin{tabular}[c]{@{}c@{}}LIVOX Horizon \\ LIVOX AVIA\end{tabular}        & 0.3867  & 7.021   & 13.56                                               & 0.433   & \textbf{6.601}   & 8.861                                                & \textbf{0.307}      & {8.051}   & \textbf{88.69}                   & /       &    /    & 0                                                   & \textbf{0.117}   & \textbf{3.656}   & \textbf{100}  \\ \cline{2-17} 
    \multicolumn{1}{c|}{}                          & \begin{tabular}[c]{@{}c@{}}OS-1 \\ LIVOX AVIA\end{tabular}                                  & 0.2255  & 6.83    & 8.108                                               & 0.804   & 7.158   & 3.436                                                & \textbf{0.201}      & \textbf{3.032}   & \textbf{97.50}                   & 0.150   & 5.109   & 100                                                 & \textbf{0.103}   & \textbf{1.248}   & \textbf{100}   \\ \cline{2-17} 
    \multicolumn{1}{c|}{}                          & \begin{tabular}[c]{@{}c@{}}OS-0 \\ OS-1\end{tabular}                                        & 0.9093  & 5.983   & 25.41                                               & 0.776   & 6.866   & 5.606                                                & \textbf{0.051}      & \textbf{1.787}   & \textbf{100  }                   & /       &    /    & 0                                                   & \textbf{0.1815}           & \textbf{4.196}            & \textbf{100}  \\ \cline{1-17} 
    \multicolumn{1}{c|}{Pandaset}                  & \begin{tabular}[c]{@{}c@{}}Pandar64\\ PandarGT\end{tabular}                                 & 0.142   & 4.318   & 15.69                                               & \textbf{0.069}   & 5.32    & 4.167                                                & {0.143}      & \textbf{3.018}   & \textbf{100  }                   & \textbf{0.025}   & \textbf{1.355}   & 100                                                 & 0.1258           & 2.046            & \textbf{100}  \\ \cline{1-17} 
    \multicolumn{1}{c|}{A2D2}                      & \begin{tabular}[c]{@{}c@{}}frontcenter\\ frontleft\end{tabular}                             & /       &    /    & 0                                                   &  /      & /       & 0                                                    & \textbf{0.721}      & \textbf{3.025}   & \textbf{30.02}                   & \textbf{0.094}   & 5.697   & 100                                                 & 0.7225           & \textbf{2.9}     & \textbf{100}    \\ \cline{1-17} 
    \multicolumn{1}{c|}{UrbanNav}                  & \begin{tabular}[c]{@{}c@{}}HDL-32E\\ Lslidar C16\end{tabular}                               & 0.9301  & 9.256   & 0.4926                                              & 0.772   & \textbf{2.708}   & 0.218                                                & \textbf{0.748}      & {9.215}   & \textbf{32}                      & /       &    /    & 0                                                   & \textbf{0.3709}  & \textbf{5.903}     & \textbf{100}   \\ \hline      

    \end{tabular}}
\end{table*}

\subsection{Study of Refinement}
\label{sec:refineexp}
In this section, we conduct experiments to evaluate the accuracy of the proposed \je and \hots for extrinsic refinement. We compare their performance with several commonly used methods, including  CROON~\cite{CROON}  and MLCC~\cite{small_fov}. In addition to this, we also conduct tests on the direct scan-scan method. It is noteworthy that the classic LiDAR calibration algorithm M-LOAM~\cite{mloam}, which can only calibrate mechanical LiDARs, is not suitable for our task setting, and therefore, it has not been included in our comparison.

\subsubsection{Quantitative experiment}

As indicated in the Tab.~\ref{table:refine}, we evaluate the extrinsic refinement methods on six types of LiDAR pairs of COLORFUL dataset and ten types of real-world datasets. We perturb the ground truth to serve as the input initial values, including a $5^\circ$ rotation and a $2m$ translation. Among the five methods, CROON and scan-scan requires single point cloud pairs as input. For \je, the number of input frames is set to five, with registration parameters consistent with those in scan-scan. Lastly,  MLCC  and \hots utilize the entire sequence to generate the final result. Specifically, MLCC and \hots require the trajectories of each LiDAR as input, which we estimate  using the Multi-LiDAR SLAM described in Sec.~\ref{sec:je}, where the required extrinsic parameters are substituted with the ground truth disturbed by the average error of \je on the corresponding dataset. 

\begin{figure}[t]
	\centering 
	\subfigure[Campus-BS (top-right)] {\includegraphics[width = 0.45\linewidth]{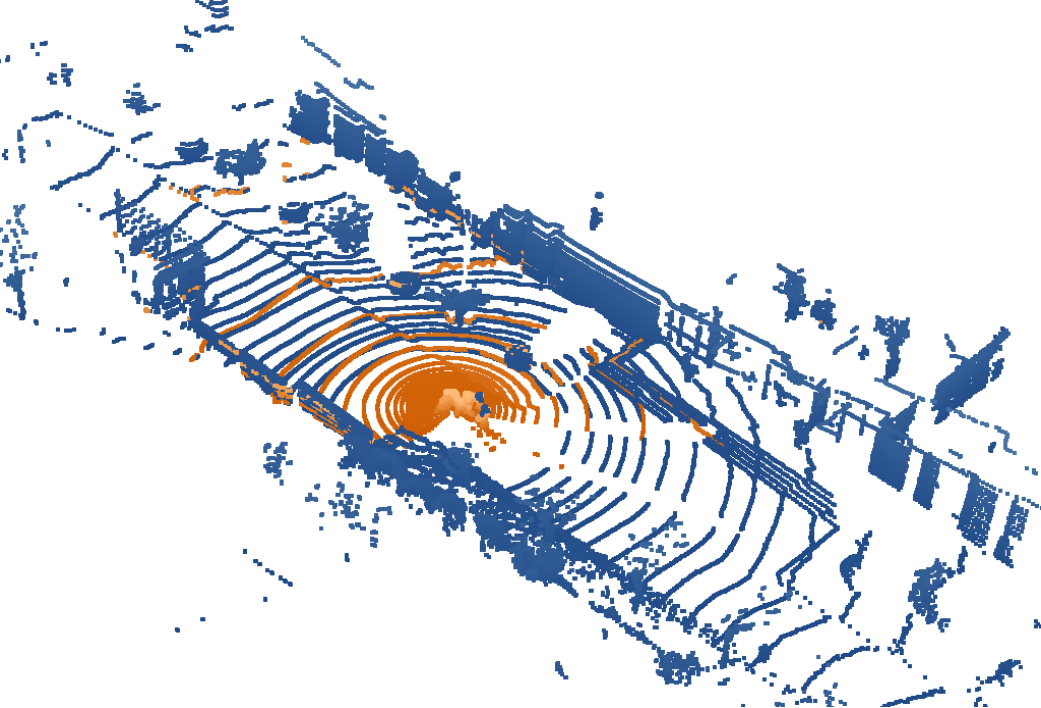}}
	\subfigure[USTC FLICAR] {\includegraphics[width = 0.45\linewidth]{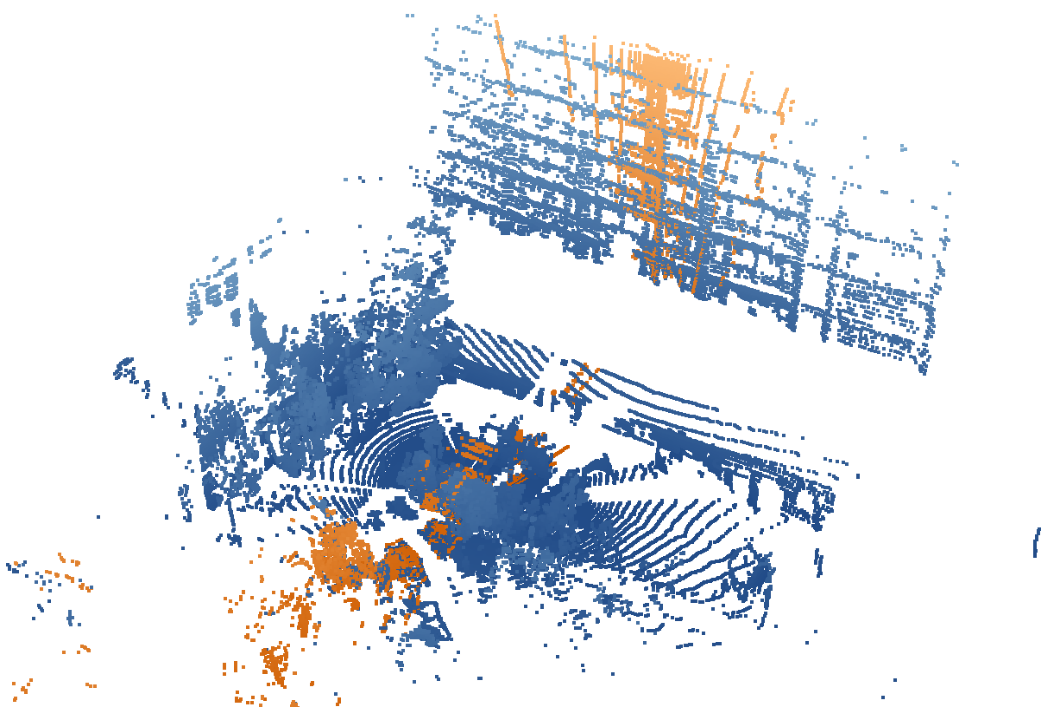}}
	\subfigure[A2D2] {\includegraphics[width = 0.45\linewidth]{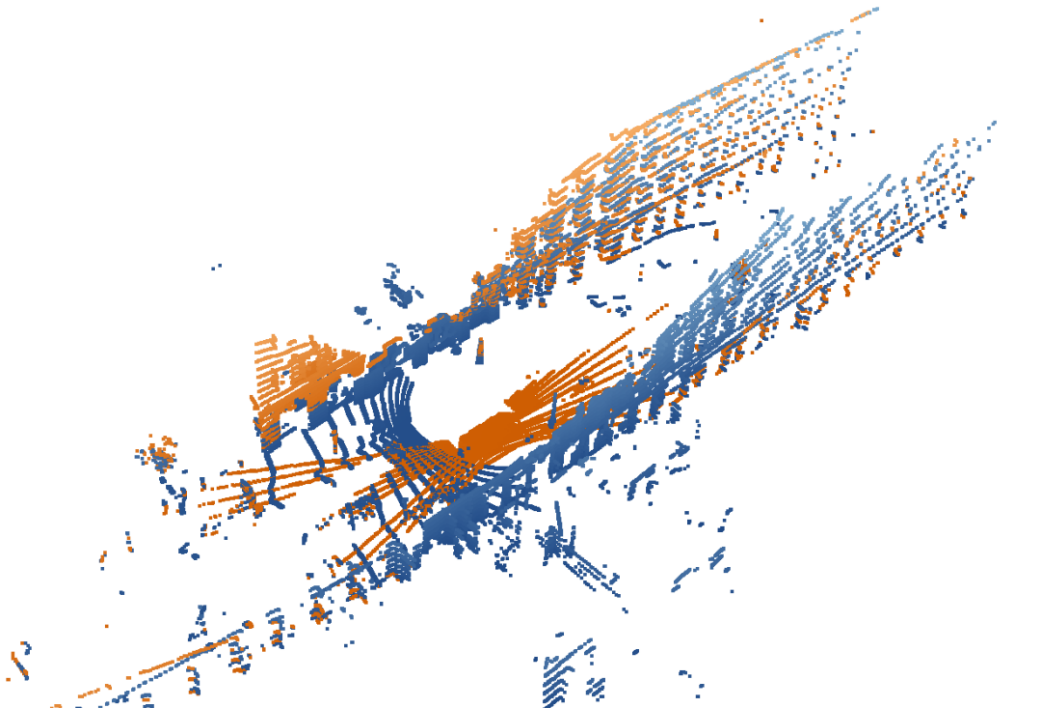}}
        \subfigure[UrbanNav]  {\includegraphics[width = 0.45\linewidth]{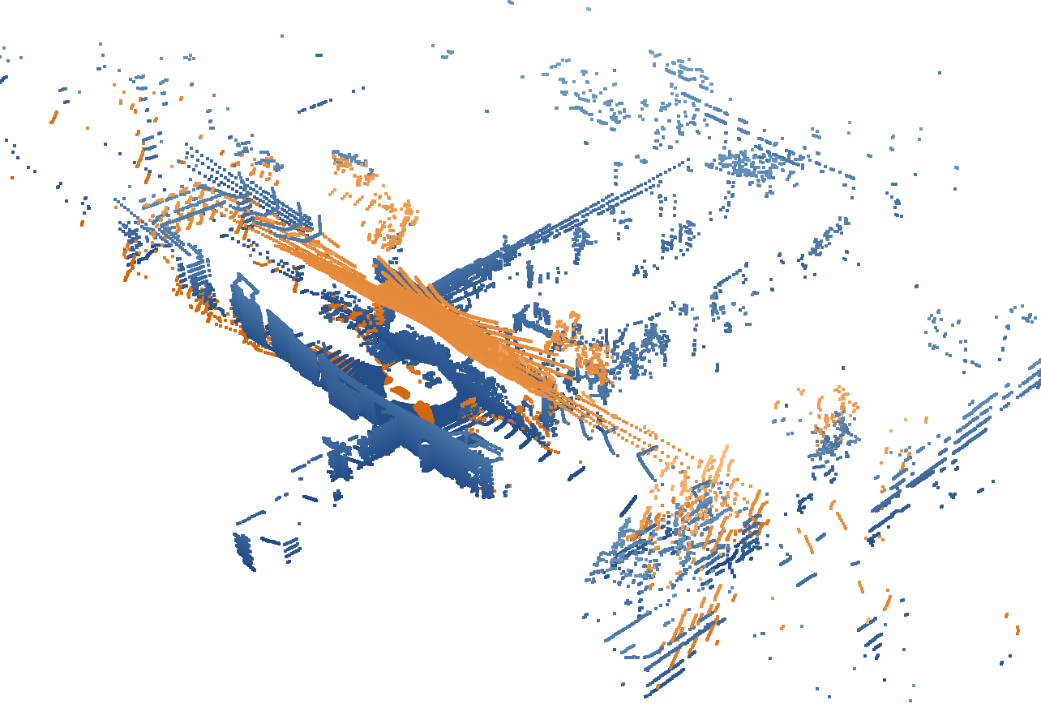}}
	\caption{The example of the difficult calibration targets in the real-world dataset. 
	}
	\label{fig:example_of_je_real_world}
\end{figure}

As shown in the table on the left, due to the poor initial values, as well as the distribution of the point cloud and the degeneracy of the scene, both CROON and the direct scan-to-scan methods have low success rates, failing entirely on some challenging tasks. In contrast, after introducing the constraints of motion, the success rate of \je significantly improves. Even in the challenging tasks such as Campus-BS, USTC FLICAR, A2D2, and UrbanNav, we maintain a success rate of over 30\%. The calibrated point clouds for these four tasks are displayed in Fig.~\ref{fig:example_of_je_real_world}.

With the use of all available data and backend optimization, MLCC  shows enhanced performance. Benefiting from the hierarchical optimization design, our approach ensures better global consistency of the map, resulting in more accurate poses and extrinsic parameters.
\subsubsection{Ablation study}
\label{sec:exp_refine_abla}
    In this section, we test the impact of initial value on \je, the  impact of the number of frames input into JEEP and the impact of time synchronization.

\begin{figure}[t]
	\centering 
	\subfigure[64-32 $e_{rot}$] {\includegraphics[width = 0.325\linewidth]{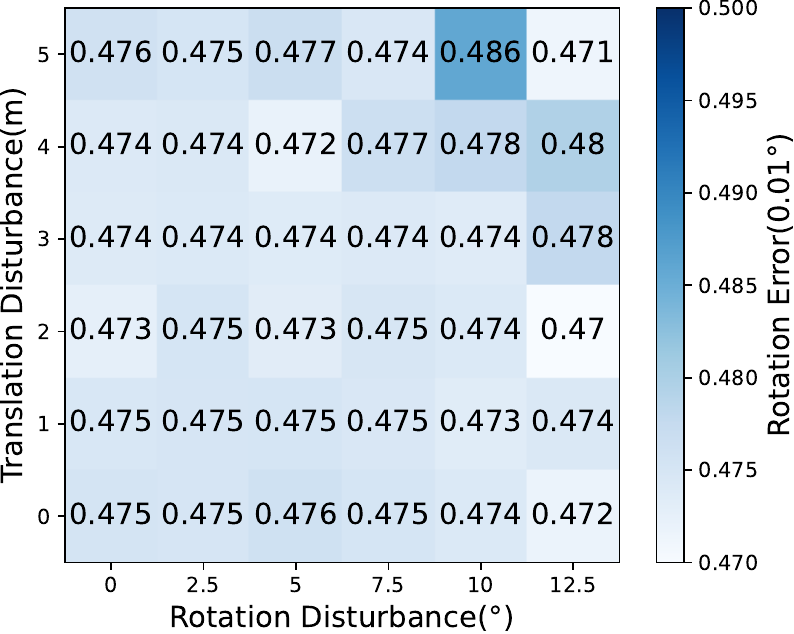}}
	\subfigure[64-32 $e_{trans}$] {\includegraphics[width = 0.32\linewidth]{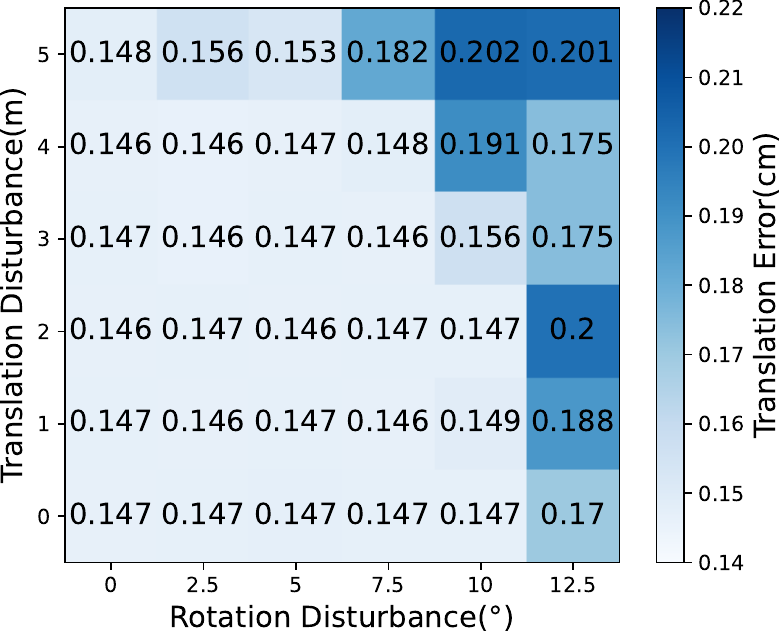}}
	\subfigure[64-32 $SR$] {\includegraphics[width = 0.317\linewidth]{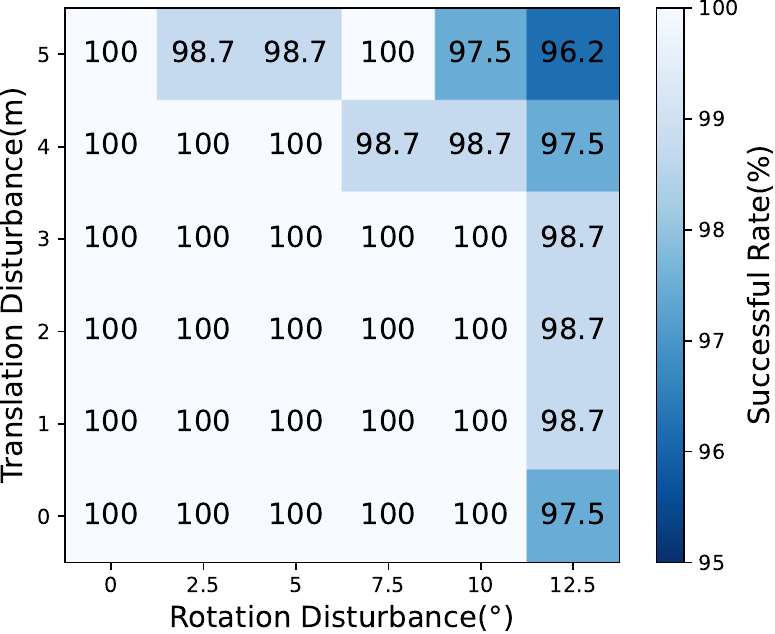}}
    \subfigure[64-Horizon $e_{rot}$] {\includegraphics[width = 0.325\linewidth]{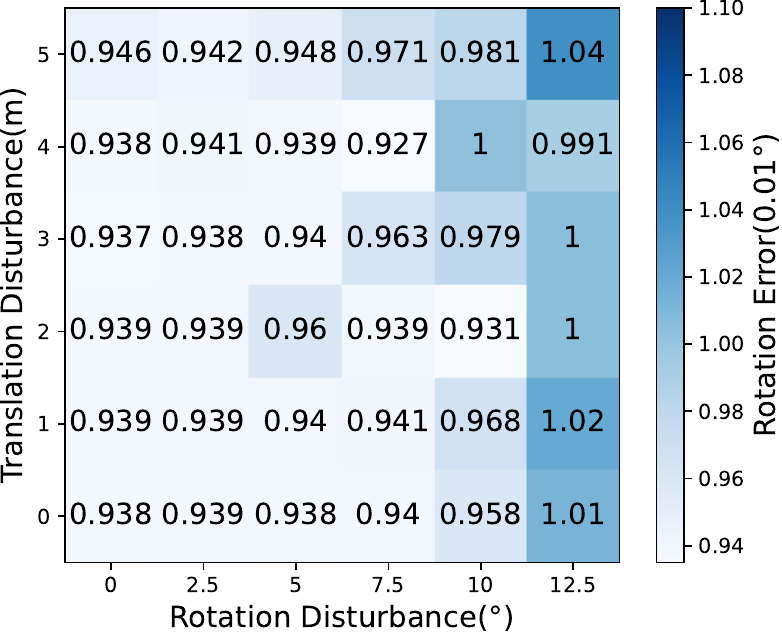}}
	\subfigure[64-Horizon $e_{trans}$] {\includegraphics[width = 0.32\linewidth]{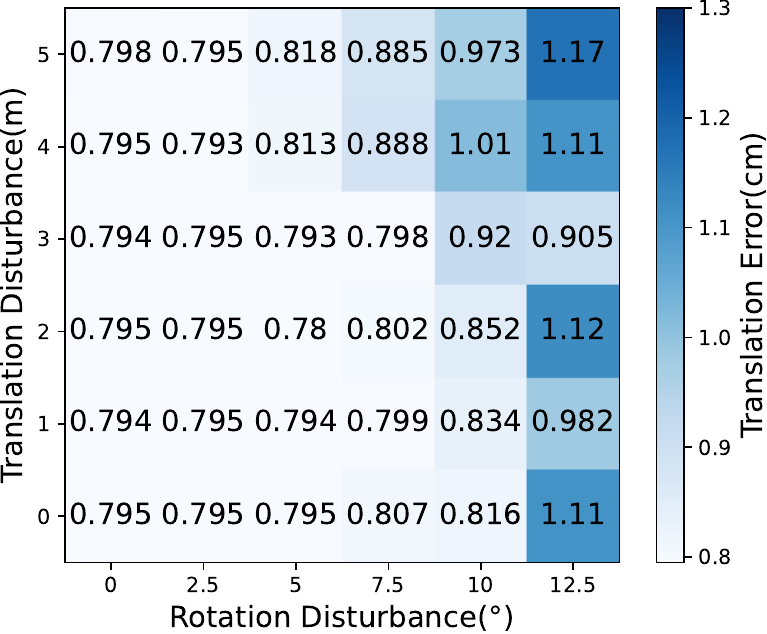}}
	\subfigure[64-Horizon $SR$] {\includegraphics[width = 0.323\linewidth]{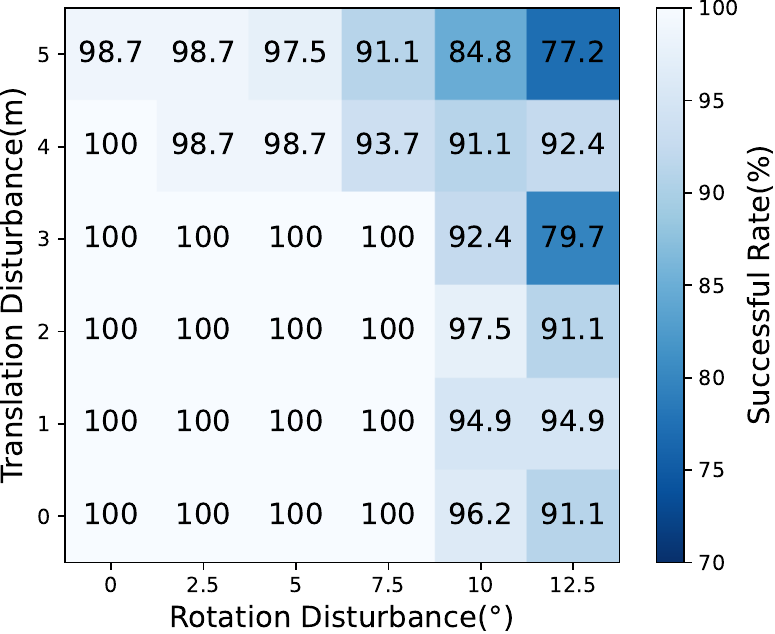}}
	\caption{Calibration results of \je under different rotation and translation disturbance in COLORFUL dataset. 
	}
	\label{fig:refine_disturb_ablation}
\end{figure}

    \textbf{The impact of initial value:} To test the robustness of \je under various initial conditions, we sample rotation disturbances from $0$ to $12.5$ degrees and translation  disturbances from $0$ to $5$ meters, conducting $6\times 6$ experiments for each LiDAR-pair. The heatmaps of rotation errors, translation errors, and success rates, as shown in Fig.~\ref{fig:refine_disturb_ablation}, which demonstrate that \je is highly robust to inaccurate initial values, achieving an $80\%$ success rate even with disturbances of $10$ degrees and $5$ meters.
    Moreover, it is crucial to note that this  experiment demonstrates that within a 5-meter range, translation disturbance doesn't significantly impact the final results. However, when rotation disturbance exceeded $10^\circ$, noticeable declines are observed in the performance of the refinement process. This phenomenon validates the correctness and necessity of performing only rotation initialization for single-platform calibration.

\begin{figure}[t]
	\centering 
 \subfigure[$SR$] {\includegraphics[width = 0.32\linewidth]{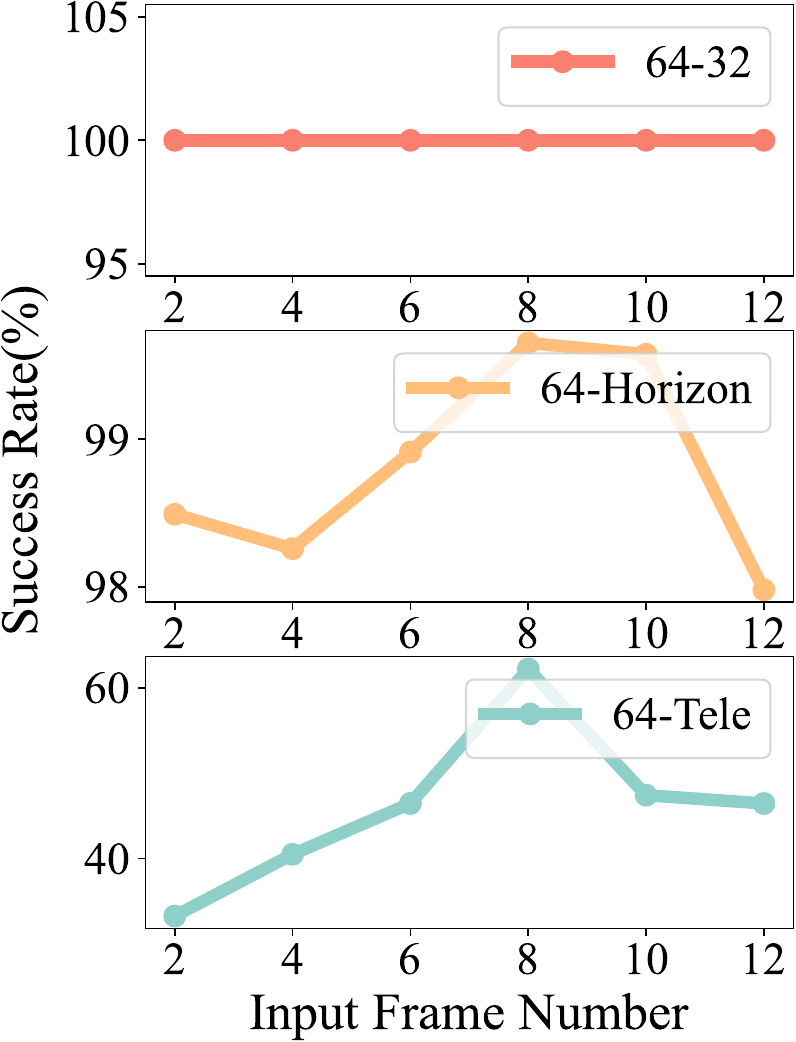}}
	\subfigure[$e_{trans}$] {\includegraphics[width = 0.32\linewidth]{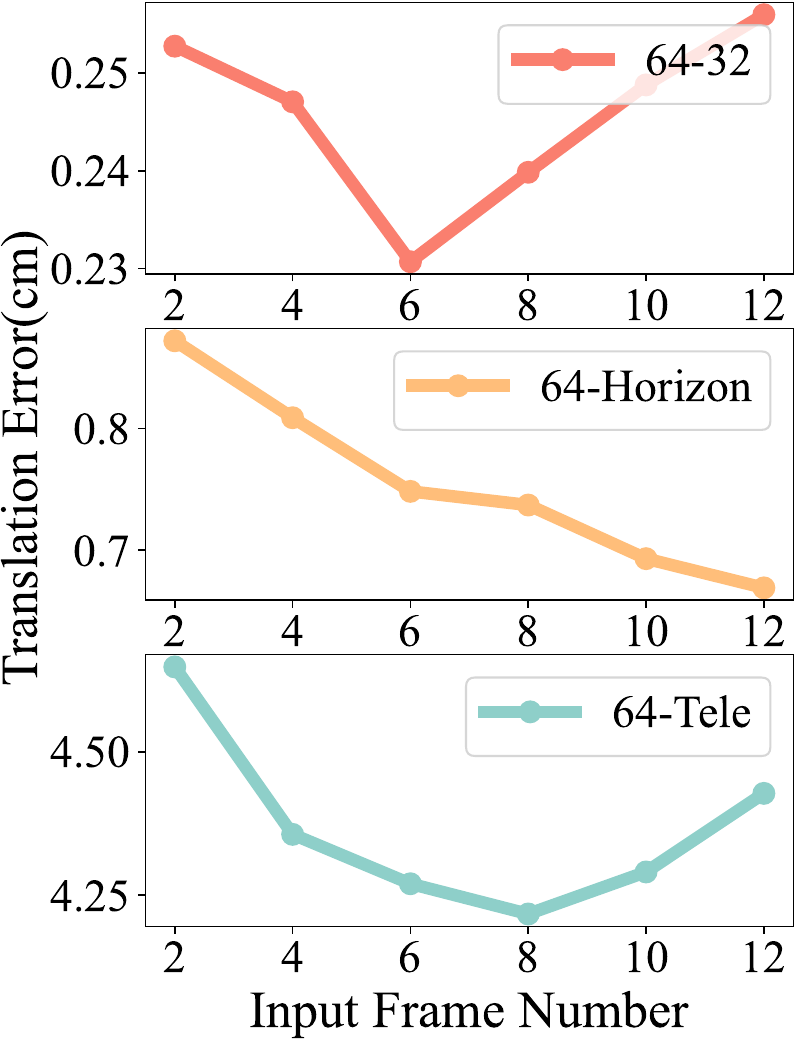}}
	\subfigure[$e_{rot}$] {\includegraphics[width = 0.32\linewidth]{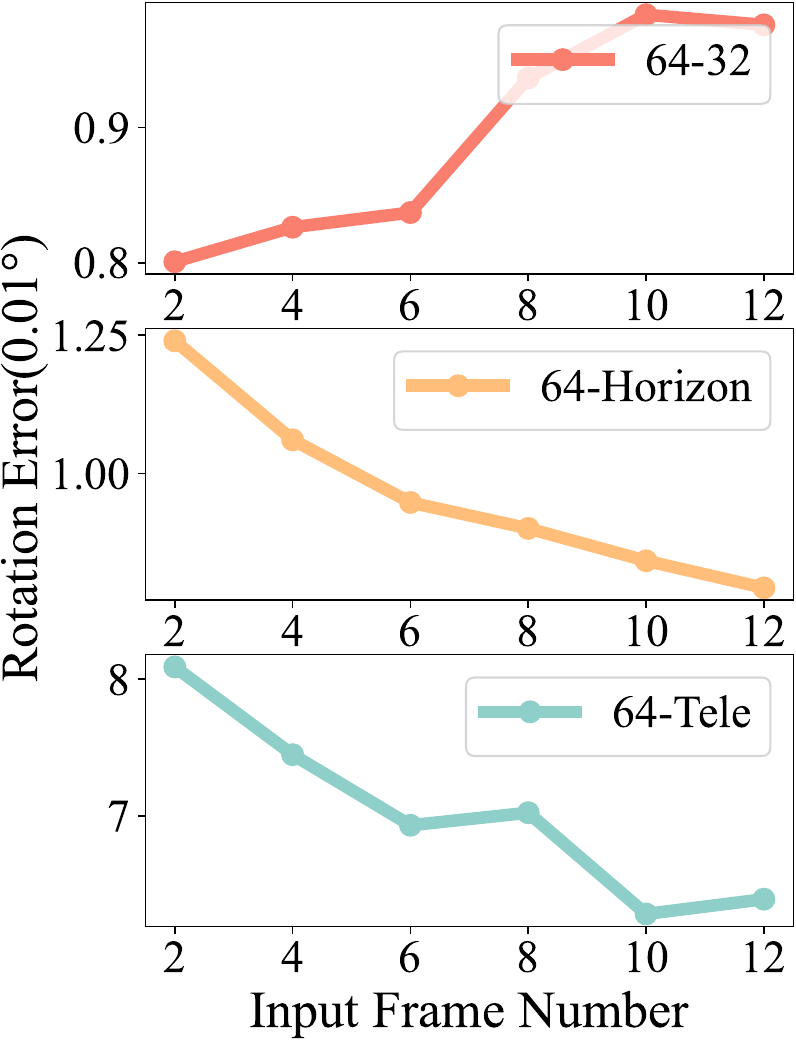}}
	\caption{Calibration results of \je for different input frame number in COLORFUL dataset.
	}
	\label{fig:refine_num_ablation}
\end{figure}

   \textbf{The impact of the frame number:} The number of 
    frames input into \je is an interesting parameter. As analyzed in Sec.~\ref{sec:je}, too few frames fail to provide sufficient constrains, while too many frames introduce cumulative errors of pose estimation to extrinsic calibration. The disturbance remain consistent with our main experiment, i.e., 2 meters and 5 degrees. We adjust the frame number from 2 to 12 to investigate the effects of varying the number of input frames.  The rotational and translational errors and successful rates are shown in the Fig.~\ref{fig:refine_num_ablation}. For tasks like 64-32 calibration, where the point cloud distribution is already uniform, fewer frames can achieve good results, and increasing the number of frames is not beneficial. However, for tasks like 64-Horizon and 64-Tele calibration, which inherently lack sufficient  constraints, a certain number of frames is necessary to achieve satisfactory results.

    \textbf{The impact of  \ts:}
To validate the effectiveness of the time synchronization introduced in Sec.~\ref{sec:time_syn}, we reduce the frequency of the data in COLORFUL dataset from $20Hz$ to $10Hz$. In specific, we remove data from two LiDARs at different times to maintain a $50ms$ interval between them. It is important to note that for $10Hz$ data, $50ms$ represents the maximum achievable time offset. We repeat the experiments on \je and \hots according to the previous experimental setup, and the results are displayed in the Tab.~\ref{table:timesyn}. 
Due to the lack of time synchronization, errors caused by motion are introduced to  calibration, resulting in \je's success rate being significantly inferior to those previously shown in Tab.~\ref{table:refine}. In fact, these successful cases, accounting for about $20\%$ of the total, are attributed to the fact that approximately $20\%$ of the data collection period involves minimal movement. Therefore, the specific values of $e_{rot}$ and $e_{trans}$ are not meaningful in explaining the calibration performance in this context. Alternatively, we use extrinsic parameters with disturbances of $1^\circ$ and $10cm$ for the multi-LiDAR SLAM, and further optimize the output pose using \ho.

\begin{figure}[t]
	\centering 
	\subfigure[64-32] {\includegraphics[width = 0.48\linewidth]{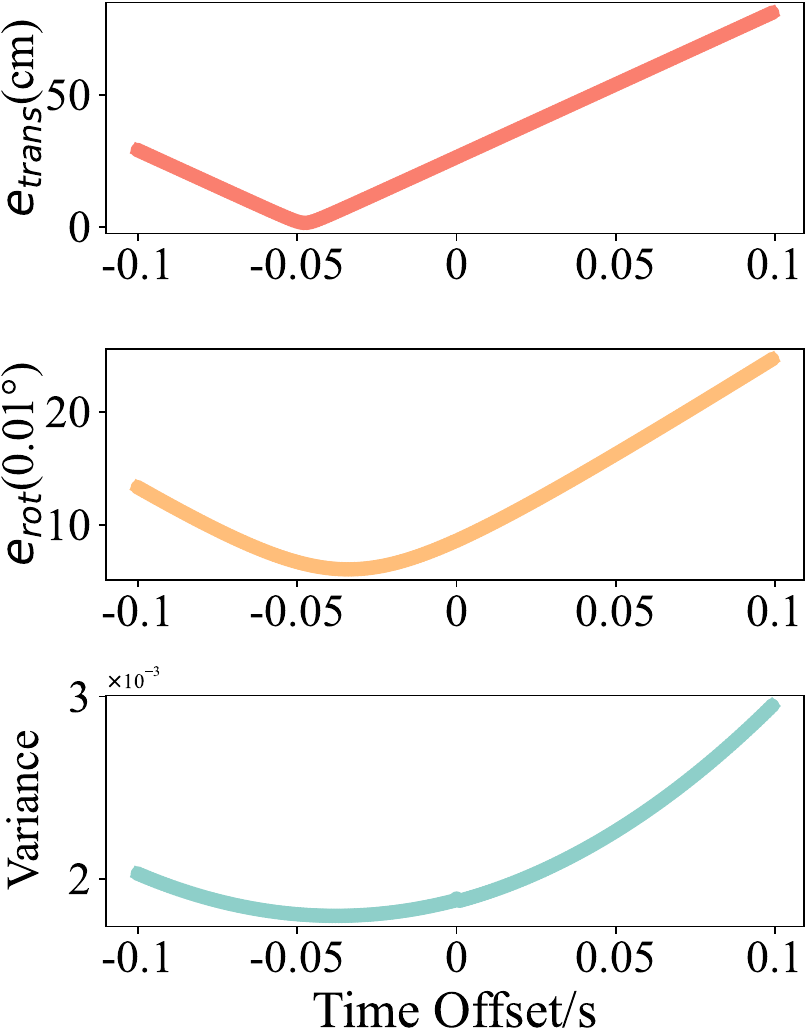}}
	\subfigure[64-Horizon] {\includegraphics[width = 0.48\linewidth]{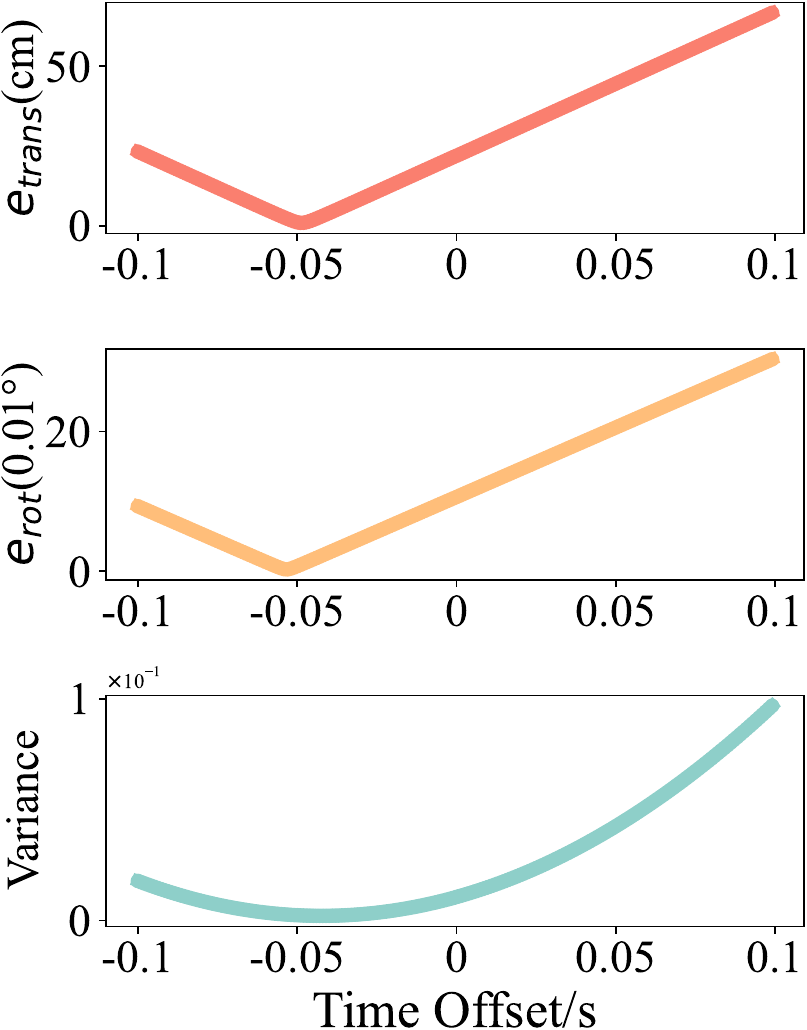}}
    \caption{The relationship between the sampled time offset and extrinsic variance. The ground truth time offset is $-50ms$.
	}
    \label{fig:time_syn_exp}
\end{figure}

After applying \ho, a noticeable improvement is achieved, yielding results comparable to previous ones. Furthermore, the estimated time offsets are close to the ground truth of $50 ms$, demonstrating that our method can achieve millisecond-level time synchronization accuracy. Fig.~\ref{fig:time_syn_exp} illustrates the relationship between the sampled time offset and extrinsic variance, translation error, and rotation error. It is observed that both translation and rotation errors reach their minimum values around $-50 ms$, and the variance shows a positive correlation with them. This indicates the validity of using variance to infer the true extrinsic parameters.


\begin{table}[t]
\center
\caption{Experiments on dataset with 50ms time offset.}
\label{table:timesyn}
\renewcommand{\arraystretch}{1.5}


\begin{tabularx}{\linewidth}{cc|>{\centering\arraybackslash}X|>{\centering\arraybackslash}X}
\hline
\multicolumn{2}{c|}{Target}                                                                                                   & 64-32   &   64-Horizon  \\ \hline
\multicolumn{1}{c}{\multirow{3}{*}{\je}} & $e_{rot}(^\circ)$\ $\downarrow$                                                                    & 0.052   &     0.033     \\  
\multicolumn{1}{c}{}                                    & $e_{trans}(cm)$\ $\downarrow$                                                           &  0.738  &     0.894     \\ 
\multicolumn{1}{c}{}                                    & $SR$(\%)\ $\uparrow$                                                          &  20.69  &  20.23        \\ \hline
\multicolumn{1}{c}{\multirow{3}{*}{\hots}} & $e_{rot}(^\circ)$\ $\downarrow$     &  0.061  &  0.022        \\  
\multicolumn{1}{c}{}                                    & $e_{trans}(cm)$\ $\downarrow$                                                           & 5.508   &   3.148    \\ 
\multicolumn{1}{c}{}                                    & $SR$(\%)\ $\uparrow$                                                           &  100    &  100          \\ \hline
\multicolumn{2}{c|}{Time Offset(ms)}                                                                                          &  -37     & -42            \\ \hline
\end{tabularx}
\end{table}

\section{Discussion and Future Work}
In this section, we discuss the limitations of the proposed methods and list some potential  future work.
\subsection{\sd}

The previous experiments confirm the robustness of \sd in various scenarios and tasks. However, \sd is not a descriptor that can be universally applied across all scenarios.
In the design of \sd, we enhance its robustness against translation between two LiDARs by filtering nearby objects. This implies that in indoor environments where the translation between two LiDARs is comparable to the room size, \sd is likely  to underperform. Additionally, for other calibration tasks, such as roadside LiDARs calibration~\cite{li2024farfusion,ren2023trajmatch}, where the translation is on the same scale as the scanning radius of the LiDARs, \sd also fails to function properly.

However, these limitations also provide new inspiration. The generation process of \sd fundamentally involves uniformly down-sampling the point cloud by angle, which means that \sd does not lose the 3D structural information of the point cloud. Currently, \sd is designed for estimating the rotation between two LiDARs, but  with the appropriate loss design, \sd still has the potential application on 6-DoF  pose estimation.
 
\subsection{Refinement}

In the first step of refinement, we employ \je effectively addressing registration failures caused by minimal overlapping areas or sparsity of point clouds by alternately estimating extrinsic parameters and poses. This method also offers the advantage of determining the success of calibration by monitoring the iteration count. 
Specifically, when the calibration parameters closely approximate the ground truth, the pose estimation process using data from both LiDARs typically converges easily. Furthermore, the accuracy of the incrementally constructed local map, facilitated by precise pose estimation, promotes convergence in the calibration step. Ideally, a single cycle of pose estimation and calibration may suffice.  In this case, we can affirm the success of the calibration and advance to the subsequent phase of accelerated multi-LiDAR SLAM.

Although calibrating during motion offers the aforementioned benefits, the effectiveness is undoubtedly influenced by the motion despite the used backend optimization steps. 
Firstly, the obvious effect is that motion causes distortion in the point cloud, which is detrimental to point cloud registration. Additionally, due to imperfect timestamp alignment, the estimated extrinsic parameters  include components of motion as shown in Fig.~\ref{fig:time_syn}. Although we reduce this effect through time synchronization steps, our practice of interpolating motion trajectories based on the assumption of the constant velocity is still not precise. 
A direct solution is to feed the algorithm with point clouds collected in a stationary state from different locations, similar to \cite{nie2022automatic, LiDAR_link}.  
However, this approach still requires a fixed calibration data collection step, unlike our approach, which is applicable  to any data segment. Furthermore, we intend to adopt the continuous-time SLAM approach~\cite{zheng2024traj, dellenbach2022ct} in the future, 
treating point cloud sequences as a continuous point stream rather than temporally discrete frame, which allows for more precise modeling of continuous-time trajectories.

\section{Conclusion}
In this paper, we design a universal LiDAR calibration method including four core steps, resolving the calibration challenges of initialization and refinement brought by different FoV and point cloud distributions due to the variety of LiDAR types. 
Firstly, we design a universal spherical descriptor \sd to initialize the rotation part of the extrinsics without any prior knowledge. Following this, a method called \je, for the joint estimation of extrinsic and poses, is proposed to achieve optimized, usable extrinsics. To further refine these parameters with additional data, we implement a straightforward multi-LiDAR SLAM approach and utilize the \ho to optimize the pose of each point cloud frame, subsequently enhancing the estimation of the extrinsics and time offsets. 
We believe our method offers a universal solution for LiDAR calibration, unaffected by the calibration environment or the type of LiDAR. This will provide a convenient and precise calibration tool for downstream tasks, allowing them to focus more on their primary objectives without being influenced by preliminary tasks.




\bibliographystyle{IEEEtran}
\bibliography{IEEEabrv, references}

\end{document}